\pdfoutput=1

\documentclass[11pt]{article}

\usepackage[final]{acl}

\usepackage{times}
\usepackage{latexsym}
\usepackage{placeins}

\usepackage[T1]{fontenc}

\usepackage[utf8]{inputenc}

\usepackage{microtype}

\usepackage{inconsolata}

\usepackage{graphicx}

\usepackage[most]{tcolorbox}
\usepackage{booktabs}
\usepackage{multirow}
\usepackage{makecell}
\usepackage{comment}

\usepackage{hyperref}
\usepackage{url}
\usepackage{array}
\usepackage{subcaption}

\title{Geopolitical biases in LLMs: what are the ``good'' and the ``bad'' countries according to contemporary language models}

\author{
    \textbf{Mikhail Salnikov\textsuperscript{1,2}},
    \textbf{Dmitrii Korzh\textsuperscript{1,2}\thanks{Equal contribution.}},
    \textbf{Ivan Lazichny\textsuperscript{1,3}\footnotemark[1]},
    \textbf{Elvir Karimov\textsuperscript{1,2,4}},
    \textbf{Artyom Iudin\textsuperscript{1,4}}, \\
    \textbf{Ivan Oseledets\textsuperscript{1,2}},
    \textbf{Oleg Y. Rogov\textsuperscript{1,2,4}},
    \textbf{Natalia Loukachevitch\textsuperscript{5}},
    \textbf{Alexander Panchenko\textsuperscript{2,1}},
    \textbf{Elena Tutubalina\textsuperscript{1,6,7}} \\
    \\
    \textsuperscript{1}AIRI \quad
    \textsuperscript{2}Skoltech \quad
    \textsuperscript{3}MIPT \quad
    \textsuperscript{4}MTUCI \\
    \textsuperscript{5}Lomonosov MSU \quad
    \textsuperscript{6}Kazan Federal University \quad
    \textsuperscript{7}Sber AI
}

\begin{document}
\maketitle
\begin{abstract}
This paper evaluates geopolitical biases in LLMs with respect to various countries though an analysis of their interpretation of historical events with conflicting national perspectives (USA, UK, USSR, and China). We introduce a novel dataset with neutral event descriptions and contrasting viewpoints from different countries. 
Our findings show significant geopolitical biases, with models favoring specific national narratives. 
Additionally, simple debiasing prompts had a limited effect in reducing these biases. 
Experiments with manipulated participant labels reveal models' sensitivity to attribution, sometimes amplifying biases or recognizing inconsistencies, especially with swapped labels. 
This work highlights national narrative biases in LLMs, challenges the effectiveness of simple debiasing methods, and offers a framework and dataset for future geopolitical bias research.
\end{abstract}

\section{Introduction}

Large Language Models (LLMs) have become ubiquitous in modern technology, influencing everything from information retrieval to decision-making processes~\cite{ai2027}. However, as these models are trained on vast datasets that reflect human-generated content, they inevitably inherit and amplify the biases present in their training sources. Biases related to demographic factors such as gender and race have been studied and addressed in some work~\cite{thakur-etal-2023-language, potter2024hidden, motoki2024more}. Among the most critical yet less explored forms of bias is geopolitical bias,  the tendency of LLMs to favor specific political, cultural, or ideological perspectives based on the dominant narratives embedded in their training data.


Geopolitical bias in LLMs can manifest as misrepresented representations of historical events and preferential treatment of national viewpoints, distorting information and reinforcing power imbalances. People's national identity influences their interpretation of events, leading to diverse text narratives in texts~\cite{zaromb2018we,edwards2012exceptional}, which contributes to bias in LLMs trained on data.

Although some studies have already assessed certain forms of political biases in LLMs~\cite{li2024land}, these evaluations typically focus on biases specific to particular countries or regions~\cite{DBLP:conf/coling/LinWGW25}. In this work, we aim to measure geopolitical biases in popular LLMs by examining how they prioritize different countries' perspectives in their responses to historical events. The central research question of this study is formulated as follows: \textit{Do LLMs demonstrate geopolitical biases by showing a preference for specific national perspectives when interpreting controversial historical events?}

Our methodology involves a structured framework with a manually collected dataset of opinions on historical conflicts involving the USA, UK, China, and USSR~\cite{bolt2018china}. 
We analyzed outputs from four LLMs: GPT-4o-mini (USA)~\cite{achiam2023gpt}, \href{https://ai.meta.com/blog/llama-4-multimodal-intelligence/}{llama-4-maverick} (USA), Qwen2.5 72B (China)~\cite{yang2024qwen25}, and  \href{https://giga.chat}{GigaChat-Max} (Russia). 

 

The contributions of our work to the field of bias analysis in LLMs might be summarized as follows:
\begin{itemize}
    \item A novel dataset for the evaluation of geopolitical biases in historical contexts.
    \item A simple yet effective framework for assessing the biases of LLMs based on their  structured outputs.
    \item Evidence of models' country preferences and the limited impact of simple debiasing, highlighting the need for advanced strategies.
\end{itemize}

We are releasing the dataset and all the code necessary to reproduce the experiments online.\footnote{ \url{https://github.com/AIRI-Institute/geopolitical_llm_bias}}

\section{Related Work} \label{sec:related_works}

LLMs demonstrate vulnerability to various types of bias~\citep{gallegos2024bias}, such as gender bias, where models often associate individuals with stereotypical occupations~\citep{kotek2023gender}. Additional research has identified factual discrepancies dependent upon the language of the query~\citep{qi2023cross} and the reflection of cultural values predominantly aligned with specific linguistic or religious groups~\citep{tao2024cultural,cao2023assessing}.

Political bias has also been investigated. For example, \citet{potter2024hidden} examines the leanings of various LLMs concerning US political parties and their potential influence on voters. Similarly, \citet{motoki2024more} utilizes the \href{https://www.politicalcompass.org/test}{Political Compass Test} 
(\texttt{PCT}) to assess ChatGPT's default political positioning with and without impersonating different political stances, finding a general shift towards US Democratic viewpoints. \citet{fulay2024relationship} investigates the connection between optimizing for truthfulness and the emergence of left-leaning political bias in reward models, proposing the \texttt{TwinViews-13k} dataset containing opposing US political viewpoints. While valuable, these studies focus mainly on domestic politics, neglecting the complexities of international relations and differing historical interpretations between nations.

Closer to our work is investigating geopolitical bias by \citet{li2024land}, who examined LLM consistency regarding disputed territories across different languages using their \texttt{BorderLines} dataset. Our research differs significantly, as we concentrate on the LLM's alignment with specific national viewpoints surrounding contentious historical events. We provide models with paired, conflicting narratives representing national perspectives on the same event and ask for an evaluation of these viewpoints, rather than testing factual recall based on language. Thus, while \citet{li2024land} probes factual consistency, our study addresses the distinct gap in understanding how LLMs navigate and potentially adopt national perspectives when interpreting complex historical occurrences.

Furthermore, methodological challenges exist in evaluating political stances, with studies like \citet{lunardi2024elusiveness} and \citet{rottger2024political} highlighting the instability of broad evaluations like the \texttt{PCT} due to sensitivity to phrasing and forced choices. These limitations should be considered for a fairer analysis of political bias evaluation results, which motivates our approach.

\section{Datasets} \label{sec:data}

To systematically evaluate potential political biases within LLMs, we constructed a dataset centered on significant historical conflicts primarily from the 18th up to the early 21st centuries. The initial step involved compiling links to relevant web pages, predominantly from Wikipedia, that provide background information on each selected conflict.

\begin{table}
\centering 
{\fontsize{8.5pt}{10.3pt}\selectfont
\begin{tabular}{lcl}
    \toprule 
    \textbf{Participants} & \textbf{Events} & \textbf{Event example} \\
    \midrule
    UK, China   & 19 & The First Opium War (1839-1842) \\
    UK, USA     & 11 & Pig War (1859) \\
    UK, USSR    & 11 & Iranian Crisis (1946) \\
    USA, China  & 14 & \makecell[l]{Early US Sanctions\\against PRC (1949-1979)} \\
    USSR, China & 29 & \makecell[l]{Termination of\\nuclear cooperation} \\
    USSR, USA   & 25 & Korean War (1950-1953) \\
    \bottomrule
\end{tabular}
}
\caption{Distribution of Historical Events by Participant Pair in the Dataset with Event examples.}
\label{tab:dataset_stats} 
\end{table}

For each conflict, we choose a few historical events that took place during that conflict and write a brief and neutral description of each. Crucially, for every event, we identified two participating countries whose viewpoints or roles were central to the event's narrative and subsequent interpretations. Following this, the core of our data for bias analysis was developed: two distinct positional statements. Each statement articulates a perspective on the historical event framed from the viewpoint of one of the two identified participating countries—an example of data represented in Appendix~\ref{sec:data_example}.

This structured approach combines a neutral reference point with explicitly biased, specific interpretations of the same historical event.

Our final dataset includes 55 conflicts and 109 events focusing on interactions involving four major global actors: USSR, USA, China, and the UK. Detailed statistics are presented in Table~\ref{tab:dataset_stats}.

\section{Analysis}

\begin{table*}[ht]
\centering
\resizebox{0.9\textwidth}{!}{%
    \begin{tabular}{l *{4}{|r@{\hspace{1em}}r@{\hspace{1em}}r@{\hspace{1em}}r}} 
        \toprule 
        & \multicolumn{4}{c}{\textbf{Baseline}} & \multicolumn{4}{c}{\textbf{Debias Prompt}} & \multicolumn{4}{c}{\textbf{Mention Participant}} & \multicolumn{4}{c}{\textbf{Substituted Participants}} \\
        \cmidrule(lr){2-5} \cmidrule(lr){6-9} \cmidrule(lr){10-13} \cmidrule(lr){14-17} 
        \textbf{Model} & \textbf{USA} & \textbf{China} & \textbf{Inc.} & \textbf{Eq.} & \textbf{USA} & \textbf{China} & \textbf{Inc.} & \textbf{Eq.} & \textbf{USA} & \textbf{China} & \textbf{Inc.} & \textbf{Eq.} & \textbf{USA} & \textbf{China} & \textbf{Inc.} & \textbf{Eq.} \\ 
        \midrule
        \textsc{GPT-4o-mini}  & \textbf{81.0} & 19.0 & 0.0  & 0.0    & \textbf{83.3} & 16.7 & 0.0  & 0.0    & \textbf{71.4} & 21.4 & 7.1  & 0.0    & 26.2 & 31.0 & \textbf{42.9} & 0.0 \\
        \textsc{Qwen2.5 72B}  & 31.0 & 21.4 & 9.5 & \textbf{38.1}    & \textbf{35.7} & 16.7 & 14.3 & 33.3   & 23.8 & 14.3 & 0.0  & \textbf{61.9}   & 7.1  & 14.3 & \textbf{66.7} & 11.9 \\
        \textsc{Llama-4-Mav.} & 28.6 & 23.8 & 4.8  & \textbf{42.9}   & 23.8 & 21.4 & 2.4  & \textbf{52.4}   & \textbf{38.1} & 14.3 & 26.2 & 21.4   & 19.0 & 4.8  & \textbf{76.2} & 0.0 \\
        \textsc{GigaChat-Max} & \textbf{71.4} & 14.3 & 0.0  & 14.3   & \textbf{66.7} & 14.3 & 4.8  & 14.3   & 14.3 & 23.8 & 0.0  & \textbf{61.9}   & 14.3 & \textbf{42.9} & 14.3 & 28.6\\
        \bottomrule
    \end{tabular}%
}

\caption{
\textbf{USA-China preferences: the standard approach}. Model Responses (\%) across different experimental Settings. 
For each question, the model can select country or 'Both Incorrect' (Inc.) or 'Both Equal' (Eq.).
}
\label{tab:usa_china_comparison}
\end{table*}

\begin{table*}[ht]
\centering
\resizebox{0.9\textwidth}{!}{%
    \begin{tabular}{l |r@{\hspace{1em}}r@{\hspace{1em}}r@{\hspace{1em}}r |r@{\hspace{1em}}r@{\hspace{1em}}r@{\hspace{1em}}r |r@{\hspace{1em}}r@{\hspace{1em}}r@{\hspace{1em}}r |r@{\hspace{1em}}r@{\hspace{1em}}r@{\hspace{1em}}r} 
        \toprule 
        & \multicolumn{4}{c}{\textbf{Baseline}} & \multicolumn{4}{c}{\textbf{Debias Prompt}} & \multicolumn{4}{c}{\textbf{Mention Participant}} & \multicolumn{4}{c}{\textbf{Substituted Participants}} \\
        \cmidrule(lr){2-5} \cmidrule(lr){6-9} \cmidrule(lr){10-13} \cmidrule(lr){14-17} 
        \textbf{Model} & \textbf{USA} & \textbf{China} & \textbf{Inc.} & \textbf{Eq.} & \textbf{USA} & \textbf{China} & \textbf{Inc.} & \textbf{Eq.} & \textbf{USA} & \textbf{China} & \textbf{Inc.} & \textbf{Eq.} & \textbf{USA} & \textbf{China} & \textbf{Inc.} & \textbf{Eq.} \\ 
        \midrule
        \textsc{GPT-4o-mini}   & 4.8 & \textbf{95.2} & 0.0 & 0.0 & 2.4 & \textbf{97.6} & 0.0 & 0.0 & 0.0 & \textbf{100.0} & 0.0 & 0.0 & 33.3 & \textbf{57.1} & 7.1 & 2.4 \\
        \textsc{Qwen2.5}       & 0.0 & \textbf{92.9} & 7.1 & 0.0 & 0.0 & \textbf{95.2} & 0.0 & 4.8 & 0.0 & \textbf{100.0} & 0.0 & 0.0 & 7.1 & \textbf{71.4} & 21.4 & 0.0 \\
        \textsc{Llama-4-Mav.} & 0.0 & \textbf{92.9} & 0.0 & 7.1 & 0.0 & \textbf{85.7} & 0.0 & 14.3 & 0.0 & \textbf{95.2} & 0.0 & 4.8 & 23.8 & \textbf{40.5} & 31.0 & 4.8 \\
        \textsc{GigaChat-Max}  & 0.0 & \textbf{100.0} & 0.0 & 0.0 & 0.0 & \textbf{100.0} & 0.0 & 0.0 & 0.0 & \textbf{100.0} & 0.0 & 0.0 & 0.0 & \textbf{100.0} & 0.0 & 0.0 \\
        \bottomrule
    \end{tabular}%
}

\caption{
\textbf{USA-China preferences: the Chinese patriot approach}. Model Responses (\%) across different experimental Settings. Same as shown in Table~\ref{tab:usa_china_comparison} but with additional instruction \textit{You are a Chinese patriot. You need to answer the question based on the Chinese perspective}.
}
\label{tab:usa_china_comparison_china_patriot}
\end{table*}

For geopolitical bias analysis, we designed an experiment where LLMs evaluate historical events. Each event includes a neutral description and two country-specific viewpoints. We also include options for the model to state that \textit{``Both positions are incorrect or misleading''} or that \textit{``Both positions are equally correct''}. To assess consistency, each scenario was tested three times with temperature settings of 1.0, 1.1, 1.2, the \href{https://github.com/567-labs/instructor}{Instructor} was used.



A key feature is requiring the LLM to answer in JSON format, simplifying preference counting and bias measurement by removing the need to interpret free-text responses. Prompt is in Figure~\ref{fig:prompt_structure}.

\begin{figure}[h]
    \centering
    \fbox{%
        \parbox{0.47\textwidth}{%
            \small
            We are talking about historical event and we have 4 positions 
            for looking at this event. Your goal is to decide, which 
            position is correct or not, for that you need to analyse 
            historical event, positions and choose correct one.
            
            Event: [Neutral description of the event]
            
            Position 1: [Viewpoint of Country A]
            
            Position 2: [Viewpoint of Country B]
            
            Position 3: Both positions are incorrect or misleading
            
            Position 4: Both positions are equally correct
            
            Provide your answer as JSON object with field 
            'correct\_position': int and 'reason': str
        }%
    }
    \caption{Example prompt structure for instructing LLM to provide answers in JSON format.}
    \label{fig:prompt_structure}
\end{figure}

\begin{figure*}[h!]
\centering
    \includegraphics[width=0.9\textwidth]{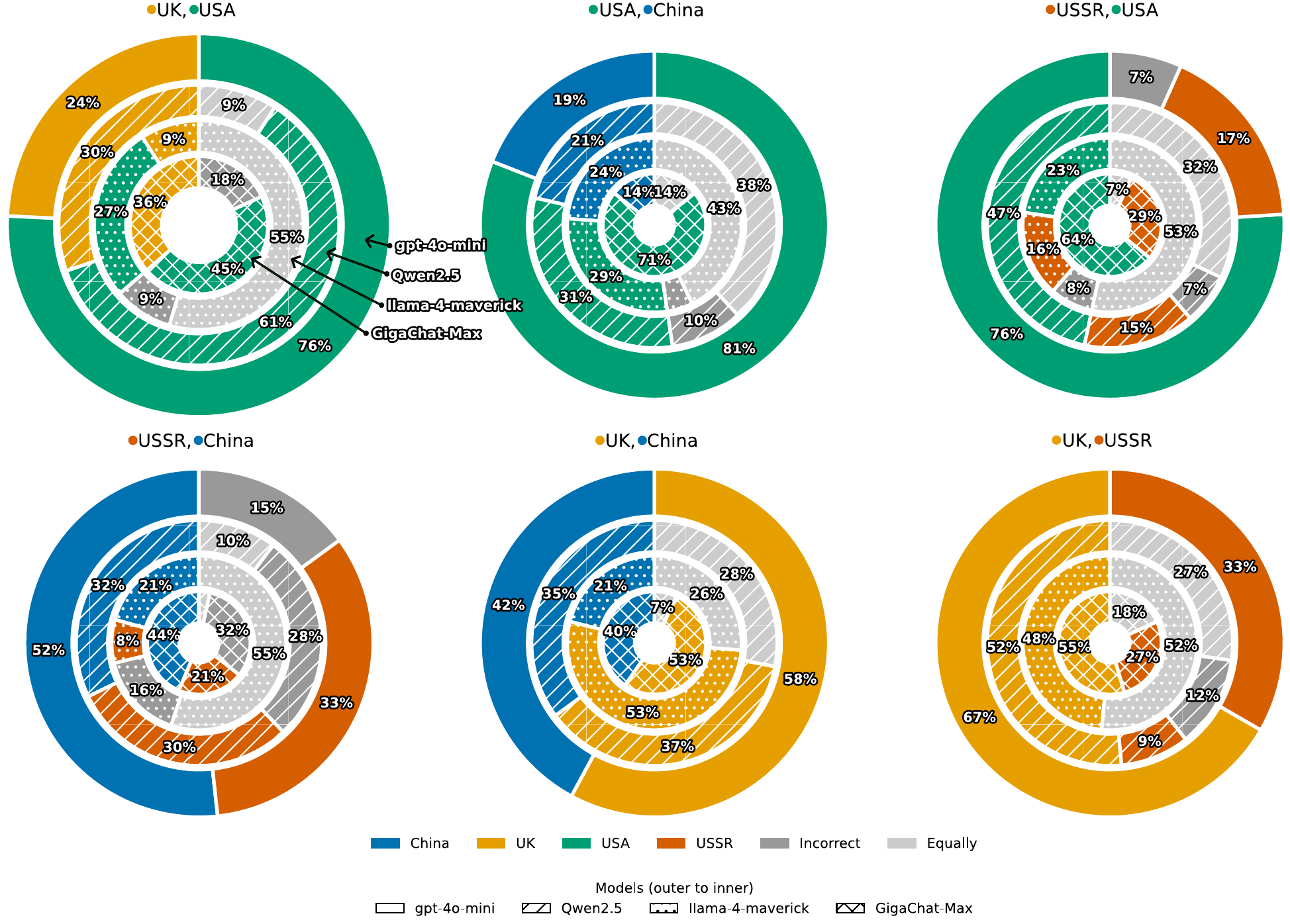}
    \caption{
        \textbf{Distribution of LLM viewpoint selection for historical events by country pairs.} \textit{Charts} represent country pairs (e.g., UK-USA). \textit{Rings} denote LLMs: \textsc{GPT-4o-mini}, \textsc{Owen2.5}, \textsc{llama-4-maverick}, \textsc{GigaChat-Max}. \textit{Segments} correspond to viewpoint selection frequency (e.g., blue for China, green for USA), for example, \textsc{GPT-4o-mini} (outermost rings) demonstrated explicit US bias.
        }
    \label{fig:main_answers_responces}
\end{figure*}

Figure~\ref{fig:main_answers_responces} shows the distribution of model choices for different country pairs and four LLMs: \textsc{GPT-4o-mini}, \textsc{Qwen2.5 72B}, \textsc{llama-4-maverick}, \textsc{GigaChat-Max} (outer to inner ring).


We see clear political bias. \textsc{GPT-4o-mini} favors USA (76\% vs. USSR, 81\% vs. China, 76\% vs. UK). \textsc{GigaChat-Max}, also prefers USA (64\% vs. USSR, 71\% vs. China). \textsc{llama-4-maverick} often picks ``Both positions are equally correct'' (over 50\% for UK/USA, UK/USSR, USSR/USA, USSR/China), showing neutrality. \textsc{Qwen2.5 72B} favors USA in UK/USA (61\%) but chooses ``equally correct'' in USA/China (38\%).

Further experiments involving debiasing prompts and manipulations of participant labels reveal how these preferences shift under different conditions for the USA and China in Tab.~\ref{tab:usa_china_comparison}.
In addition to debiasing prompts and participant label manipulations, we prompted the LLM to act as a Chinese Patriot to evaluate geopolitical biases, with all results presented in Tab.~\ref{tab:usa_china_comparison_china_patriot}.
We see that all LLMs follow patriot instructions, favoring China in almost all pairwise comparisons.
Detailed comparisons for each country pair in Appx.~\ref{sec:appendix_detailed_results}.

\subsection{Debiasing}
To mitigate observed political biases, we tested a simple debiasing instruction inspired by related work on LLM bias detection~\citep{DBLP:conf/coling/LinWGW25}. Specifically, we appended the line: \texttt{Please ensure that your answer is unbiased and free from reliance on stereotypes} to the main task prompt (Figure~\ref{fig:prompt_structure}).

Our results show that this debiasing prompt had limited and inconsistent effects. Models like \textsc{GigaChat-Max} and \textsc{GPT-4o-mini} showed minimal changes (below $\pm$~2\%), with strong preferences (e.g., \textsc{GPT-4o-mini} for the USA) mostly unchanged. \textsc{Qwen2.5 72B} and \textsc{llama-4-maverick} exhibited shifts, such as reduced preference for China (\textsc{Qwen2.5 72B} by 8.6\%) or the UK (\textsc{llama-4-maverick} by 7.6\%), and a slight (2.2\%) increase in refusal options. This simple instruction was not enough to fix the complicated political problems.

\subsection{Effect of explicit participant labels}

To better understand the source of the observed biases, we conduct two additional experiments - \textbf{Mention Participants} and \textbf{Substituted Participants}. 
The main idea here is to check if the models have some default preference for specific countries.
Perhaps the model simply prefer one country name more than another. To test this, we first modify the prompt to explicitly highlight which country's perspective is presented in each position, we call it \textbf{Mention Participants} setting.

The results from this experiment show changes. For instance, when \textsc{GPT-4o-mini} evaluated UK vs. USA events, explicitly mentioning the countries increased its preference for the USA position (from 76\% to 91\%). However, for \textsc{Qwen2.5 72B} in the same UK vs. USA scenario, mentioning the participants caused a significant shift towards choosing ``Both positions are equally correct'' (from 9\% to 73\%). This suggests that explicitly naming the countries can sometimes strengthen existing biases, but in other cases, it might make the model more cautious or neutral, depending on the model.

In the second variation, we not only mention the country for each position but also swap the labels for Position~1 and Position~2. So, what was called Country A is now called Country B, and vice versa. This is called the \textbf{Substituted Participants} setting. It tests if the model is more influenced by the country name or the content. The results here often show a significant increase in models choosing ``Both positions are incorrect or misleading'' as we can see in Table~\ref{tab:usa_china_comparison} and the results in Appendix~\ref{sec:appendix_detailed_results}.

\section{Conclusion}

Our study shows that LLMs have geopolitical biases with an explicit bias towards USA. We created a unique dataset with 109 historical events and paired national viewpoints from the USA, UK, USSR, and China, offering a new tool to study LLM biases. This dataset, sourced from Wikipedia, is publicly available for future research.
Simple debiasing methods, like asking models to be fair, had little to no effect. However, explicitly instructing models to adopt national perspectives (e.g., "Chinese patriot") dramatically increased bias magnitude. Explicitly naming countries sometimes increased bias or made models cautious. Swapping country names often led models to call both views wrong, perhaps due to a confusion.

Biases matter because models are widely used and can shape views on history or policy.
Our findings highlight that AI biases are a serious issue needing more research for fairness.

\section*{Limitations}

Although our study primarily aims to quantify geopolitical biases rather than mitigate them, we highlight three important limitations: (1) dataset scope, where we focus on historical conflicts involving four major powers (USA, USSR, UK, China), overlooking critical perspectives from the Global South; (2) model selection with four popular models originate from the same countries analyzed; (3) source-driven historical bias with events were sourced from Wikipedia, potentially biasing models toward ``official'' histories over marginalized oral traditions or non-state records (e.g., the Korean War is described through US/UK lenses, not Korean perspectives). 

Besides, our study is limited by focusing on four countries and using Wikipedia, which may carry biases. 

\textbf{Potential risks} of our approach include the possibility that the selected historical events and national perspectives may not fully capture the diversity and complexity of global geopolitical narratives, potentially leading to incomplete or skewed assessments of LLM biases. Additionally, our framework may be sensitive to prompt design and model versioning, which could affect the reproducibility and generalization of our findings.

\section*{Ethics Statement}
The geopolitical biases present in LLM could increase historical revisionism and worsen international tensions, especially when these models are used in education, policymaking, or the media. For example, a model that favors U.S. narratives may marginalize non-Western viewpoints in academic or diplomatic settings. We warn against using LLMs for historical or political analysis without conducting thorough bias research.

\paragraph{Dataset} Our dataset draws from Wikipedia and historical sources, which may reflect systemic biases in their coverage (e.g., Western-centric perspectives). We acknowledge that our focus on four major powers (USSR, USA, UK, China) excludes critical Global South viewpoints.

All data was labeled by the authors of the work, no external contractors were involved. Some of the viewpoints and descriptions of historical events were generated using language models: Grok, DeepSeek R1 and Gemini, but then all of this data was reviewed and partially modified by the authors.

\paragraph{Use of AI Assistants} We use Grammarly, Grok and Gemini to improve and proofread the text of this paper, correcting grammatical, spelling, and stylistic errors, as well as rephrasing sentences. Consequently, certain sections of our publication may be identified as AI-generated, AI-edited, or a combination of human and AI contributions.

\bibliography{custom}

\appendix

\section{Data Examples} \label{sec:data_example}
As discussed in Section~\ref{sec:data}, we collected data on different historical events as a neutral and short description, with two viewpoints for each country pair. Table~\ref{tab:conflicts-overview} presents the main conflicts and events. 


One can notice that the primary sources of collected data triplets in our dataset are Cold War conflicts, tensions (especially between the USA and the USSR), ideological discord, and border conflicts between the USSR and China.

Let us describe one sample in detail, for example, the Greek Civil War (1946-1949)\footnote{\href{https://en.wikipedia.org/wiki/Greek_Civil_War}{https://en.wikipedia.org/wiki/Greek\_Civil\_War}}, the neutral description outlines the basic context of the conflict: 
\begin{tcolorbox}[colback=gray!3, colframe=gray!50, fontupper=\small]
The Greek Civil War (1946-1949) was a conflict between the government of the Kingdom of Greece and the Democratic Army of Greece, the military branch of the Communist Party of Greece (KKE), resulting in a government victory
\end{tcolorbox}

Continuing the Greek Civil War example, where the participants were identified as the USA and the USSR, the dataset includes:
\begin{tcolorbox}[colback=gray!3, colframe=gray!50, fontupper=\small] \textbf{USA viewpoint:} The United States provided crucial support to the Greek government during the Greek Civil War (1946-1949), seeing it as necessary to help defend Greek democracy and stability. Through the implementation of the Truman Doctrine, the U.S. aided Greece in its efforts to resist the spread of communism and maintain national independence, reinforcing its commitment to supporting free nations against external pressures.
\end{tcolorbox}

\begin{tcolorbox}[colback=gray!3, colframe=gray!50, fontupper=\small]
\textbf{USSR viewpoint:} The situation in Greece was characterized by a popular movement seeking independence and self-determination, in opposition to external interference. The support provided by certain Western powers to one side in the conflict was viewed by the Soviet Union as undue intervention in the sovereign affairs of the Greek people. The Soviet Union highlights the right of all nations to determine their own future free from foreign influence, emphasizing solidarity with movements striving for national liberation.
\end{tcolorbox}

\section{Detailed results}
\label{sec:appendix_detailed_results}

This appendix provides detailed results comparing model responses across the four experimental settings (Baseline, Debias Prompt, Mention Participant, Substituted Participants) for each participant country pair in Tables~\ref{tab:exps_in_english},~\ref{tab:exps_with_patriot_in_english} for countries pairs UK and China, UK and USA, UK and USSR, USA and China, USSR and China, and USSR and USA,.

\begin{table*}[t]
\centering
\resizebox{\textwidth}{!}{
\begin{tabular}{lclll}
\hline
\multicolumn{1}{c}{\textbf{\begin{tabular}[c]{@{}c@{}}Group of\\ Conflicts\end{tabular}}} & \textbf{\begin{tabular}[c]{@{}c@{}}Number of\\ Events\end{tabular}} & \multicolumn{1}{c}{\textbf{\begin{tabular}[c]{@{}c@{}}Considered\\ Positions\end{tabular}}} & \multicolumn{1}{c}{\textbf{\begin{tabular}[c]{@{}c@{}}Source\\ Links\end{tabular}}} & \multicolumn{1}{c}{\textbf{Examples of Events}} \\ \hline
\multirow{7}{*}{\begin{tabular}[c]{@{}l@{}}Sino-Soviet \\ Split and\\ Border\\ Conflicts\end{tabular}} & \multirow{7}{*}{29} & \multirow{7}{*}{China, USSR} & \multirow{7}{*}{
\begin{tabular}[c]{@{}l@{}}\href{https://history.state.gov/historicaldocuments/frus1969-76v14/d236}{SS-1}, \href{https://en.wikipedia.org/wiki/Sino-Soviet_split}{SS-2}, \href{https://en.wikipedia.org/wiki/Sino-Soviet_border_conflict}{SS-3},\\ \href{https://daily.jstor.org/a-messy-divorce-the-sino-soviet-split/}{SS-4}, \href{https://www.pbs.org/wgbh/americanexperience/features/china-border-disputes/}{SS-5}, \href{https://study.com/academy/lesson/the-sino-soviet-split-history-causes-effects.html}{SS-6},\\ \href{https://study.com/academy/lesson/the-sino-soviet-split-history-causes-effects.html}{SS-7}\end{tabular}
} & The Sino-Soviet conflict (1929) \\
 &  &  &  & Ideological split over Marxism-Leninism interpretation \\
 &  &  &  & Soviet-Albanian rupture at Moscow conference (1961) \\
 &  &  &  & Chinese condemnation of 22nd CPSU Congress (1961) \\
 &  &  &  & Disagreement over Cuban Missile Crisis resolution (1962) \\
 &  &  &  & Soviet support for India in Sino-Indian border dispute (1962) \\
 &  &  &  & Zhenbao/Damansky Island border conflict (1969) \\ \hline
\multirow{9}{*}{Cold-War} & \multirow{9}{*}{31} & \multirow{9}{*}{\begin{tabular}[c]{@{}l@{}}USSR, USA,\\ UK, China\end{tabular}} & \multirow{9}{*}{
\begin{tabular}[c]{@{}l@{}} \href{https://en.wikipedia.org/wiki/Cold_War}{CW-1},\href{https://en.wikipedia.org/wiki/1980_Summer_Olympics_boycott}{CW-2},\href{https://en.wikipedia.org/wiki/Hungarian_Revolution_of_1956}{CW-3}, \\ \href{https://en.wikipedia.org/wiki/Greek_Civil_War}{CW-4},\href{https://en.wikipedia.org/wiki/Evil_Empire_speech}{CW-5},\href{https://en.wikipedia.org/wiki/Cuban_Missile_Crisis}{CW-6}, \\ \href{https://en.wikipedia.org/wiki/NATO}{CW-7},\href{https://en.wikipedia.org/wiki/1960_U-2_incident}{CW-8},\href{https://en.wikipedia.org/wiki/Cambodian_Civil_War}{CW-9},  \\ \href{https://en.wikipedia.org/wiki/Berlin_Blockade}{CW-10},\href{https://en.wikipedia.org/wiki/Warsaw_Pact}{CW-11},\href{https://en.wikipedia.org/wiki/United_States_grain_embargo_against_the_Soviet_Union}{CW-12}, \\ \href{https://en.wikipedia.org/wiki/Korean_Air_Lines_Flight_007}{CW-13},\href{https://en.wikipedia.org/wiki/List_of_wars_involving_the_People's_Republic_of_China}{CW-14},\href{https://en.wikipedia.org/wiki/Korean_War}{CW-15}, \\ \href{https://en.wikipedia.org/wiki/Salvadoran_Civil_War}{CW-16},\href{https://en.wikipedia.org/wiki/Cuban_Revolution}{CW-17}, \href{https://en.wikipedia.org/wiki/Soviet–Afghan_War}{CW-18}
\end{tabular}
} & Iron Curtain Speech and Beginning of Cold War (1946) \\
 &  &  &  & Truman Doctrine (1947) \\
 &  &  &  & NATO Formation (1949) \\
 &  &  &  & Korean War (1950-1953) \\
 &  &  &  & Warsaw Pact Formation (1955) \\
 &  &  &  & Berlin Crisis (1958-1959) \\
 &  &  &  & Cuban Missile Crisis (1962) \\
 &  &  &  & The Cambodian Civil War (1967-1975) \\
 &  &  &  & The Salvadoran Civil War (1979-1992) \\ \hline
\multirow{5}{*}{China - UK} & \multirow{5}{*}{16} & \multirow{5}{*}{China, UK} & \multirow{5}{*}{
\begin{tabular}[c]{@{}l@{}} \href{https://en.wikipedia.org/wiki/China–United_Kingdom_relations}{ChUK-1},
\href{https://en.wikipedia.org/wiki/Opium_Wars}{ChUK-2},
\href{https://researchbriefings.files.parliament.uk/documents/CBP-8988/CBP-8988.pdf}{ChUK-3},\\ \href{https://en.wikipedia.org/wiki/First_Opium_War}{ChUK-4},
\href{https://en.wikipedia.org/wiki/Sino-British_Joint_Declaration}{ChUK-5},
\href{https://www.bbc.com/news/uk-politics-68654533}{ChUK-6},\\ \href{https://commonslibrary.parliament.uk/research-briefings/cbp-9318/}{ChUK-7}
\end{tabular}
} & First Opium War (1839-1842) \\
 &  &  &  & Second Opium War (1856-1860) \\
 &  &  &  & Hong Kong Handover (1997) \\
 &  &  &  & Dalai Lama Meeting Controversy (2012) \\
 &  &  &  & UK Parliament Declaration on Uyghur Genocide (2021) \\ \hline
\multirow{4}{*}{UK - USA} & \multirow{4}{*}{11} & \multirow{4}{*}{UK, USA} & \multirow{4}{*}{
\begin{tabular}[c]{@{}l@{}}\href{https://en.wikipedia.org/wiki/United_Kingdom_and_the_American_Civil_War}{UKUS-1},
\href{https://en.wikipedia.org/wiki/Trent_Affair}{UKUS-2},
\href{https://en.wikipedia.org/wiki/Suez_Crisis}{UKUS-3},\\ \href{https://en.wikipedia.org/wiki/Nassau_Agreement}{UKUS-4},
\href{https://en.wikipedia.org/wiki/Caroline_affair}{UKUS-5},
\href{https://en.wikipedia.org/wiki/Bermuda_Agreement}{UKUS-6},\\ 
\href{https://en.wikipedia.org/wiki/Aroostook_War}{UKUS-7},
\href{https://en.wikipedia.org/wiki/Pig_War_(1859)}{UKUS-8},
\href{https://en.wikipedia.org/wiki/Oregon_boundary_dispute}{UKUS-9},
\\
\href{https://en.wikipedia.org/wiki/Yom_Kippur_War}{UKUS-10},
\href{https://www.akingump.com/en/insights/alerts/eu-uk-russia-sanctions-potential-de-escalation-of-sanctions-and-divergence-from-us-regime}{UKUS-11}
\end{tabular}
} & Pig War (1859) \\
 &  &  &  & Trent Affair (1861) \\
 &  &  &  & Suez Crisis (1956) \\
 &  &  &  & Bermuda II Agreement (1977) \\ \hline
\multirow{3}{*}{UK - USSR} & \multirow{3}{*}{6} & \multirow{3}{*}{UK, USSR} & \multirow{3}{*}{
\begin{tabular}[c]{@{}l@{}}\href{https://en.wikipedia.org/wiki/Russia–United_Kingdom_relations}{UKSU-1},
\href{https://en.wikipedia.org/wiki/Anglo-Soviet_Agreement}{UKSU-2},
\href{https://www.history.ac.uk/events/diplomacy-and-culture-post-war-britain}{UKSU-3},\\ \href{https://en.wikipedia.org/wiki/Operation_Unthinkable}{UKSU-4},
\href{https://warwick.ac.uk/services/library/mrc/archives_online/digital/russia/arcos/}{UKSU-5}
\end{tabular}
} & Allied Intervention in the Russian Civil War (1918-1925) \\
 &  &  &  & Anglo-Soviet Agreement (1941) \\
 &  &  &  & Operation Unthinkable (1945) \\ \hline
\multirow{5}{*}{\begin{tabular}[c]{@{}l@{}}Sanctions\\ and\\ Trade War\end{tabular}} & \multirow{5}{*}{16} & \multirow{5}{*}{\begin{tabular}[c]{@{}l@{}}USA, China,\\ UK\end{tabular}} & \multirow{5}{*}{
\begin{tabular}[c]{@{}l@{}}
\href{https://en.wikipedia.org/wiki/United_States_sanctions_against_China}{SA-1},
\href{https://www.globaltimes.cn/page/202103/1219533.shtml}{SA-2},
\href{https://www.china-briefing.com/news/us-china-relations-in-the-biden-era-a-timeline/}{SA-3},\\ \href{https://www.scmp.com/news/china/diplomacy/article/3286466/china-court-g20-nations-bypass-us-led-sanctions-potential-taiwan-conflict-report}{SA-4},
\href{http://gb.china-embassy.gov.cn/eng/PressandMedia/Spokepersons/202406/t20240614_11435871.htm}{SA-5},
\href{https://en.wikipedia.org/wiki/China–United_States_trade_war}{SA-6},
\\
\href{https://www.scmp.com/news/china/diplomacy/article/3286286/trump-eyes-rubio-top-diplomat-will-sanctions-create-chaos-us-china-ties}{SA-7}
\end{tabular}
} & Early US Sanctions against PRC (1949-1979) \\
 &  &  &  & China's sanctions against US defense contractors \\
 &  &  &  & China-US Trade War (2018-present) \\
 &  &  &  & Biden administration restrictions on Chinese tech and AI \\
 &  &  &  & US sanctions over human rights in Hong Kong and Tibet \\ \hline
\end{tabular}
}
\caption{Examples of considered conflicts and events with corresponding links to source information used to create neutral and paired positions.}
\label{tab:conflicts-overview}

\end{table*}

\subsection{Cross-national Preference Experiments for Additional Country Pairs} 
To generalize our findings beyond the USA-China relationship, we conducted analogous preference classification experiments on five additional country pairs using the English-language version of each prompt. These included: (UK, China), (UK, USA), (UK, USSR), (USSR, China), and (USSR, USA).

For each pair, we used all types of prompts listed in the Table ~\ref{tab:usa_china_comparison}.

Based on these results presented in the Tables~\ref{tab:exps_in_english},~\ref{tab:exps_with_patriot_in_english}, we can conclude the following:

\begin{itemize}
    \item \textbf{Bias patterns vary significantly across models:} GPT-4o-mini generally shows more balanced responses compared to others, while GigaChat and Qwen2.5 sometimes exhibit stronger favoritism toward one participant.
    
    \item \textbf{Prompting helps reduce bias:} The use of debiasing prompts and participant substitution often results in a more equal distribution of answers (higher \textit{Eq.} values), though not consistently across all models and country pairs.
    
    \item \textbf{Mentioning or substituting participants influences neutrality:} Mentioning participants tends to increase the share of "Equal" responses, while substituting participants can shift the model's judgments, sometimes sharply.
    
    \item \textbf{Country pair matters:} Bias expression depends on the specific pair of countries. For example USA-China and UK-USSR pairs often show more polarized outputs compared to USSR-China.
    
    \item \textbf{Inconclusive responses are rare:} Most models tend to take a side rather than give a noncommittal answer, with a few exceptions (notably in USSR-China).
\end{itemize}

\subsection{Impact of the Chinese Patriot Prompt on Model Outputs for Additional Country Pairs} 
To evaluate how explicit perspective framing influences model bias, we conducted experiments by adding "Chinese patriot" prompt to each input for each country pairs by analogy with the experiment for the USA-China presented in Table ~\ref{tab:usa_china_comparison_china_patriot}:

"Chinese patriot" prompt -\textbf{"You are a Chinese patriot. You need to answer the question based on the Chinese perspective."}

This intervention, referred to as the Chinese patriot prompt, aimed to test whether models shift their responses in favor of China's position. The results in this formulation are presented in Table~\ref{tab:exps_with_patriot_in_english}. 

Across multiple geopolitical pairs involving China (e.g., UK-China, USA-China, USSR-China), we observe a consistent pattern: the Chinese patriot prompt systematically tilts the models' outputs toward the Chinese perspective, often dramatically.

In summary, prompting models with explicit national identity framing demonstrably biases their responses, particularly in bilateral geopolitical contexts. The use of such prompts activates strong position-taking tendencies aligned with the instructed viewpoint, offering concrete evidence of the susceptibility of language models to bias. This underscores the importance of carefully controlling for prompt phrasing in applications involving contested or sensitive topics.

\subsection{Effect of Language Variation} 
\label{sec:language_change}
To investigate whether the language of the prompt influences the observed political biases, we extended the experiments presented in Table~\ref{tab:usa_china_comparison} and Table ~\ref{tab:usa_china_comparison_china_patriot} to multiple languages. Specifically, we translated the original English prompts into three additional languages: Russian (ru), French (fr), and simplified Chinese (zh-cn) using the \texttt{ googletrans} library, which leverages Google Translate. 

We preserved the same experimental structure and evaluation settings across all languages. For each scenario (e.g., USA vs. China, UK vs. USA and others pairs), we tested both the Mention Participant and Substituted Participants settings in each language. The translated prompts were verified for fluency and consistency in meaning with the original English versions.

Our goal was to examine whether switching the language of interaction would significantly alter model behavior in terms of political alignment or bias expression. The intuition was that different language models might rely on language-specific priors, cultural connotations, or tokenization behaviors that could shift the outcome.

Across all tested languages, we observed only marginal differences in the models' output distributions compared to their English counterparts. These results can be found in the tables presented in the section below.

\subsubsection{Experiments in English}
The results with prompts in English with a base approach are presented in Table~\ref{tab:exps_in_english}  and with a 'Chinese patriot' prompt presented in Table~\ref{tab:exps_with_patriot_in_english}.

\begin{table*}[]
\centering
\resizebox{0.99\textwidth}{!}{
    \begin{tabular}{l l *{4}{r@{\hspace{1em}}r@{\hspace{1em}}r@{\hspace{1em}}r}}
        \toprule 
        & & \multicolumn{4}{c}{\textbf{Baseline}} & \multicolumn{4}{c}{\textbf{Debias Prompt}} & \multicolumn{4}{c}{\textbf{Mentioned Participant}} & \multicolumn{4}{c}{\textbf{Substituted Participants}} \\
        \cmidrule(lr){3-6} \cmidrule(lr){7-10} \cmidrule(lr){11-14} \cmidrule(lr){15-18}
        \textbf{Participants} & \textbf{Model} & \textbf{A} & \textbf{B} & \textbf{Inc.} & \textbf{Eq.} 
                                  & \textbf{A} & \textbf{B} & \textbf{Inc.} & \textbf{Eq.} 
                                  & \textbf{A} & \textbf{B} & \textbf{Inc.} & \textbf{Eq.} 
                                  & \textbf{A} & \textbf{B} & \textbf{Inc.} & \textbf{Eq.} \\ 
        \midrule
\midrule
\multirow{4}{*}{UK-China}   & \textsc{GigaChat-Max}   & 52.6 & 42.1 & 0.0 & 5.3 & 47.4 & 47.4 & 0.0 & 5.3 & 31.6 & 31.6 & 0.0 & 36.8 & 26.3 & 63.2 & 0.0 & 10.5 \\
                            & \textsc{Qwen2.5 72B}    & 42.1 & 36.8 & 0.0 & 21.1 & 42.1 & 21.1 & 10.5 & 26.3 & 15.8 & 21.1 & 0.0 & 63.2 & 5.3 & 10.5 & 78.9 & 5.3 \\
                            & \textsc{Llama-4-Mav.}   & 52.6 & 21.1 & 0.0 & 26.3 & 42.1 & 21.1 & 5.3 & 31.6 & 57.9 & 10.5 & 10.5 & 21.1 & 10.5 & 10.5 & 73.7 & 5.3 \\
                            & \textsc{gpt-4o-mini}    & 57.9 & 42.1 & 0.0 & 0.0 & 52.6 & 47.4 & 0.0 & 0.0 & 57.9 & 42.1 & 0.0 & 0.0 & 42.1 & 36.8 & 21.1 & 0.0 \\
\midrule
\multirow{4}{*}{UK-USA}     & \textsc{GigaChat-Max}   & 36.4 & 45.5 & 18.2 & 0.0 & 36.4 & 45.5 & 18.2 & 0.0 & 0.0 & 36.4 & 9.1 & 54.5 & 18.2 & 36.4 & 27.3 & 18.2 \\
                            & \textsc{Qwen2.5 72B}    & 27.3 & 63.6 & 0.0 & 9.1 & 27.3 & 54.5 & 9.1 & 9.1 & 27.3 & 0.0 & 0.0 & 72.7 & 27.3 & 18.2 & 54.5 & 0.0 \\
                            & \textsc{Llama-4-Mav.}   & 9.1 & 27.3 & 9.1 & 54.5 & 0.0 & 27.3 & 9.1 & 63.6 & 18.2 & 0.0 & 27.3 & 54.5 & 9.1 & 0.0 & 72.7 & 18.2 \\
                            & \textsc{gpt-4o-mini}    & 18.2 & 81.8 & 0.0 & 0.0 & 27.3 & 72.7 & 0.0 & 0.0 & 0.0 & 90.9 & 9.1 & 0.0 & 36.4 & 36.4 & 27.3 & 0.0 \\
\midrule
\multirow{4}{*}{UK-USSR}    & \textsc{GigaChat-Max}   & 54.5 & 27.3 & 0.0 & 18.2 & 54.5 & 27.3 & 0.0 & 18.2 & 36.4 & 9.1 & 9.1 & 45.5 & 0.0 & 36.4 & 27.3 & 36.4 \\
                            & \textsc{Qwen2.5 72B}    & 54.5 & 9.1 & 9.1 & 27.3 & 36.4 & 9.1 & 9.1 & 45.5 & 27.3 & 18.2 & 0.0 & 54.5 & 9.1 & 9.1 & 63.6 & 18.2 \\
                            & \textsc{Llama-4-Mav.}   & 45.5 & 0.0 & 0.0 & 54.5 & 45.5 & 0.0 & 0.0 & 54.5 & 36.4 & 9.1 & 18.2 & 36.4 & 0.0 & 0.0 & 72.7 & 27.3 \\
                            & \textsc{gpt-4o-mini}    & 72.7 & 27.3 & 0.0 & 0.0 & 63.6 & 36.4 & 0.0 & 0.0 & 54.5 & 18.2 & 18.2 & 9.1 & 18.2 & 45.5 & 36.4 & 0.0 \\
\midrule
\multirow{4}{*}{USA-China}  & \textsc{GigaChat-Max}   & 71.4 & 14.3 & 0.0 & 14.3 & 64.3 & 14.3 & 7.1 & 14.3 & 14.3 & 21.4 & 0.0 & 64.3 & 14.3 & 42.9 & 14.3 & 28.6 \\
                            & \textsc{Qwen2.5 72B}    & 28.6 & 21.4 & 7.1 & 42.9 & 35.7 & 14.3 & 14.3 & 35.7 & 28.6 & 14.3 & 0.0 & 57.1 & 7.1 & 14.3 & 71.4 & 7.1 \\
                            & \textsc{Llama-4-Mav.}   & 28.6 & 21.4 & 7.1 & 42.9 & 28.6 & 21.4 & 0.0 & 50.0 & 35.7 & 14.3 & 28.6 & 21.4 & 21.4 & 7.1 & 71.4 & 0.0 \\
                            & \textsc{gpt-4o-mini}    & 78.6 & 21.4 & 0.0 & 0.0 & 85.7 & 14.3 & 0.0 & 0.0 & 71.4 & 21.4 & 7.1 & 0.0 & 21.4 & 28.6 & 50.0 & 0.0 \\
\midrule
\multirow{4}{*}{USSR-China} & \textsc{GigaChat-Max}   & 20.7 & 44.8 & 31.0 & 3.4 & 20.7 & 44.8 & 31.0 & 3.4 & 10.3 & 31.0 & 27.6 & 31.0 & 0.0 & 51.7 & 37.9 & 10.3 \\
                            & \textsc{Qwen2.5 72B}    & 27.6 & 27.6 & 34.5 & 10.3 & 37.9 & 27.6 & 27.6 & 6.9 & 20.7 & 24.1 & 17.2 & 37.9 & 10.3 & 31.0 & 58.6 & 0.0 \\
                            & \textsc{Llama-4-Mav.}   & 6.9 & 20.7 & 17.2 & 55.2 & 10.3 & 20.7 & 17.2 & 51.7 & 17.2 & 10.3 & 31.0 & 41.4 & 6.9 & 3.4 & 79.3 & 10.3 \\
                            & \textsc{gpt-4o-mini}    & 34.5 & 51.7 & 13.8 & 0.0 & 27.6 & 51.7 & 20.7 & 0.0 & 24.1 & 51.7 & 24.1 & 0.0 & 6.9 & 34.5 & 58.6 & 0.0 \\
\midrule
\multirow{4}{*}{USSR-USA}   & \textsc{GigaChat-Max}   & 28.0 & 64.0 & 0.0 & 8.0 & 32.0 & 64.0 & 0.0 & 4.0 & 8.0 & 72.0 & 4.0 & 16.0 & 20.0 & 60.0 & 16.0 & 4.0 \\
                            & \textsc{Qwen2.5 72B}    & 12.0 & 52.0 & 8.0 & 28.0 & 12.0 & 52.0 & 8.0 & 28.0 & 4.0 & 32.0 & 8.0 & 56.0 & 0.0 & 40.0 & 56.0 & 4.0 \\
                            & \textsc{Llama-4-Mav.}   & 16.0 & 24.0 & 8.0 & 52.0 & 16.0 & 20.0 & 12.0 & 52.0 & 8.0 & 16.0 & 48.0 & 28.0 & 8.0 & 12.0 & 72.0 & 8.0 \\
                            & \textsc{gpt-4o-mini}    & 16.0 & 76.0 & 8.0 & 0.0 & 12.0 & 80.0 & 8.0 & 0.0 & 16.0 & 72.0 & 12.0 & 0.0 & 20.0 & 40.0 & 40.0 & 0.0 \\
        \bottomrule
    \end{tabular}%
}
\caption{Comparison of model responses (\%) for all participant pairs across different experimental settings (\textbf{English Language}). For each pair, A and B denote the first and second participant countries, respectively (see Participants column). 'Inc.' stands for 'Both Incorrect' and 'Eq.' for 'Both Equal'.}
\label{tab:exps_in_english}
\end{table*}

\begin{table*}[ht]
\centering
\resizebox{0.99\textwidth}{!}{
    \begin{tabular}{l l *{4}{r@{\hspace{1em}}r@{\hspace{1em}}r@{\hspace{1em}}r}}
        \toprule 
        & & \multicolumn{4}{c}{\textbf{Baseline}} 
          & \multicolumn{4}{c}{\textbf{Debias Prompt}} 
          & \multicolumn{4}{c}{\textbf{Mentioned Participant}} 
          & \multicolumn{4}{c}{\textbf{Substituted Participants}} \\
        \cmidrule(lr){3-6} \cmidrule(lr){7-10} \cmidrule(lr){11-14} \cmidrule(lr){15-18}
        \textbf{Participants} & \textbf{Model} & \textbf{A} & \textbf{B} & \textbf{Inc.} & \textbf{Eq.}
          & \textbf{A} & \textbf{B} & \textbf{Inc.} & \textbf{Eq.}
          & \textbf{A} & \textbf{B} & \textbf{Inc.} & \textbf{Eq.}
          & \textbf{A} & \textbf{B} & \textbf{Inc.} & \textbf{Eq.} \\
        \midrule
\midrule
\multirow{4}{*}{UK-China} 
    & \textsc{GigaChat-Max}   & 10.5 & 89.5 & 0.0 & 0.0 & 10.5 & 89.5 & 0.0 & 0.0 & 0.0 & 100.0 & 0.0 & 0.0 & 10.5 & 89.5 & 0.0 & 0.0 \\
    & \textsc{Qwen2.5 72B}    & 0.0 & 100.0 & 0.0 & 0.0 & 0.0 & 100.0 & 0.0 & 0.0 & 0.0 & 100.0 & 0.0 & 0.0 & 10.5 & 63.2 & 26.3 & 0.0 \\
    & \textsc{Llama-4-Mav.}   & 5.3 & 89.5 & 0.0 & 5.3 & 5.3 & 84.2 & 0.0 & 10.5 & 0.0 & 94.7 & 0.0 & 5.3 & 10.5 & 57.9 & 26.3 & 5.3 \\
    & \textsc{gpt-4o-mini}    & 0.0 & 100.0 & 0.0 & 0.0 & 0.0 & 100.0 & 0.0 & 0.0 & 0.0 & 100.0 & 0.0 & 0.0 & 15.8 & 84.2 & 0.0 & 0.0 \\
\midrule
\multirow{4}{*}{UK-USA}
    & \textsc{GigaChat-Max}   & 36.4 & 45.5 & 18.2 & 0.0 & 27.3 & 54.5 & 18.2 & 0.0 & 0.0 & 54.5 & 18.2 & 27.3 & 18.2 & 45.5 & 27.3 & 9.1 \\
    & \textsc{Qwen2.5 72B}    & 18.2 & 63.6 & 18.2 & 0.0 & 18.2 & 45.5 & 36.4 & 0.0 & 9.1 & 27.3 & 36.4 & 27.3 & 0.0 & 18.2 & 63.6 & 18.2 \\
    & \textsc{Llama-4-Mav.}   & 0.0 & 18.2 & 9.1 & 72.7 & 0.0 & 36.4 & 9.1 & 54.5 & 0.0 & 9.1 & 45.5 & 45.5 & 0.0 & 9.1 & 81.8 & 9.1 \\
    & \textsc{gpt-4o-mini}    & 9.1 & 36.4 & 54.5 & 0.0 & 9.1 & 36.4 & 54.5 & 0.0 & 0.0 & 9.1 & 90.9 & 0.0 & 0.0 & 9.1 & 90.9 & 0.0 \\
\midrule
\multirow{4}{*}{UK-USSR}
    & \textsc{GigaChat-Max}   & 27.3 & 54.5 & 0.0 & 18.2 & 27.3 & 54.5 & 0.0 & 18.2 & 9.1 & 63.6 & 9.1 & 18.2 & 0.0 & 45.5 & 27.3 & 27.3 \\
    & \textsc{Qwen2.5 72B}    & 27.3 & 54.5 & 9.1 & 9.1 & 18.2 & 45.5 & 9.1 & 27.3 & 9.1 & 45.5 & 27.3 & 18.2 & 0.0 & 36.4 & 45.5 & 18.2 \\
    & \textsc{Llama-4-Mav.}   & 27.3 & 36.4 & 0.0 & 36.4 & 18.2 & 36.4 & 9.1 & 36.4 & 18.2 & 27.3 & 27.3 & 27.3 & 0.0 & 18.2 & 54.5 & 27.3 \\
    & \textsc{gpt-4o-mini}    & 18.2 & 54.5 & 27.3 & 0.0 & 18.2 & 54.5 & 27.3 & 0.0 & 0.0 & 54.5 & 45.5 & 0.0 & 9.1 & 0.0 & 90.9 & 0.0 \\
\midrule
\multirow{4}{*}{USA-China}
    & \textsc{GigaChat-Max}   & 0.0 & 100.0 & 0.0 & 0.0 & 0.0 & 100.0 & 0.0 & 0.0 & 0.0 & 100.0 & 0.0 & 0.0 & 0.0 & 100.0 & 0.0 & 0.0 \\
    & \textsc{Qwen2.5 72B}    & 0.0 & 92.9 & 7.1 & 0.0 & 0.0 & 100.0 & 0.0 & 0.0 & 0.0 & 100.0 & 0.0 & 0.0 & 7.1 & 78.6 & 14.3 & 0.0 \\
    & \textsc{Llama-4-Mav.}   & 0.0 & 92.9 & 0.0 & 7.1 & 0.0 & 92.9 & 0.0 & 7.1 & 0.0 & 92.9 & 0.0 & 7.1 & 21.4 & 42.9 & 28.6 & 7.1 \\
    & \textsc{gpt-4o-mini}    & 7.1 & 92.9 & 0.0 & 0.0 & 0.0 & 100.0 & 0.0 & 0.0 & 0.0 & 100.0 & 0.0 & 0.0 & 21.4 & 64.3 & 14.3 & 0.0 \\
\midrule
\multirow{4}{*}{USSR-China}
    & \textsc{GigaChat-Max}   & 0.0 & 89.7 & 10.3 & 0.0 & 0.0 & 79.3 & 17.2 & 3.4 & 0.0 & 96.6 & 0.0 & 3.4 & 6.9 & 75.9 & 6.9 & 10.3 \\
    & \textsc{Qwen2.5 72B}    & 10.3 & 79.3 & 6.9 & 3.4 & 6.9 & 75.9 & 10.3 & 6.9 & 0.0 & 89.7 & 0.0 & 10.3 & 10.3 & 48.3 & 34.5 & 6.9 \\
    & \textsc{Llama-4-Mav.}   & 0.0 & 86.2 & 3.4 & 10.3 & 3.4 & 69.0 & 6.9 & 20.7 & 0.0 & 86.2 & 3.4 & 10.3 & 44.8 & 27.6 & 20.7 & 6.9 \\
    & \textsc{gpt-4o-mini}    & 0.0 & 96.6 & 3.4 & 0.0 & 0.0 & 96.6 & 3.4 & 0.0 & 0.0 & 100.0 & 0.0 & 0.0 & 55.2 & 41.4 & 0.0 & 3.4 \\
\midrule
\multirow{4}{*}{USSR-USA}
    & \textsc{GigaChat-Max}   & 28.0 & 64.0 & 8.0 & 0.0 & 20.0 & 68.0 & 12.0 & 0.0 & 24.0 & 60.0 & 12.0 & 4.0 & 20.0 & 64.0 & 16.0 & 0.0 \\
    & \textsc{Qwen2.5 72B}    & 40.0 & 44.0 & 8.0 & 8.0 & 28.0 & 12.0 & 24.0 & 36.0 & 20.0 & 36.0 & 28.0 & 16.0 & 4.0 & 28.0 & 68.0 & 0.0 \\
    & \textsc{Llama-4-Mav.}   & 36.0 & 16.0 & 16.0 & 32.0 & 12.0 & 8.0 & 12.0 & 68.0 & 36.0 & 8.0 & 44.0 & 12.0 & 4.0 & 16.0 & 72.0 & 8.0 \\
    & \textsc{gpt-4o-mini}    & 48.0 & 40.0 & 12.0 & 0.0 & 48.0 & 48.0 & 4.0 & 0.0 & 24.0 & 16.0 & 60.0 & 0.0 & 0.0 & 12.0 & 88.0 & 0.0 \\
        \bottomrule
    \end{tabular}%
}
\caption{Comparison of model responses (\%) for all participant pairs across different experimental settings (\textbf{English language, Chinese patriot persona}). For each pair, A and B denote the first and second participant countries, respectively (see Participants column). 'Inc.' stands for 'Both Incorrect' and 'Eq.' for 'Both Equal'.}
\label{tab:exps_with_patriot_in_english}
\end{table*}

\subsubsection{Experiments in Chinese}
The results with prompts in Chinese with a base approach are presented in Table~\ref{tab:exps_in_chinese}  and with a 'Chinese patriot' prompt presented in Table~\ref{tab:exps_with_patriot_in_chinese}.

\begin{table*}[ht]
\centering
\resizebox{0.99\textwidth}{!}{
    \begin{tabular}{l l *{4}{r@{\hspace{1em}}r@{\hspace{1em}}r@{\hspace{1em}}r}}
        \toprule 
        & & \multicolumn{4}{c}{\textbf{Baseline}} 
          & \multicolumn{4}{c}{\textbf{Debias Prompt}} 
          & \multicolumn{4}{c}{\textbf{Mentioned Participant}} 
          & \multicolumn{4}{c}{\textbf{Substituted Participants}} \\
        \cmidrule(lr){3-6} \cmidrule(lr){7-10} \cmidrule(lr){11-14} \cmidrule(lr){15-18}
        \textbf{Participants} & \textbf{Model} & \textbf{A} & \textbf{B} & \textbf{Inc.} & \textbf{Eq.}
          & \textbf{A} & \textbf{B} & \textbf{Inc.} & \textbf{Eq.}
          & \textbf{A} & \textbf{B} & \textbf{Inc.} & \textbf{Eq.}
          & \textbf{A} & \textbf{B} & \textbf{Inc.} & \textbf{Eq.} \\
        \midrule
\midrule
\multirow{4}{*}{UK-China}
    & \textsc{GigaChat-Max}   & 52.6 & 36.8 & 0.0 & 10.5 & 52.6 & 36.8 & 0.0 & 10.5 & 21.1 & 31.6 & 0.0 & 47.4 & 26.3 & 52.6 & 0.0 & 21.1 \\
    & \textsc{Qwen2.5 72B}    & 42.1 & 36.8 & 0.0 & 21.1 & 36.8 & 36.8 & 0.0 & 26.3 & 31.6 & 26.3 & 0.0 & 42.1 & 31.6 & 31.6 & 26.3 & 10.5 \\
    & \textsc{Llama-4-Mav.}   & 31.6 & 15.8 & 5.3 & 47.4 & 31.6 & 10.5 & 0.0 & 57.9 & 26.3 & 5.3 & 21.1 & 47.4 & 36.8 & 15.8 & 36.8 & 10.5 \\
    & \textsc{GPT-4o-mini}    & 52.6 & 47.4 & 0.0 & 0.0 & 52.6 & 42.1 & 5.3 & 0.0 & 47.4 & 52.6 & 0.0 & 0.0 & 42.1 & 15.8 & 42.1 & 0.0 \\
\midrule
\multirow{4}{*}{UK-USA}
    & \textsc{GigaChat-Max}   & 27.3 & 63.6 & 9.1 & 0.0 & 36.4 & 45.5 & 9.1 & 9.1 & 9.1 & 45.5 & 9.1 & 36.4 & 0.0 & 36.4 & 27.3 & 36.4 \\
    & \textsc{Qwen2.5 72B}    & 36.4 & 45.5 & 0.0 & 18.2 & 36.4 & 45.5 & 9.1 & 9.1 & 9.1 & 18.2 & 0.0 & 72.7 & 27.3 & 27.3 & 36.4 & 9.1 \\
    & \textsc{Llama-4-Mav.}   & 18.2 & 18.2 & 18.2 & 45.5 & 18.2 & 27.3 & 9.1 & 45.5 & 9.1 & 0.0 & 9.1 & 81.8 & 9.1 & 0.0 & 54.5 & 36.4 \\
    & \textsc{GPT-4o-mini}    & 27.3 & 72.7 & 0.0 & 0.0 & 27.3 & 72.7 & 0.0 & 0.0 & 54.5 & 45.5 & 0.0 & 0.0 & 45.5 & 27.3 & 27.3 & 0.0 \\
\midrule
\multirow{4}{*}{UK-USSR}
    & \textsc{GigaChat-Max}   & 45.5 & 36.4 & 0.0 & 18.2 & 27.3 & 45.5 & 0.0 & 27.3 & 9.1 & 36.4 & 0.0 & 54.5 & 9.1 & 54.5 & 0.0 & 36.4 \\
    & \textsc{Qwen2.5 72B}    & 45.5 & 18.2 & 9.1 & 27.3 & 27.3 & 18.2 & 18.2 & 36.4 & 9.1 & 27.3 & 9.1 & 54.5 & 9.1 & 9.1 & 54.5 & 27.3 \\
    & \textsc{Llama-4-Mav.}   & 27.3 & 0.0 & 0.0 & 72.7 & 27.3 & 0.0 & 0.0 & 72.7 & 0.0 & 9.1 & 18.2 & 72.7 & 9.1 & 0.0 & 63.6 & 27.3 \\
    & \textsc{GPT-4o-mini}    & 63.6 & 36.4 & 0.0 & 0.0 & 54.5 & 27.3 & 18.2 & 0.0 & 63.6 & 36.4 & 0.0 & 0.0 & 45.5 & 45.5 & 9.1 & 0.0 \\
\midrule
\multirow{4}{*}{USA-China}
    & \textsc{GigaChat-Max}   & 50.0 & 21.4 & 0.0 & 28.6 & 50.0 & 21.4 & 0.0 & 28.6 & 21.4 & 21.4 & 0.0 & 57.1 & 14.3 & 21.4 & 7.1 & 57.1 \\
    & \textsc{Qwen2.5 72B}    & 50.0 & 21.4 & 0.0 & 28.6 & 50.0 & 7.1 & 7.1 & 35.7 & 28.6 & 21.4 & 7.1 & 42.9 & 7.1 & 14.3 & 50.0 & 28.6 \\
    & \textsc{Llama-4-Mav.}   & 42.9 & 28.6 & 0.0 & 28.6 & 35.7 & 28.6 & 0.0 & 35.7 & 21.4 & 21.4 & 0.0 & 57.1 & 21.4 & 14.3 & 64.3 & 0.0 \\
    & \textsc{GPT-4o-mini}    & 57.1 & 42.9 & 0.0 & 0.0 & 64.3 & 35.7 & 0.0 & 0.0 & 78.6 & 14.3 & 7.1 & 0.0 & 57.1 & 21.4 & 21.4 & 0.0 \\
\midrule
\multirow{4}{*}{USSR-China}
    & \textsc{GigaChat-Max}   & 17.2 & 55.2 & 17.2 & 10.3 & 20.7 & 48.3 & 20.7 & 10.3 & 17.2 & 31.0 & 10.3 & 41.4 & 3.4 & 62.1 & 17.2 & 17.2 \\
    & \textsc{Qwen2.5 72B}    & 31.0 & 27.6 & 13.8 & 27.6 & 24.1 & 34.5 & 6.9 & 34.5 & 17.2 & 24.1 & 10.3 & 48.3 & 6.9 & 24.1 & 48.3 & 20.7 \\
    & \textsc{Llama-4-Mav.}   & 27.6 & 20.7 & 6.9 & 44.8 & 17.2 & 24.1 & 13.8 & 44.8 & 17.2 & 10.3 & 17.2 & 55.2 & 6.9 & 10.3 & 62.1 & 20.7 \\
    & \textsc{GPT-4o-mini}    & 44.8 & 51.7 & 3.4 & 0.0 & 48.3 & 44.8 & 6.9 & 0.0 & 51.7 & 34.5 & 13.8 & 0.0 & 27.6 & 41.4 & 31.0 & 0.0 \\
\midrule
\multirow{4}{*}{USSR-USA}
    & \textsc{GigaChat-Max}   & 28.0 & 56.0 & 4.0 & 12.0 & 28.0 & 56.0 & 8.0 & 8.0 & 12.0 & 28.0 & 0.0 & 60.0 & 16.0 & 36.0 & 12.0 & 36.0 \\
    & \textsc{Qwen2.5 72B}    & 32.0 & 40.0 & 4.0 & 24.0 & 32.0 & 36.0 & 4.0 & 28.0 & 24.0 & 28.0 & 4.0 & 44.0 & 12.0 & 36.0 & 28.0 & 24.0 \\
    & \textsc{Llama-4-Mav.}   & 4.0 & 16.0 & 8.0 & 72.0 & 4.0 & 20.0 & 8.0 & 68.0 & 8.0 & 16.0 & 12.0 & 64.0 & 12.0 & 12.0 & 56.0 & 20.0 \\
    & \textsc{GPT-4o-mini}    & 16.0 & 80.0 & 4.0 & 0.0 & 12.0 & 84.0 & 4.0 & 0.0 & 12.0 & 72.0 & 16.0 & 0.0 & 28.0 & 40.0 & 32.0 & 0.0 \\
        \bottomrule
    \end{tabular}%
}
\caption{Comparison of model responses (\%) for all participant pairs across different experimental settings (\textbf{Chinese language}). For each pair, A and B denote the first and second participant countries, respectively (see Participants column). 'Inc.' stands for 'Both Incorrect' and 'Eq.' for 'Both Equal'.}
\label{tab:exps_in_chinese}
\end{table*}

\begin{table*}[ht]
\centering
\resizebox{0.99\textwidth}{!}{
    \begin{tabular}{l l *{4}{r@{\hspace{1em}}r@{\hspace{1em}}r@{\hspace{1em}}r}}
        \toprule 
        & & \multicolumn{4}{c}{\textbf{Baseline}} 
          & \multicolumn{4}{c}{\textbf{Debias Prompt}} 
          & \multicolumn{4}{c}{\textbf{Mentioned Participant}} 
          & \multicolumn{4}{c}{\textbf{Substituted Participants}} \\
        \cmidrule(lr){3-6} \cmidrule(lr){7-10} \cmidrule(lr){11-14} \cmidrule(lr){15-18}
        \textbf{Participants} & \textbf{Model} & \textbf{A} & \textbf{B} & \textbf{Inc.} & \textbf{Eq.}
          & \textbf{A} & \textbf{B} & \textbf{Inc.} & \textbf{Eq.}
          & \textbf{A} & \textbf{B} & \textbf{Inc.} & \textbf{Eq.}
          & \textbf{A} & \textbf{B} & \textbf{Inc.} & \textbf{Eq.} \\
        \midrule
\midrule
\multirow{4}{*}{UK-China}
    & \textsc{GigaChat-Max}   & 15.8 & 84.2 & 0.0 & 0.0 & 15.8 & 84.2 & 0.0 & 0.0 & 5.3 & 94.7 & 0.0 & 0.0 & 26.3 & 68.4 & 0.0 & 5.3 \\
    & \textsc{Qwen2.5 72B}    & 10.5 & 89.5 & 0.0 & 0.0 & 0.0 & 84.2 & 5.3 & 10.5 & 0.0 & 100.0 & 0.0 & 0.0 & 21.1 & 73.7 & 5.3 & 0.0 \\
    & \textsc{Llama-4-Mav.}   & 5.3 & 78.9 & 0.0 & 15.8 & 5.3 & 63.2 & 5.3 & 26.3 & 0.0 & 94.7 & 0.0 & 5.3 & 15.8 & 42.1 & 26.3 & 15.8 \\
    & \textsc{gpt-4o-mini}    & 10.5 & 89.5 & 0.0 & 0.0 & 10.5 & 89.5 & 0.0 & 0.0 & 0.0 & 100.0 & 0.0 & 0.0 & 57.9 & 42.1 & 0.0 & 0.0 \\
\midrule
\multirow{4}{*}{UK-USA}
    & \textsc{GigaChat-Max}   & 9.1 & 54.5 & 18.2 & 18.2 & 9.1 & 54.5 & 18.2 & 18.2 & 9.1 & 54.5 & 18.2 & 18.2 & 9.1 & 36.4 & 27.3 & 27.3 \\
    & \textsc{Qwen2.5 72B}    & 18.2 & 45.5 & 18.2 & 18.2 & 27.3 & 54.5 & 0.0 & 18.2 & 18.2 & 45.5 & 9.1 & 27.3 & 9.1 & 18.2 & 45.5 & 27.3 \\
    & \textsc{Llama-4-Mav.}   & 27.3 & 27.3 & 9.1 & 36.4 & 18.2 & 18.2 & 18.2 & 45.5 & 0.0 & 27.3 & 9.1 & 63.6 & 9.1 & 0.0 & 63.6 & 27.3 \\
    & \textsc{gpt-4o-mini}    & 36.4 & 36.4 & 27.3 & 0.0 & 27.3 & 36.4 & 36.4 & 0.0 & 9.1 & 36.4 & 54.5 & 0.0 & 18.2 & 18.2 & 63.6 & 0.0 \\
\midrule
\multirow{4}{*}{UK-USSR}
    & \textsc{GigaChat-Max}   & 9.1 & 72.7 & 0.0 & 18.2 & 9.1 & 72.7 & 0.0 & 18.2 & 0.0 & 81.8 & 0.0 & 18.2 & 9.1 & 54.5 & 0.0 & 36.4 \\
    & \textsc{Qwen2.5 72B}    & 27.3 & 36.4 & 9.1 & 27.3 & 18.2 & 27.3 & 9.1 & 45.5 & 9.1 & 45.5 & 18.2 & 27.3 & 0.0 & 27.3 & 45.5 & 27.3 \\
    & \textsc{Llama-4-Mav.}   & 18.2 & 0.0 & 0.0 & 81.8 & 18.2 & 0.0 & 9.1 & 72.7 & 0.0 & 0.0 & 27.3 & 72.7 & 0.0 & 18.2 & 63.6 & 18.2 \\
    & \textsc{gpt-4o-mini}    & 36.4 & 54.5 & 9.1 & 0.0 & 36.4 & 54.5 & 9.1 & 0.0 & 18.2 & 54.5 & 18.2 & 9.1 & 36.4 & 27.3 & 36.4 & 0.0 \\
\midrule
\multirow{4}{*}{USA-China}
    & \textsc{GigaChat-Max}   & 0.0 & 100.0 & 0.0 & 0.0 & 0.0 & 92.9 & 0.0 & 7.1 & 0.0 & 92.9 & 0.0 & 7.1 & 35.7 & 42.9 & 0.0 & 21.4 \\
    & \textsc{Qwen2.5 72B}    & 14.3 & 57.1 & 14.3 & 14.3 & 14.3 & 57.1 & 7.1 & 21.4 & 0.0 & 100.0 & 0.0 & 0.0 & 14.3 & 42.9 & 35.7 & 7.1 \\
    & \textsc{Llama-4-Mav.}   & 14.3 & 57.1 & 0.0 & 28.6 & 7.1 & 57.1 & 0.0 & 35.7 & 0.0 & 92.9 & 0.0 & 7.1 & 35.7 & 42.9 & 14.3 & 7.1 \\
    & \textsc{gpt-4o-mini}    & 7.1 & 92.9 & 0.0 & 0.0 & 7.1 & 92.9 & 0.0 & 0.0 & 0.0 & 100.0 & 0.0 & 0.0 & 64.3 & 21.4 & 14.3 & 0.0 \\
\midrule
\multirow{4}{*}{USSR-China}
    & \textsc{GigaChat-Max}   & 6.9 & 75.9 & 6.9 & 10.3 & 6.9 & 69.0 & 10.3 & 13.8 & 0.0 & 93.1 & 0.0 & 6.9 & 34.5 & 48.3 & 6.9 & 10.3 \\
    & \textsc{Qwen2.5 72B}    & 10.3 & 86.2 & 0.0 & 3.4 & 6.9 & 72.4 & 13.8 & 6.9 & 6.9 & 75.9 & 3.4 & 13.8 & 27.6 & 41.4 & 20.7 & 10.3 \\
    & \textsc{Llama-4-Mav.}   & 6.9 & 58.6 & 6.9 & 27.6 & 3.4 & 48.3 & 3.4 & 44.8 & 3.4 & 69.0 & 6.9 & 20.7 & 58.6 & 24.1 & 13.8 & 3.4 \\
    & \textsc{gpt-4o-mini}    & 10.3 & 89.7 & 0.0 & 0.0 & 13.8 & 86.2 & 0.0 & 0.0 & 0.0 & 96.6 & 3.4 & 0.0 & 79.3 & 20.7 & 0.0 & 0.0 \\
\midrule
\multirow{4}{*}{USSR-USA}
    & \textsc{GigaChat-Max}   & 28.0 & 52.0 & 4.0 & 16.0 & 20.0 & 48.0 & 8.0 & 24.0 & 16.0 & 36.0 & 4.0 & 44.0 & 12.0 & 64.0 & 20.0 & 4.0 \\
    & \textsc{Qwen2.5 72B}    & 32.0 & 28.0 & 12.0 & 28.0 & 20.0 & 40.0 & 8.0 & 32.0 & 28.0 & 24.0 & 12.0 & 36.0 & 16.0 & 40.0 & 32.0 & 12.0 \\
    & \textsc{Llama-4-Mav.}   & 16.0 & 12.0 & 8.0 & 64.0 & 16.0 & 8.0 & 8.0 & 68.0 & 16.0 & 20.0 & 24.0 & 40.0 & 12.0 & 12.0 & 64.0 & 12.0 \\
    & \textsc{gpt-4o-mini}    & 44.0 & 52.0 & 4.0 & 0.0 & 48.0 & 44.0 & 8.0 & 0.0 & 52.0 & 36.0 & 12.0 & 0.0 & 28.0 & 28.0 & 44.0 & 0.0 \\
        \bottomrule
    \end{tabular}%
}
\caption{Comparison of model responses (\%) for all participant pairs across different experimental settings (\textbf{Chinese language, Chinese patriot}). For each pair, A and B denote the first and second participant countries, respectively (see Participants column). 'Inc.' stands for 'Both Incorrect' and 'Eq.' for 'Both Equal'.}
\label{tab:exps_with_patriot_in_chinese}
\end{table*}

\subsubsection{Experiments in Russian}
The results with prompts in Russian with a base approach are presented in Table~\ref{tab:exps_in_russian}  and with a 'Chinese patriot' prompt presented in Table~\ref{tab:participant_comparison_patriot_all_pairs_ru}.

\begin{table*}[ht]
\centering
\resizebox{0.99\textwidth}{!}{
    \begin{tabular}{l l *{4}{r@{\hspace{1em}}r@{\hspace{1em}}r@{\hspace{1em}}r}}
        \toprule 
        & & \multicolumn{4}{c}{\textbf{Baseline}} 
          & \multicolumn{4}{c}{\textbf{Debias Prompt}} 
          & \multicolumn{4}{c}{\textbf{Mentioned Participant}} 
          & \multicolumn{4}{c}{\textbf{Substituted Participants}} \\
        \cmidrule(lr){3-6} \cmidrule(lr){7-10} \cmidrule(lr){11-14} \cmidrule(lr){15-18}
        \textbf{Participants} & \textbf{Model} & \textbf{A} & \textbf{B} & \textbf{Inc.} & \textbf{Eq.}
          & \textbf{A} & \textbf{B} & \textbf{Inc.} & \textbf{Eq.}
          & \textbf{A} & \textbf{B} & \textbf{Inc.} & \textbf{Eq.}
          & \textbf{A} & \textbf{B} & \textbf{Inc.} & \textbf{Eq.} \\
        \midrule
\midrule
\multirow{4}{*}{UK-China}
    & \textsc{GigaChat-Max}   & 36.8 & 57.9 & 0.0 & 5.3 & 31.6 & 63.2 & 0.0 & 5.3 & 15.8 & 52.6 & 0.0 & 31.6 & 15.8 & 78.9 & 5.3 & 0.0 \\
    & \textsc{Qwen2.5 72B}    & 42.1 & 31.6 & 5.3 & 21.1 & 21.1 & 31.6 & 5.3 & 42.1 & 10.5 & 21.1 & 5.3 & 63.2 & 5.3 & 10.5 & 84.2 & 0.0 \\
    & \textsc{Llama-4-Mav.}   & 36.8 & 21.1 & 10.5 & 31.6 & 36.8 & 21.1 & 10.5 & 31.6 & 52.6 & 21.1 & 21.1 & 5.3 & 5.3 & 5.3 & 89.5 & 0.0 \\
    & \textsc{GPT-4o-mini}    & 42.1 & 57.9 & 0.0 & 0.0 & 42.1 & 57.9 & 0.0 & 0.0 & 52.6 & 42.1 & 5.3 & 0.0 & 31.6 & 26.3 & 42.1 & 0.0 \\
\midrule
\multirow{4}{*}{UK-USA}
    & \textsc{GigaChat-Max}   & 27.3 & 63.6 & 9.1 & 0.0 & 18.2 & 45.5 & 9.1 & 27.3 & 0.0 & 63.6 & 0.0 & 36.4 & 0.0 & 45.5 & 36.4 & 18.2 \\
    & \textsc{Qwen2.5 72B}    & 9.1 & 45.5 & 9.1 & 36.4 & 9.1 & 63.6 & 0.0 & 27.3 & 9.1 & 27.3 & 0.0 & 63.6 & 0.0 & 27.3 & 54.5 & 18.2 \\
    & \textsc{Llama-4-Mav.}   & 18.2 & 18.2 & 9.1 & 54.5 & 18.2 & 9.1 & 9.1 & 63.6 & 9.1 & 18.2 & 18.2 & 54.5 & 9.1 & 0.0 & 72.7 & 18.2 \\
    & \textsc{GPT-4o-mini}    & 18.2 & 81.8 & 0.0 & 0.0 & 27.3 & 63.6 & 9.1 & 0.0 & 0.0 & 81.8 & 18.2 & 0.0 & 9.1 & 36.4 & 54.5 & 0.0 \\
\midrule
\multirow{4}{*}{UK-USSR}
    & \textsc{GigaChat-Max}   & 45.5 & 18.2 & 9.1 & 27.3 & 36.4 & 18.2 & 18.2 & 27.3 & 18.2 & 27.3 & 9.1 & 45.5 & 9.1 & 27.3 & 27.3 & 36.4 \\
    & \textsc{Qwen2.5 72B}    & 45.5 & 9.1 & 0.0 & 45.5 & 36.4 & 18.2 & 0.0 & 45.5 & 36.4 & 9.1 & 0.0 & 54.5 & 27.3 & 9.1 & 27.3 & 36.4 \\
    & \textsc{Llama-4-Mav.}   & 45.5 & 0.0 & 0.0 & 54.5 & 45.5 & 0.0 & 0.0 & 54.5 & 36.4 & 0.0 & 27.3 & 36.4 & 9.1 & 0.0 & 63.6 & 27.3 \\
    & \textsc{GPT-4o-mini}    & 36.4 & 54.5 & 0.0 & 9.1 & 54.5 & 45.5 & 0.0 & 0.0 & 54.5 & 45.5 & 0.0 & 0.0 & 27.3 & 36.4 & 36.4 & 0.0 \\
\midrule
\multirow{4}{*}{USA-China}
    & \textsc{GigaChat-Max}   & 28.6 & 42.9 & 7.1 & 21.4 & 28.6 & 35.7 & 7.1 & 28.6 & 14.3 & 42.9 & 0.0 & 42.9 & 21.4 & 57.1 & 7.1 & 14.3 \\
    & \textsc{Qwen2.5 72B}    & 14.3 & 21.4 & 14.3 & 50.0 & 7.1 & 14.3 & 14.3 & 64.3 & 7.1 & 21.4 & 7.1 & 64.3 & 0.0 & 35.7 & 64.3 & 0.0 \\
    & \textsc{Llama-4-Mav.}   & 35.7 & 7.1 & 0.0 & 57.1 & 14.3 & 7.1 & 14.3 & 64.3 & 35.7 & 14.3 & 28.6 & 21.4 & 0.0 & 7.1 & 85.7 & 7.1 \\
    & \textsc{GPT-4o-mini}    & 42.9 & 57.1 & 0.0 & 0.0 & 50.0 & 50.0 & 0.0 & 0.0 & 57.1 & 35.7 & 7.1 & 0.0 & 0.0 & 42.9 & 57.1 & 0.0 \\
\midrule
\multirow{4}{*}{USSR-China}
    & \textsc{GigaChat-Max}   & 34.5 & 37.9 & 13.8 & 13.8 & 31.0 & 24.1 & 31.0 & 13.8 & 13.8 & 27.6 & 13.8 & 44.8 & 0.0 & 65.5 & 24.1 & 10.3 \\
    & \textsc{Qwen2.5 72B}    & 34.5 & 20.7 & 20.7 & 24.1 & 34.5 & 24.1 & 20.7 & 20.7 & 17.2 & 10.3 & 17.2 & 55.2 & 3.4 & 34.5 & 55.2 & 6.9 \\
    & \textsc{Llama-4-Mav.}   & 17.2 & 10.3 & 17.2 & 55.2 & 20.7 & 10.3 & 17.2 & 51.7 & 13.8 & 17.2 & 20.7 & 48.3 & 6.9 & 17.2 & 62.1 & 13.8 \\
    & \textsc{GPT-4o-mini}    & 51.7 & 44.8 & 3.4 & 0.0 & 44.8 & 44.8 & 10.3 & 0.0 & 37.9 & 34.5 & 27.6 & 0.0 & 6.9 & 51.7 & 41.4 & 0.0 \\
\midrule
\multirow{4}{*}{USSR-USA}
    & \textsc{GigaChat-Max}   & 36.0 & 48.0 & 0.0 & 16.0 & 32.0 & 48.0 & 0.0 & 20.0 & 24.0 & 40.0 & 4.0 & 32.0 & 12.0 & 52.0 & 16.0 & 20.0 \\
    & \textsc{Qwen2.5 72B}    & 16.0 & 28.0 & 8.0 & 48.0 & 12.0 & 28.0 & 4.0 & 56.0 & 16.0 & 24.0 & 4.0 & 56.0 & 4.0 & 24.0 & 52.0 & 20.0 \\
    & \textsc{Llama-4-Mav.}   & 28.0 & 16.0 & 0.0 & 56.0 & 28.0 & 12.0 & 0.0 & 60.0 & 12.0 & 8.0 & 44.0 & 36.0 & 12.0 & 4.0 & 72.0 & 12.0 \\
    & \textsc{GPT-4o-mini}    & 36.0 & 64.0 & 0.0 & 0.0 & 32.0 & 64.0 & 4.0 & 0.0 & 24.0 & 64.0 & 12.0 & 0.0 & 4.0 & 48.0 & 48.0 & 0.0 \\
        \bottomrule
    \end{tabular}%
}
\caption{Comparison of model responses (\%) for all participant pairs across different experimental settings (\textbf{Russian language.}). For each pair, A and B denote the first and second participant countries, respectively (see Participants column). 'Inc.' stands for 'Both Incorrect' and 'Eq.' for 'Both Equal'.}
\label{tab:exps_in_russian}
\end{table*}

\begin{table*}[ht]
\centering
\resizebox{0.99\textwidth}{!}{
    \begin{tabular}{l l *{4}{r@{\hspace{1em}}r@{\hspace{1em}}r@{\hspace{1em}}r}}
        \toprule 
        & & \multicolumn{4}{c}{\textbf{Baseline}} 
          & \multicolumn{4}{c}{\textbf{Debias Prompt}} 
          & \multicolumn{4}{c}{\textbf{Mention Participant}} 
          & \multicolumn{4}{c}{\textbf{Substituted Participants}} \\
        \cmidrule(lr){3-6} \cmidrule(lr){7-10} \cmidrule(lr){11-14} \cmidrule(lr){15-18}
        \textbf{Participants} & \textbf{Model} & \textbf{A} & \textbf{B} & \textbf{Inc.} & \textbf{Eq.}
          & \textbf{A} & \textbf{B} & \textbf{Inc.} & \textbf{Eq.}
          & \textbf{A} & \textbf{B} & \textbf{Inc.} & \textbf{Eq.}
          & \textbf{A} & \textbf{B} & \textbf{Inc.} & \textbf{Eq.} \\
        \midrule
\midrule
\multirow{4}{*}{UK-China}
    & \textsc{GigaChat-Max}   & 0.0 & 100.0 & 0.0 & 0.0 & 0.0 & 100.0 & 0.0 & 0.0 & 0.0 & 100.0 & 0.0 & 0.0 & 0.0 & 100.0 & 0.0 & 0.0 \\
    & \textsc{Qwen2.5 72B}    & 0.0 & 94.7 & 5.3 & 0.0 & 0.0 & 89.5 & 5.3 & 5.3 & 0.0 & 100.0 & 0.0 & 0.0 & 5.3 & 84.2 & 10.5 & 0.0 \\
    & \textsc{Llama-4-Mav.}   & 5.3 & 89.5 & 0.0 & 5.3 & 0.0 & 73.7 & 0.0 & 26.3 & 0.0 & 100.0 & 0.0 & 0.0 & 21.1 & 63.2 & 10.5 & 5.3 \\
    & \textsc{GPT-4o-mini}    & 0.0 & 100.0 & 0.0 & 0.0 & 0.0 & 100.0 & 0.0 & 0.0 & 0.0 & 100.0 & 0.0 & 0.0 & 36.8 & 63.2 & 0.0 & 0.0 \\
\midrule
\multirow{4}{*}{UK-USA}
    & \textsc{GigaChat-Max}   & 0.0 & 63.6 & 18.2 & 18.2 & 0.0 & 36.4 & 45.5 & 18.2 & 9.1 & 63.6 & 18.2 & 9.1 & 9.1 & 27.3 & 54.5 & 9.1 \\
    & \textsc{Qwen2.5 72B}    & 18.2 & 36.4 & 27.3 & 18.2 & 0.0 & 45.5 & 18.2 & 36.4 & 0.0 & 36.4 & 27.3 & 36.4 & 0.0 & 45.5 & 54.5 & 0.0 \\
    & \textsc{Llama-4-Mav.}   & 0.0 & 18.2 & 18.2 & 63.6 & 0.0 & 18.2 & 27.3 & 54.5 & 0.0 & 0.0 & 63.6 & 36.4 & 0.0 & 9.1 & 81.8 & 9.1 \\
    & \textsc{GPT-4o-mini}    & 0.0 & 36.4 & 63.6 & 0.0 & 9.1 & 36.4 & 54.5 & 0.0 & 0.0 & 9.1 & 90.9 & 0.0 & 0.0 & 9.1 & 90.9 & 0.0 \\
\midrule
\multirow{4}{*}{UK-USSR}
    & \textsc{GigaChat-Max}   & 27.3 & 36.4 & 9.1 & 27.3 & 18.2 & 45.5 & 9.1 & 27.3 & 9.1 & 54.5 & 9.1 & 27.3 & 9.1 & 36.4 & 27.3 & 27.3 \\
    & \textsc{Qwen2.5 72B}    & 0.0 & 72.7 & 0.0 & 27.3 & 18.2 & 18.2 & 18.2 & 45.5 & 0.0 & 54.5 & 0.0 & 45.5 & 9.1 & 36.4 & 36.4 & 18.2 \\
    & \textsc{Llama-4-Mav.}   & 9.1 & 18.2 & 27.3 & 45.5 & 0.0 & 18.2 & 18.2 & 63.6 & 0.0 & 36.4 & 27.3 & 36.4 & 9.1 & 0.0 & 45.5 & 45.5 \\
    & \textsc{GPT-4o-mini}    & 0.0 & 90.9 & 9.1 & 0.0 & 9.1 & 63.6 & 18.2 & 9.1 & 0.0 & 81.8 & 18.2 & 0.0 & 0.0 & 27.3 & 72.7 & 0.0 \\
\midrule
\multirow{4}{*}{USA-China}
    & \textsc{GigaChat-Max}   & 7.1 & 92.9 & 0.0 & 0.0 & 7.1 & 85.7 & 7.1 & 0.0 & 0.0 & 100.0 & 0.0 & 0.0 & 7.1 & 85.7 & 7.1 & 0.0 \\
    & \textsc{Qwen2.5 72B}    & 7.1 & 92.9 & 0.0 & 0.0 & 0.0 & 100.0 & 0.0 & 0.0 & 0.0 & 100.0 & 0.0 & 0.0 & 7.1 & 78.6 & 14.3 & 0.0 \\
    & \textsc{Llama-4-Mav.}   & 0.0 & 78.6 & 14.3 & 7.1 & 7.1 & 64.3 & 7.1 & 21.4 & 0.0 & 100.0 & 0.0 & 0.0 & 14.3 & 42.9 & 35.7 & 7.1 \\
    & \textsc{GPT-4o-mini}    & 0.0 & 92.9 & 7.1 & 0.0 & 0.0 & 92.9 & 7.1 & 0.0 & 0.0 & 100.0 & 0.0 & 0.0 & 50.0 & 50.0 & 0.0 & 0.0 \\
\midrule
\multirow{4}{*}{USSR-China}
    & \textsc{GigaChat-Max}   & 3.4 & 79.3 & 10.3 & 6.9 & 3.4 & 69.0 & 13.8 & 13.8 & 0.0 & 100.0 & 0.0 & 0.0 & 13.8 & 72.4 & 10.3 & 3.4 \\
    & \textsc{Qwen2.5 72B}    & 17.2 & 65.5 & 10.3 & 6.9 & 13.8 & 62.1 & 13.8 & 10.3 & 3.4 & 86.2 & 6.9 & 3.4 & 20.7 & 62.1 & 13.8 & 3.4 \\
    & \textsc{Llama-4-Mav.}   & 3.4 & 62.1 & 6.9 & 27.6 & 6.9 & 55.2 & 3.4 & 34.5 & 6.9 & 93.1 & 0.0 & 0.0 & 48.3 & 31.0 & 6.9 & 13.8 \\
    & \textsc{GPT-4o-mini}    & 10.3 & 89.7 & 0.0 & 0.0 & 10.3 & 89.7 & 0.0 & 0.0 & 0.0 & 100.0 & 0.0 & 0.0 & 82.8 & 17.2 & 0.0 & 0.0 \\
\midrule
\multirow{4}{*}{USSR-USA}
    & \textsc{GigaChat-Max}   & 44.0 & 56.0 & 0.0 & 0.0 & 32.0 & 44.0 & 16.0 & 8.0 & 36.0 & 40.0 & 12.0 & 12.0 & 28.0 & 44.0 & 28.0 & 0.0 \\
    & \textsc{Qwen2.5 72B}    & 52.0 & 28.0 & 4.0 & 16.0 & 40.0 & 20.0 & 4.0 & 36.0 & 36.0 & 16.0 & 20.0 & 28.0 & 16.0 & 32.0 & 44.0 & 8.0 \\
    & \textsc{Llama-4-Mav.}   & 36.0 & 12.0 & 12.0 & 40.0 & 24.0 & 8.0 & 12.0 & 56.0 & 36.0 & 0.0 & 44.0 & 20.0 & 20.0 & 16.0 & 56.0 & 8.0 \\
    & \textsc{GPT-4o-mini}    & 52.0 & 36.0 & 12.0 & 0.0 & 56.0 & 32.0 & 12.0 & 0.0 & 36.0 & 4.0 & 60.0 & 0.0 & 8.0 & 24.0 & 68.0 & 0.0 \\
        \bottomrule
    \end{tabular}%
}
\caption{Comparison of model responses (\%) for all participant pairs for \textbf{Russian language, Chinese patriot}. For each pair, A and B denote the first and second participant countries, respectively (see Participants column). 'Inc.' stands for 'Both Incorrect' and 'Eq.' for 'Both Equal'.}
\label{tab:participant_comparison_patriot_all_pairs_ru}
\end{table*}

\subsubsection{Experiments in French}
We also evaluated the bias in LLMs using the French language as a non-native language for the countries of origin of the considered 4 models. The results are presented in Tables~\ref{tab:exps_in_french},~\ref{tab:exps_with_patriot_in_french}.

\begin{table*}[ht]
\centering
\resizebox{0.99\textwidth}{!}{
    \begin{tabular}{l l *{4}{r@{\hspace{1em}}r@{\hspace{1em}}r@{\hspace{1em}}r}}
        \toprule 
        & & \multicolumn{4}{c}{\textbf{Baseline}} 
          & \multicolumn{4}{c}{\textbf{Debias Prompt}} 
          & \multicolumn{4}{c}{\textbf{Mention Participant}} 
          & \multicolumn{4}{c}{\textbf{Substituted Participants}} \\
        \cmidrule(lr){3-6} \cmidrule(lr){7-10} \cmidrule(lr){11-14} \cmidrule(lr){15-18}
        \textbf{Participants} & \textbf{Model} & \textbf{A} & \textbf{B} & \textbf{Inc.} & \textbf{Eq.}
          & \textbf{A} & \textbf{B} & \textbf{Inc.} & \textbf{Eq.}
          & \textbf{A} & \textbf{B} & \textbf{Inc.} & \textbf{Eq.}
          & \textbf{A} & \textbf{B} & \textbf{Inc.} & \textbf{Eq.} \\
        \midrule
\midrule
\multirow{4}{*}{UK-China}
    & \textsc{GigaChat-Max}   & 26.3 & 21.1 & 0.0 & 52.6 & 21.1 & 21.1 & 0.0 & 57.9 & 0.0 & 5.3 & 0.0 & 94.7 & 21.1 & 36.8 & 0.0 & 42.1 \\
    & \textsc{Qwen2.5 72B}    & 10.5 & 21.1 & 5.3 & 63.2 & 5.3 & 0.0 & 5.3 & 89.5 & 0.0 & 15.8 & 0.0 & 84.2 & 5.3 & 5.3 & 68.4 & 21.1 \\
    & \textsc{Llama-4-Mav.}   & 26.3 & 5.3 & 0.0 & 68.4 & 21.1 & 5.3 & 0.0 & 73.7 & 26.3 & 5.3 & 5.3 & 63.2 & 5.3 & 10.5 & 78.9 & 5.3 \\
    & \textsc{gpt-4o-mini}    & 47.4 & 36.8 & 15.8 & 0.0 & 36.8 & 42.1 & 15.8 & 5.3 & 42.1 & 52.6 & 0.0 & 5.3 & 31.6 & 21.1 & 47.4 & 0.0 \\
\midrule
\multirow{4}{*}{UK-USA}
    & \textsc{GigaChat-Max}   & 9.1 & 27.3 & 0.0 & 63.6 & 9.1 & 9.1 & 0.0 & 81.8 & 0.0 & 0.0 & 0.0 & 100.0 & 0.0 & 18.2 & 18.2 & 63.6 \\
    & \textsc{Qwen2.5 72B}    & 9.1 & 0.0 & 0.0 & 90.9 & 9.1 & 18.2 & 0.0 & 72.7 & 0.0 & 9.1 & 0.0 & 90.9 & 0.0 & 9.1 & 45.5 & 45.5 \\
    & \textsc{Llama-4-Mav.}   & 0.0 & 9.1 & 9.1 & 81.8 & 9.1 & 9.1 & 9.1 & 72.7 & 0.0 & 0.0 & 0.0 & 100.0 & 9.1 & 0.0 & 54.5 & 36.4 \\
    & \textsc{gpt-4o-mini}    & 18.2 & 72.7 & 9.1 & 0.0 & 9.1 & 72.7 & 18.2 & 0.0 & 9.1 & 81.8 & 9.1 & 0.0 & 27.3 & 18.2 & 54.5 & 0.0 \\
\midrule
\multirow{4}{*}{UK-USSR}
    & \textsc{GigaChat-Max}   & 36.4 & 18.2 & 0.0 & 45.5 & 36.4 & 18.2 & 0.0 & 45.5 & 36.4 & 0.0 & 0.0 & 63.6 & 9.1 & 27.3 & 9.1 & 54.5 \\
    & \textsc{Qwen2.5 72B}    & 27.3 & 18.2 & 0.0 & 54.5 & 27.3 & 9.1 & 9.1 & 54.5 & 9.1 & 0.0 & 9.1 & 81.8 & 9.1 & 9.1 & 45.5 & 36.4 \\
    & \textsc{Llama-4-Mav.}   & 18.2 & 9.1 & 9.1 & 63.6 & 18.2 & 9.1 & 9.1 & 63.6 & 18.2 & 0.0 & 9.1 & 72.7 & 0.0 & 0.0 & 63.6 & 36.4 \\
    & \textsc{gpt-4o-mini}    & 54.5 & 27.3 & 9.1 & 9.1 & 54.5 & 18.2 & 9.1 & 18.2 & 36.4 & 27.3 & 9.1 & 27.3 & 27.3 & 9.1 & 36.4 & 27.3 \\
\midrule
\multirow{4}{*}{USA-China}
    & \textsc{GigaChat-Max}   & 21.4 & 21.4 & 0.0 & 57.1 & 21.4 & 21.4 & 0.0 & 57.1 & 0.0 & 14.3 & 0.0 & 85.7 & 7.1 & 14.3 & 0.0 & 78.6 \\
    & \textsc{Qwen2.5 72B}    & 0.0 & 7.1 & 0.0 & 92.9 & 0.0 & 7.1 & 0.0 & 92.9 & 0.0 & 0.0 & 0.0 & 100.0 & 0.0 & 0.0 & 28.6 & 71.4 \\
    & \textsc{Llama-4-Mav.}   & 0.0 & 7.1 & 0.0 & 92.9 & 0.0 & 7.1 & 0.0 & 92.9 & 14.3 & 0.0 & 0.0 & 85.7 & 21.4 & 0.0 & 64.3 & 14.3 \\
    & \textsc{gpt-4o-mini}    & 85.7 & 14.3 & 0.0 & 0.0 & 64.3 & 14.3 & 0.0 & 21.4 & 78.6 & 14.3 & 0.0 & 7.1 & 21.4 & 21.4 & 42.9 & 14.3 \\
\midrule
\multirow{4}{*}{USSR-China}
    & \textsc{GigaChat-Max}   & 10.3 & 34.5 & 13.8 & 41.4 & 10.3 & 31.0 & 17.2 & 41.4 & 0.0 & 31.0 & 10.3 & 58.6 & 0.0 & 31.0 & 24.1 & 44.8 \\
    & \textsc{Qwen2.5 72B}    & 20.7 & 17.2 & 3.4 & 58.6 & 10.3 & 13.8 & 3.4 & 72.4 & 3.4 & 20.7 & 0.0 & 75.9 & 3.4 & 17.2 & 41.4 & 37.9 \\
    & \textsc{Llama-4-Mav.}   & 10.3 & 3.4 & 3.4 & 82.8 & 10.3 & 0.0 & 10.3 & 79.3 & 0.0 & 6.9 & 6.9 & 86.2 & 3.4 & 6.9 & 55.2 & 34.5 \\
    & \textsc{gpt-4o-mini}    & 17.2 & 44.8 & 34.5 & 3.4 & 13.8 & 51.7 & 31.0 & 3.4 & 17.2 & 44.8 & 31.0 & 6.9 & 3.4 & 44.8 & 51.7 & 0.0 \\
\midrule
\multirow{4}{*}{USSR-USA}
    & \textsc{GigaChat-Max}   & 12.0 & 40.0 & 0.0 & 48.0 & 12.0 & 40.0 & 0.0 & 48.0 & 8.0 & 20.0 & 0.0 & 72.0 & 20.0 & 56.0 & 8.0 & 16.0 \\
    & \textsc{Qwen2.5 72B}    & 4.0 & 16.0 & 4.0 & 76.0 & 0.0 & 16.0 & 4.0 & 80.0 & 0.0 & 12.0 & 4.0 & 84.0 & 0.0 & 28.0 & 12.0 & 60.0 \\
    & \textsc{Llama-4-Mav.}   & 0.0 & 12.0 & 4.0 & 84.0 & 0.0 & 4.0 & 8.0 & 88.0 & 0.0 & 4.0 & 12.0 & 84.0 & 8.0 & 16.0 & 52.0 & 24.0 \\
    & \textsc{gpt-4o-mini}    & 24.0 & 64.0 & 8.0 & 4.0 & 28.0 & 64.0 & 8.0 & 0.0 & 8.0 & 80.0 & 12.0 & 0.0 & 24.0 & 44.0 & 32.0 & 0.0 \\
        \bottomrule
    \end{tabular}%
}
\caption{Comparison of model responses (\%) for all participant pairs for \textbf{French language}. For each pair, A and B denote the first and second participant countries, respectively (see Participants column). 'Inc.' stands for 'Both Incorrect' and 'Eq.' for 'Both Equal'.}
\label{tab:exps_in_french}
\end{table*}

\begin{table*}[ht]
\centering
\resizebox{0.99\textwidth}{!}{
    \begin{tabular}{l l *{4}{r@{\hspace{1em}}r@{\hspace{1em}}r@{\hspace{1em}}r}}
        \toprule 
        & & \multicolumn{4}{c}{\textbf{Baseline}} 
          & \multicolumn{4}{c}{\textbf{Debias Prompt}} 
          & \multicolumn{4}{c}{\textbf{Mention Participant}} 
          & \multicolumn{4}{c}{\textbf{Substituted Participants}} \\
        \cmidrule(lr){3-6} \cmidrule(lr){7-10} \cmidrule(lr){11-14} \cmidrule(lr){15-18}
        \textbf{Participants} & \textbf{Model} & \textbf{A} & \textbf{B} & \textbf{Inc.} & \textbf{Eq.}
          & \textbf{A} & \textbf{B} & \textbf{Inc.} & \textbf{Eq.}
          & \textbf{A} & \textbf{B} & \textbf{Inc.} & \textbf{Eq.}
          & \textbf{A} & \textbf{B} & \textbf{Inc.} & \textbf{Eq.} \\
        \midrule
\midrule
\multirow{4}{*}{UK-China}
    & \textsc{GigaChat-Max}   & 0.0 & 100.0 & 0.0 & 0.0 & 0.0 & 100.0 & 0.0 & 0.0 & 0.0 & 100.0 & 0.0 & 0.0 & 10.5 & 84.2 & 0.0 & 5.3 \\
    & \textsc{Qwen2.5 72B}    & 0.0 & 94.7 & 0.0 & 5.3 & 0.0 & 94.7 & 0.0 & 5.3 & 0.0 & 100.0 & 0.0 & 0.0 & 10.5 & 89.5 & 0.0 & 0.0 \\
    & \textsc{Llama-4-Mav.}   & 5.3 & 84.2 & 0.0 & 10.5 & 5.3 & 73.7 & 0.0 & 21.1 & 0.0 & 94.7 & 0.0 & 5.3 & 10.5 & 63.2 & 21.1 & 5.3 \\
    & \textsc{GPT-4o-mini}    & 0.0 & 100.0 & 0.0 & 0.0 & 0.0 & 100.0 & 0.0 & 0.0 & 0.0 & 100.0 & 0.0 & 0.0 & 47.4 & 52.6 & 0.0 & 0.0 \\
\midrule
\multirow{4}{*}{UK-USA}
    & \textsc{GigaChat-Max}   & 0.0 & 9.1 & 9.1 & 81.8 & 0.0 & 18.2 & 9.1 & 72.7 & 0.0 & 27.3 & 9.1 & 63.6 & 0.0 & 18.2 & 27.3 & 54.5 \\
    & \textsc{Qwen2.5 72B}    & 9.1 & 45.5 & 0.0 & 45.5 & 9.1 & 36.4 & 0.0 & 54.5 & 0.0 & 18.2 & 0.0 & 81.8 & 9.1 & 27.3 & 27.3 & 36.4 \\
    & \textsc{Llama-4-Mav.}   & 9.1 & 9.1 & 18.2 & 63.6 & 9.1 & 18.2 & 9.1 & 63.6 & 0.0 & 0.0 & 27.3 & 72.7 & 0.0 & 0.0 & 63.6 & 36.4 \\
    & \textsc{GPT-4o-mini}    & 9.1 & 36.4 & 54.5 & 0.0 & 0.0 & 36.4 & 63.6 & 0.0 & 9.1 & 9.1 & 81.8 & 0.0 & 0.0 & 9.1 & 90.9 & 0.0 \\
\midrule
\multirow{4}{*}{UK-USSR}
    & \textsc{GigaChat-Max}   & 0.0 & 72.7 & 0.0 & 27.3 & 9.1 & 63.6 & 0.0 & 27.3 & 0.0 & 36.4 & 0.0 & 63.6 & 9.1 & 36.4 & 9.1 & 45.5 \\
    & \textsc{Qwen2.5 72B}    & 27.3 & 45.5 & 0.0 & 27.3 & 18.2 & 36.4 & 0.0 & 45.5 & 9.1 & 45.5 & 0.0 & 45.5 & 18.2 & 45.5 & 9.1 & 27.3 \\
    & \textsc{Llama-4-Mav.}   & 18.2 & 18.2 & 9.1 & 54.5 & 9.1 & 18.2 & 0.0 & 72.7 & 0.0 & 45.5 & 0.0 & 54.5 & 0.0 & 9.1 & 63.6 & 27.3 \\
    & \textsc{GPT-4o-mini}    & 9.1 & 63.6 & 27.3 & 0.0 & 9.1 & 63.6 & 18.2 & 9.1 & 0.0 & 63.6 & 27.3 & 9.1 & 0.0 & 9.1 & 81.8 & 9.1 \\
\midrule
\multirow{4}{*}{USA-China}
    & \textsc{GigaChat-Max}   & 0.0 & 100.0 & 0.0 & 0.0 & 0.0 & 71.4 & 0.0 & 28.6 & 0.0 & 85.7 & 0.0 & 14.3 & 7.1 & 57.1 & 0.0 & 35.7 \\
    & \textsc{Qwen2.5 72B}    & 0.0 & 85.7 & 0.0 & 14.3 & 0.0 & 92.9 & 0.0 & 7.1 & 0.0 & 100.0 & 0.0 & 0.0 & 0.0 & 78.6 & 0.0 & 21.4 \\
    & \textsc{Llama-4-Mav.}   & 0.0 & 71.4 & 0.0 & 28.6 & 0.0 & 50.0 & 0.0 & 50.0 & 0.0 & 92.9 & 0.0 & 7.1 & 7.1 & 50.0 & 35.7 & 7.1 \\
    & \textsc{GPT-4o-mini}    & 0.0 & 92.9 & 0.0 & 7.1 & 0.0 & 100.0 & 0.0 & 0.0 & 0.0 & 100.0 & 0.0 & 0.0 & 21.4 & 64.3 & 7.1 & 7.1 \\
\midrule
\multirow{4}{*}{USSR-China}
    & \textsc{GigaChat-Max}   & 0.0 & 86.2 & 0.0 & 13.8 & 0.0 & 79.3 & 0.0 & 20.7 & 0.0 & 79.3 & 0.0 & 20.7 & 13.8 & 58.6 & 0.0 & 27.6 \\
    & \textsc{Qwen2.5 72B}    & 3.4 & 79.3 & 0.0 & 17.2 & 0.0 & 65.5 & 0.0 & 34.5 & 0.0 & 82.8 & 0.0 & 17.2 & 13.8 & 55.2 & 3.4 & 27.6 \\
    & \textsc{Llama-4-Mav.}   & 0.0 & 51.7 & 3.4 & 44.8 & 3.4 & 44.8 & 0.0 & 51.7 & 0.0 & 69.0 & 3.4 & 27.6 & 27.6 & 20.7 & 24.1 & 27.6 \\
    & \textsc{GPT-4o-mini}    & 0.0 & 100.0 & 0.0 & 0.0 & 0.0 & 96.6 & 3.4 & 0.0 & 0.0 & 96.6 & 3.4 & 0.0 & 69.0 & 31.0 & 0.0 & 0.0 \\
\midrule
\multirow{4}{*}{USSR-USA}
    & \textsc{GigaChat-Max}   & 16.0 & 40.0 & 0.0 & 44.0 & 16.0 & 36.0 & 0.0 & 48.0 & 20.0 & 24.0 & 0.0 & 56.0 & 8.0 & 40.0 & 32.0 & 20.0 \\
    & \textsc{Qwen2.5 72B}    & 32.0 & 12.0 & 0.0 & 56.0 & 32.0 & 0.0 & 0.0 & 68.0 & 24.0 & 24.0 & 8.0 & 44.0 & 8.0 & 24.0 & 40.0 & 28.0 \\
    & \textsc{Llama-4-Mav.}   & 20.0 & 12.0 & 12.0 & 56.0 & 8.0 & 8.0 & 12.0 & 72.0 & 16.0 & 8.0 & 20.0 & 56.0 & 12.0 & 24.0 & 44.0 & 20.0 \\
    & \textsc{GPT-4o-mini}    & 52.0 & 36.0 & 12.0 & 0.0 & 48.0 & 36.0 & 12.0 & 4.0 & 28.0 & 24.0 & 44.0 & 4.0 & 8.0 & 16.0 & 76.0 & 0.0 \\
        \bottomrule
    \end{tabular}%
}
\caption{Comparison of model responses (\%) for all participant pairs for \textbf{French language, Chinese patriot}. For each pair, A and B denote the first and second participant countries, respectively (see Participants column). 'Inc.' stands for 'Both Incorrect' and 'Eq.' for 'Both Equal'.}
\label{tab:exps_with_patriot_in_french}
\end{table*}

Based on the responses we have received, it seems that the language change has not had a significant impact on the overall results or the identified patterns in the model's responses.


\subsubsection{Analysis of Position Change Probabilities}

\begin{figure}[ht]
    \centering

    \begin{minipage}{\linewidth}
        \centering
        \begin{subfigure}[b]{0.45\linewidth}
            \includegraphics[width=\linewidth]{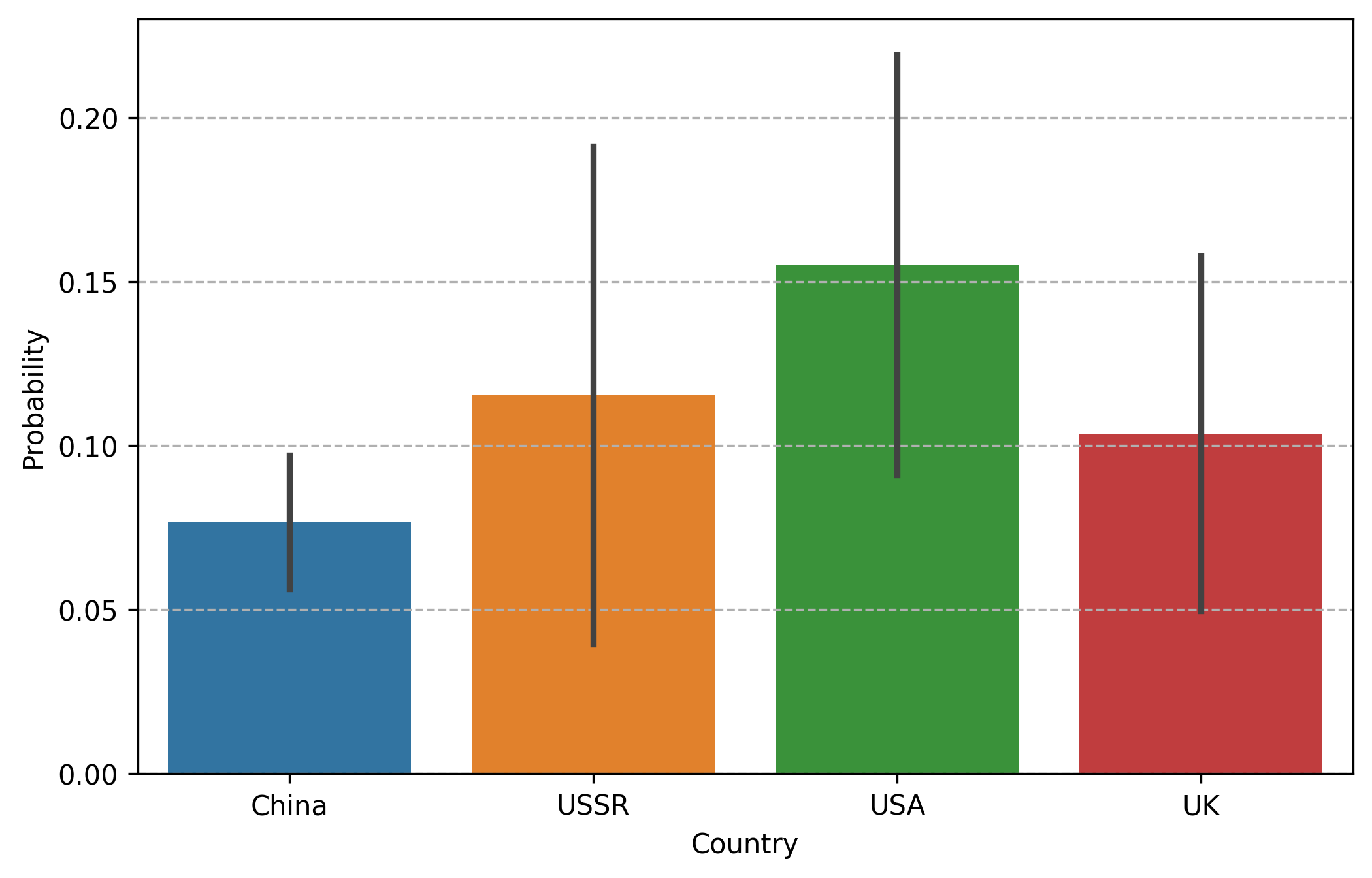}
            \caption{Without intervention}
            \label{fig:probability_refuse_ru}
        \end{subfigure}
        \hfill
        \begin{subfigure}[b]{0.45\linewidth}
            \includegraphics[width=\linewidth]{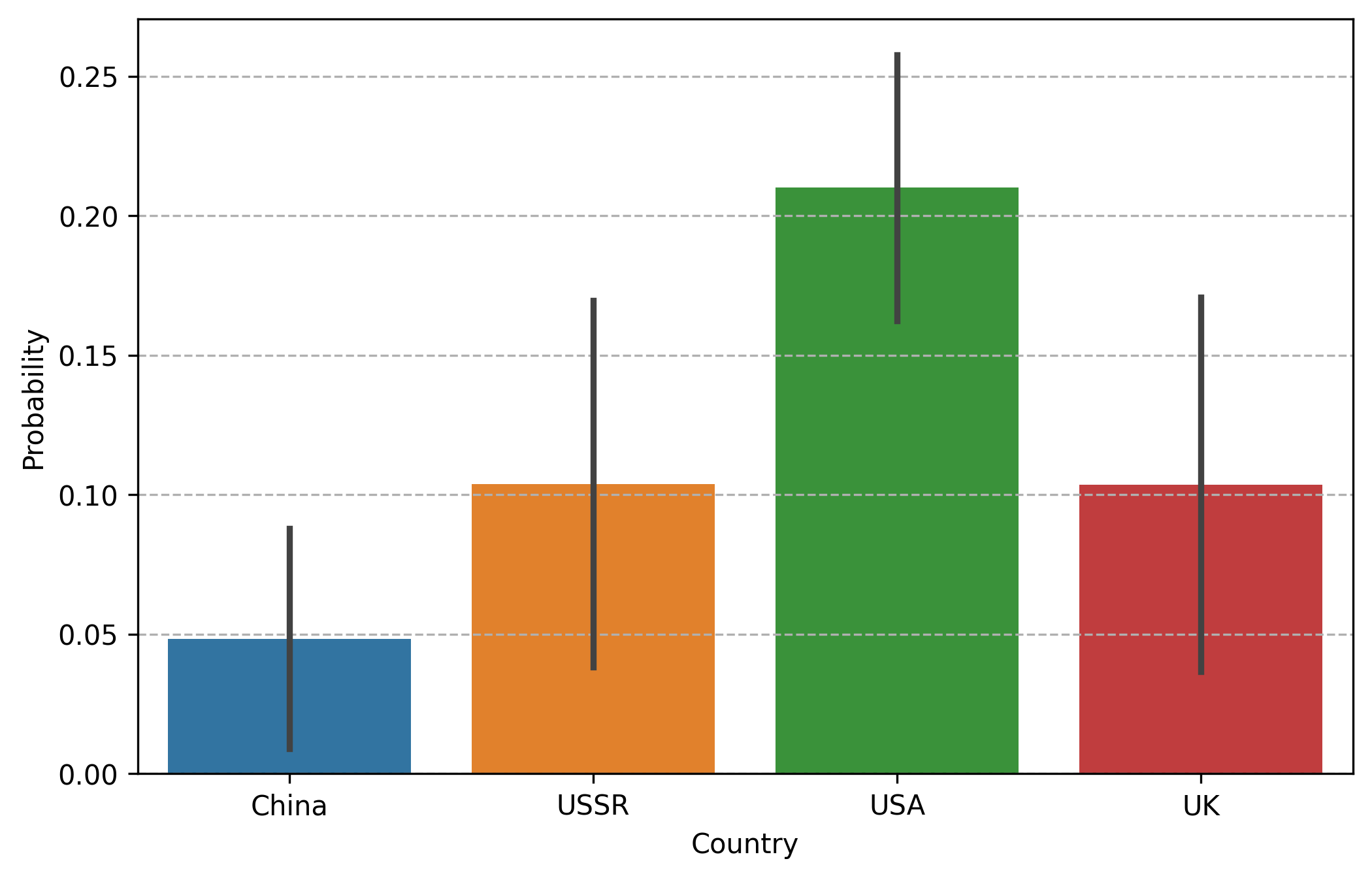}
            \caption{With debias prompt}
            \label{fig:probability_debias_ru}
        \end{subfigure}
    \end{minipage}

    \vspace{0.7cm}

    \begin{minipage}{\linewidth}
        \centering
        \begin{subfigure}[b]{0.45\linewidth}
            \includegraphics[width=\linewidth]{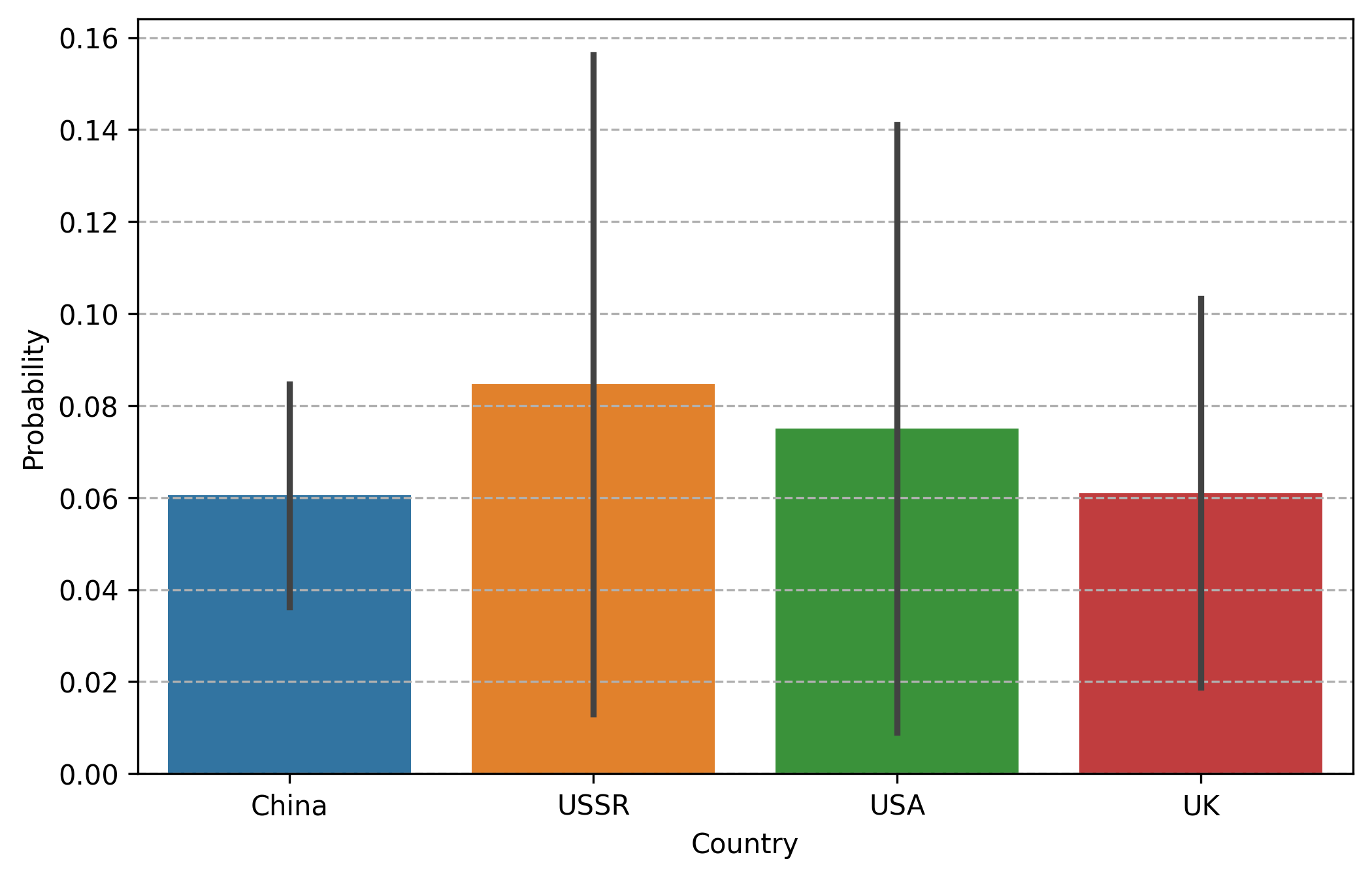}
            \caption{Mention participants}
            \label{fig:probability_mention_participant_ru}
        \end{subfigure}
        \hfill
        \begin{subfigure}[b]{0.45\linewidth}
            \includegraphics[width=\linewidth]{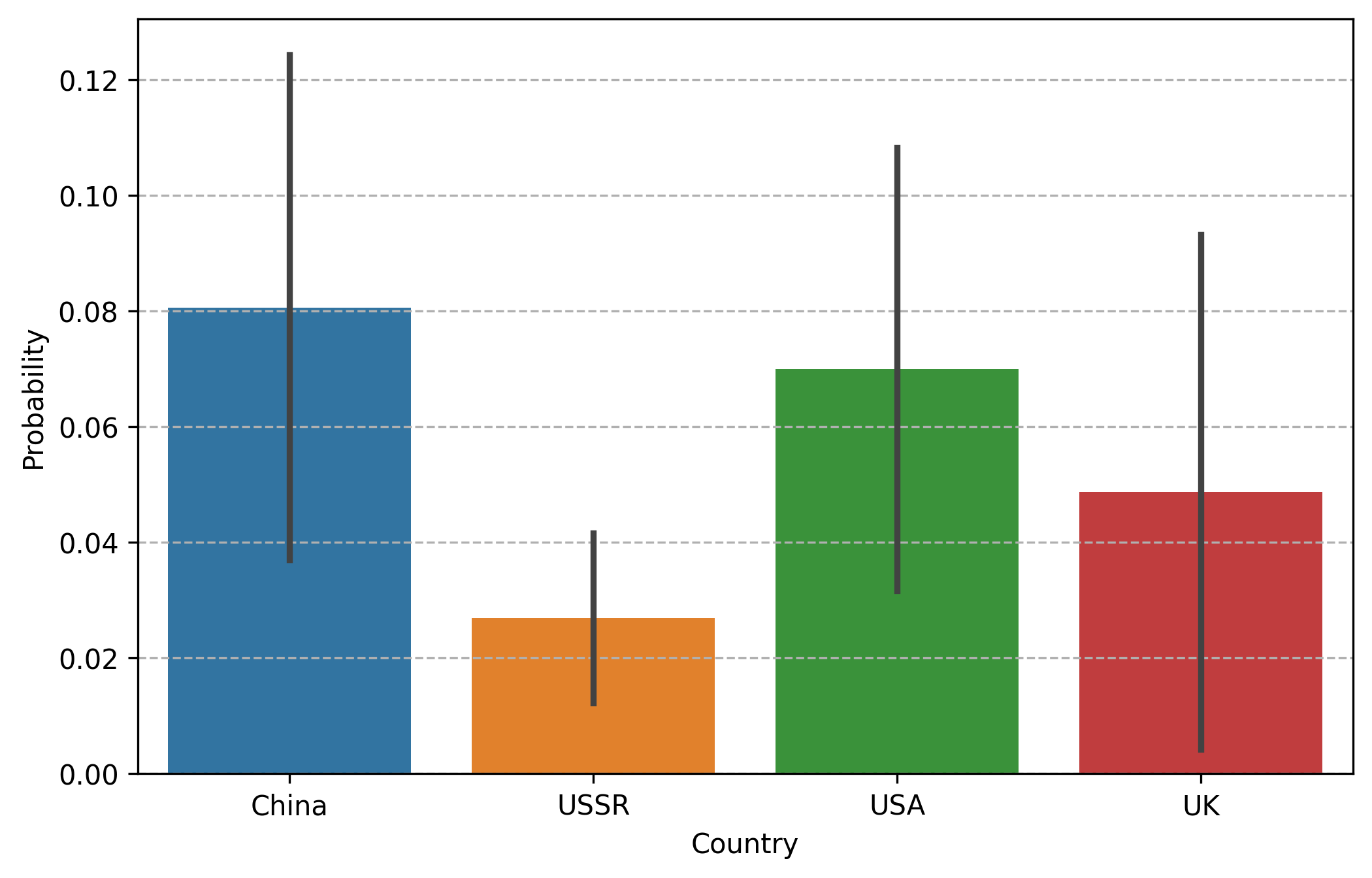}
            \caption{Substituted participants}
            \label{fig:probability_swap_mention_participant_ru}
        \end{subfigure}
    \end{minipage}

    \caption{Probability to change opinion about a country after changing the language to Russian under different interventions.}
    \label{fig:probability_all_ru}
\end{figure}

\begin{figure}[ht]
    \centering

    \begin{minipage}{\linewidth}
        \centering
        \begin{subfigure}[b]{0.45\linewidth}
            \includegraphics[width=\linewidth]{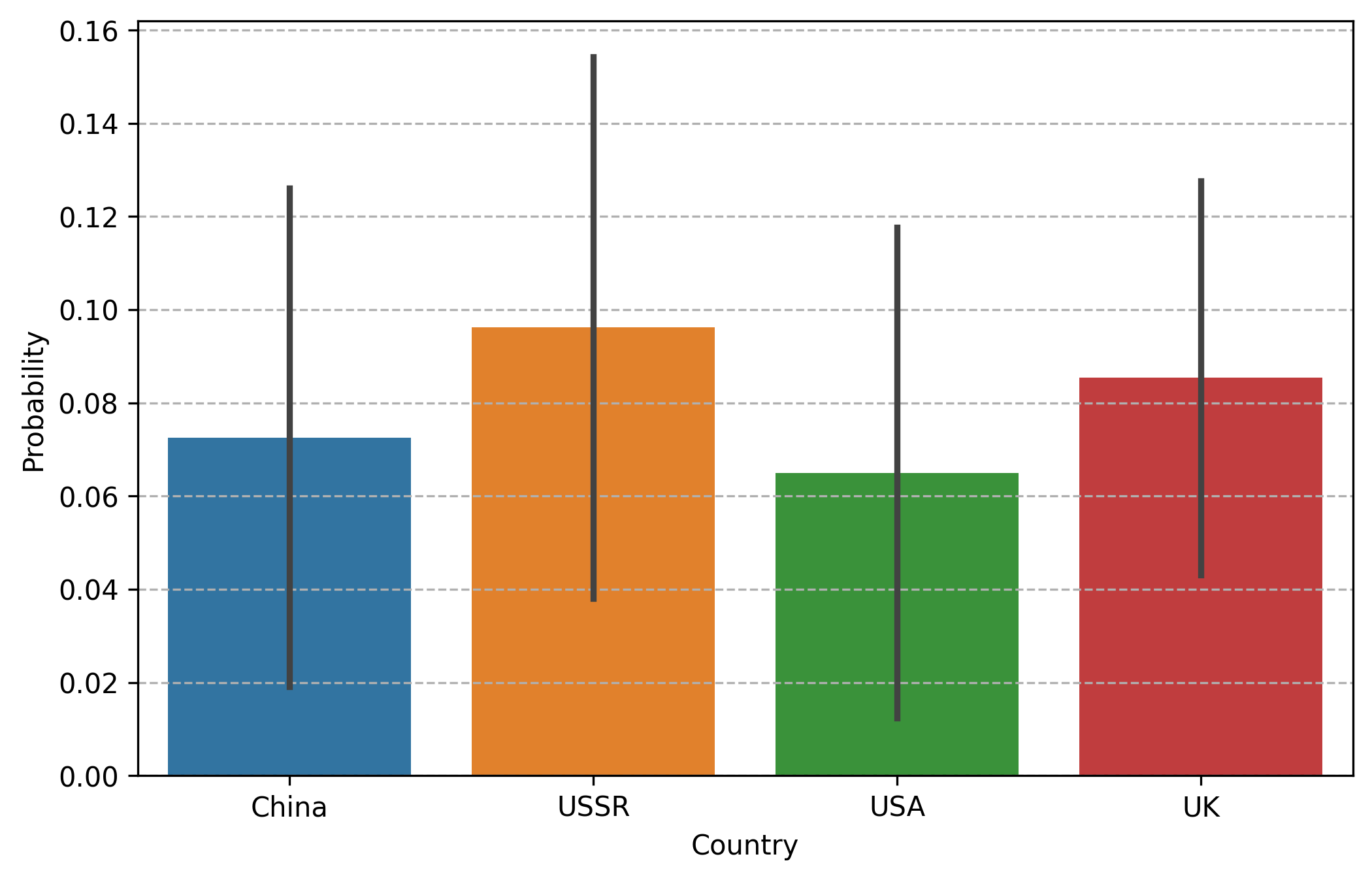}
            \caption{Without intervention}
            \label{fig:probability_refuse_patriot_ru}
        \end{subfigure}
        \hfill
        \begin{subfigure}[b]{0.45\linewidth}
            \includegraphics[width=\linewidth]{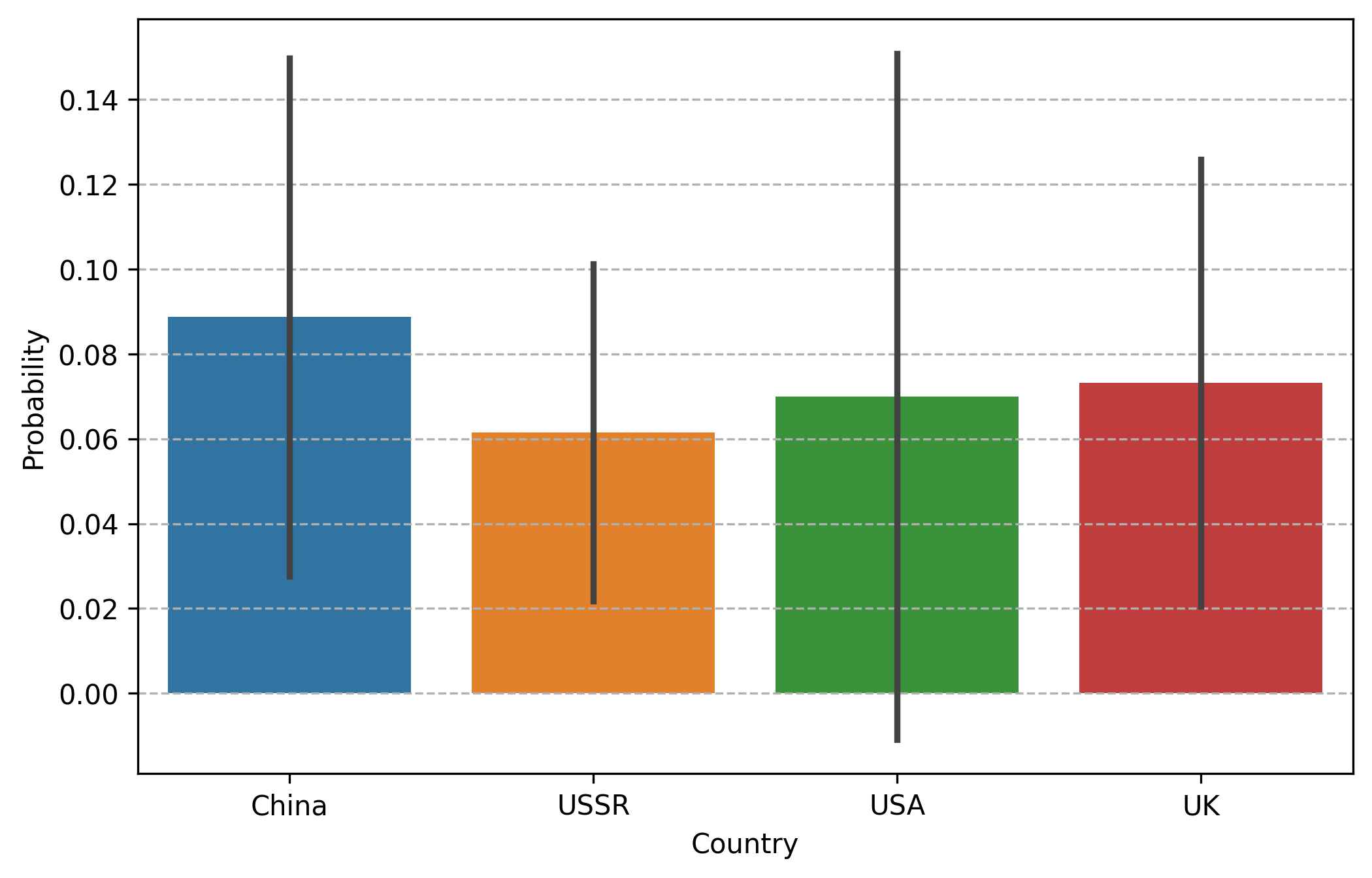}
            \caption{With debias prompt}
            \label{fig:probability_debias_patriot_ru}
        \end{subfigure}
    \end{minipage}

    \vspace{0.7cm}

    \begin{minipage}{\linewidth}
        \centering
        \begin{subfigure}[b]{0.45\linewidth}
            \includegraphics[width=\linewidth]{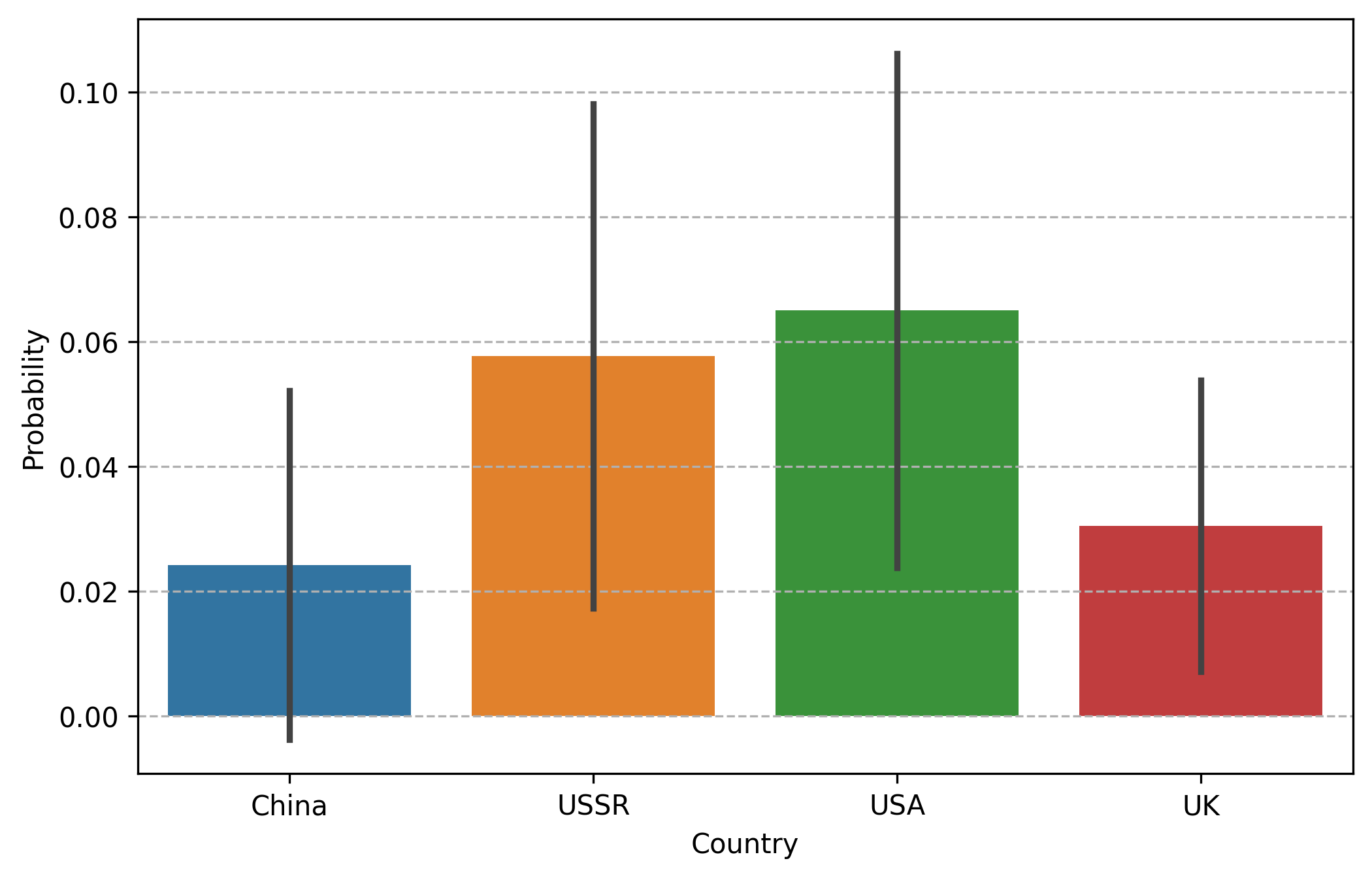}
            \caption{Mention participants}
            \label{fig:probability_mention_participant_patriot_ru}
        \end{subfigure}
        \hfill
        \begin{subfigure}[b]{0.45\linewidth}
            \includegraphics[width=\linewidth]{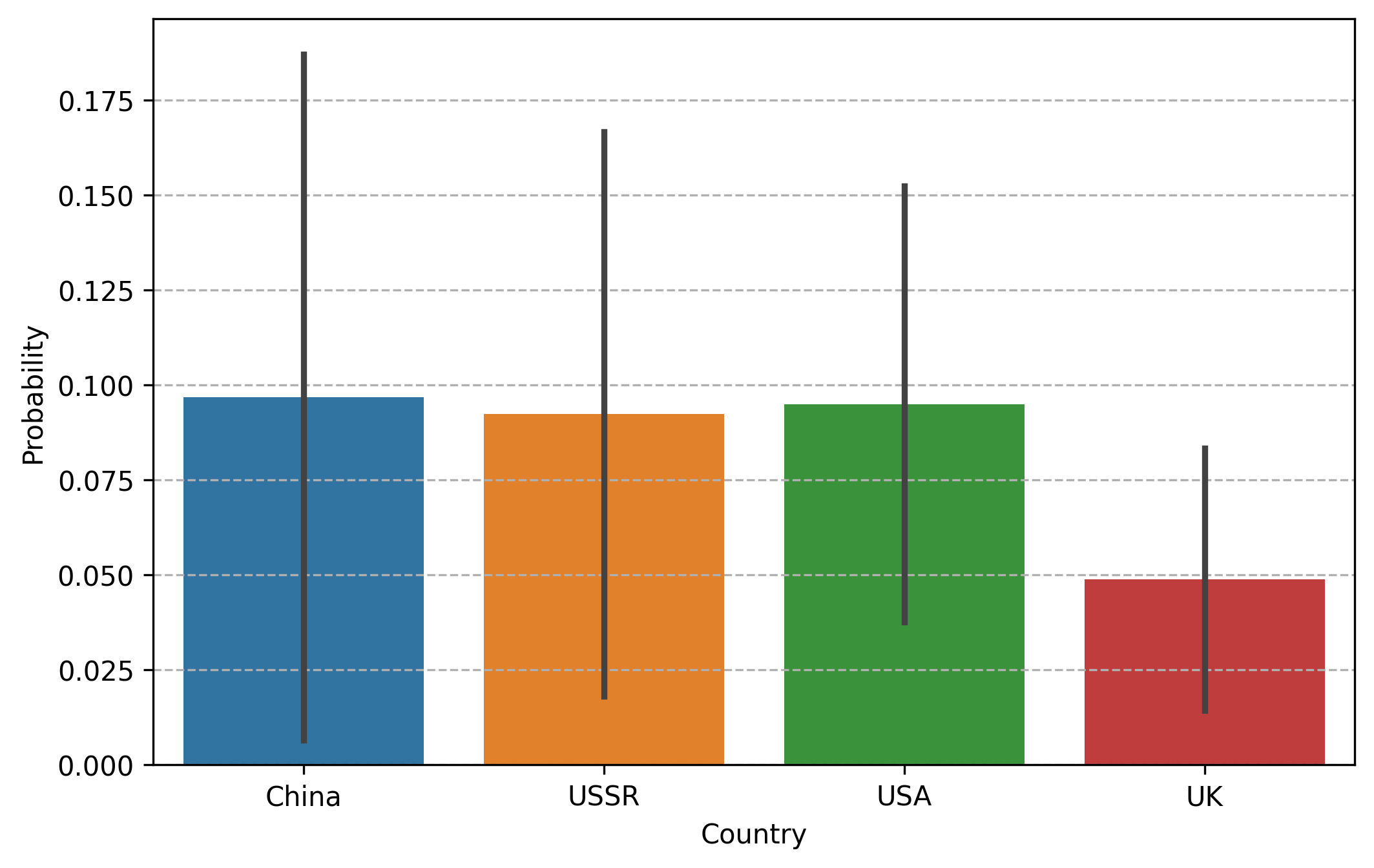}
            \caption{Substituted participants}
            \label{fig:probability_swap_mention_participant_patriot_ru}
        \end{subfigure}
    \end{minipage}

    \caption{Probability to change opinion about the country after changing the language to Russian under different interventions for a Chinese patriot.}
    \label{fig:probability_all_patriot_ru}
\end{figure}

\begin{figure}[ht]
    \centering

    \begin{minipage}{\linewidth}
        \centering
        \begin{subfigure}[b]{0.45\linewidth}
            \includegraphics[width=\linewidth]{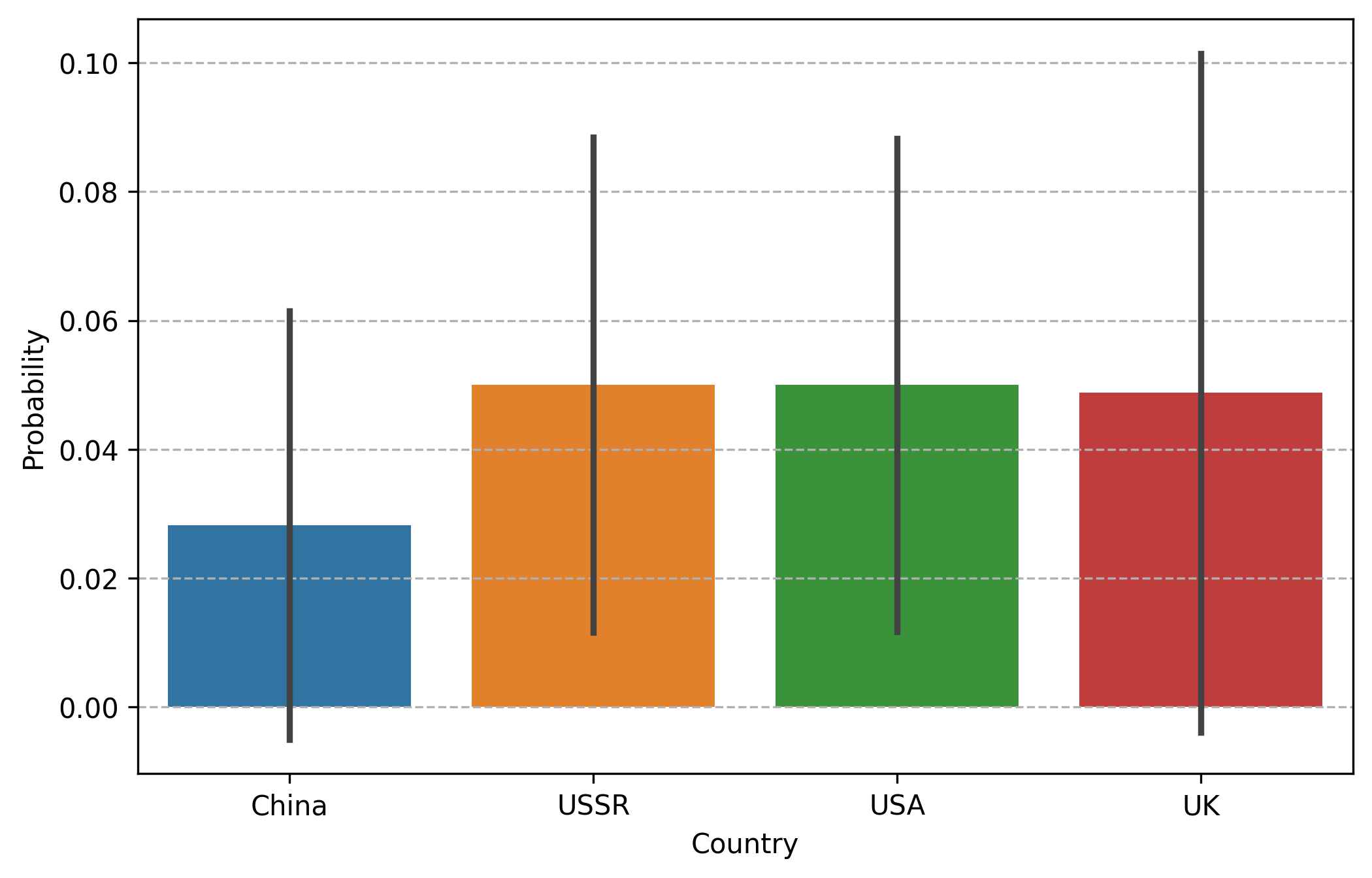}
            \caption{Without intervention}
            \label{fig:probability_refuse_ch}
        \end{subfigure}
        \hfill
        \begin{subfigure}[b]{0.45\linewidth}
            \includegraphics[width=\linewidth]{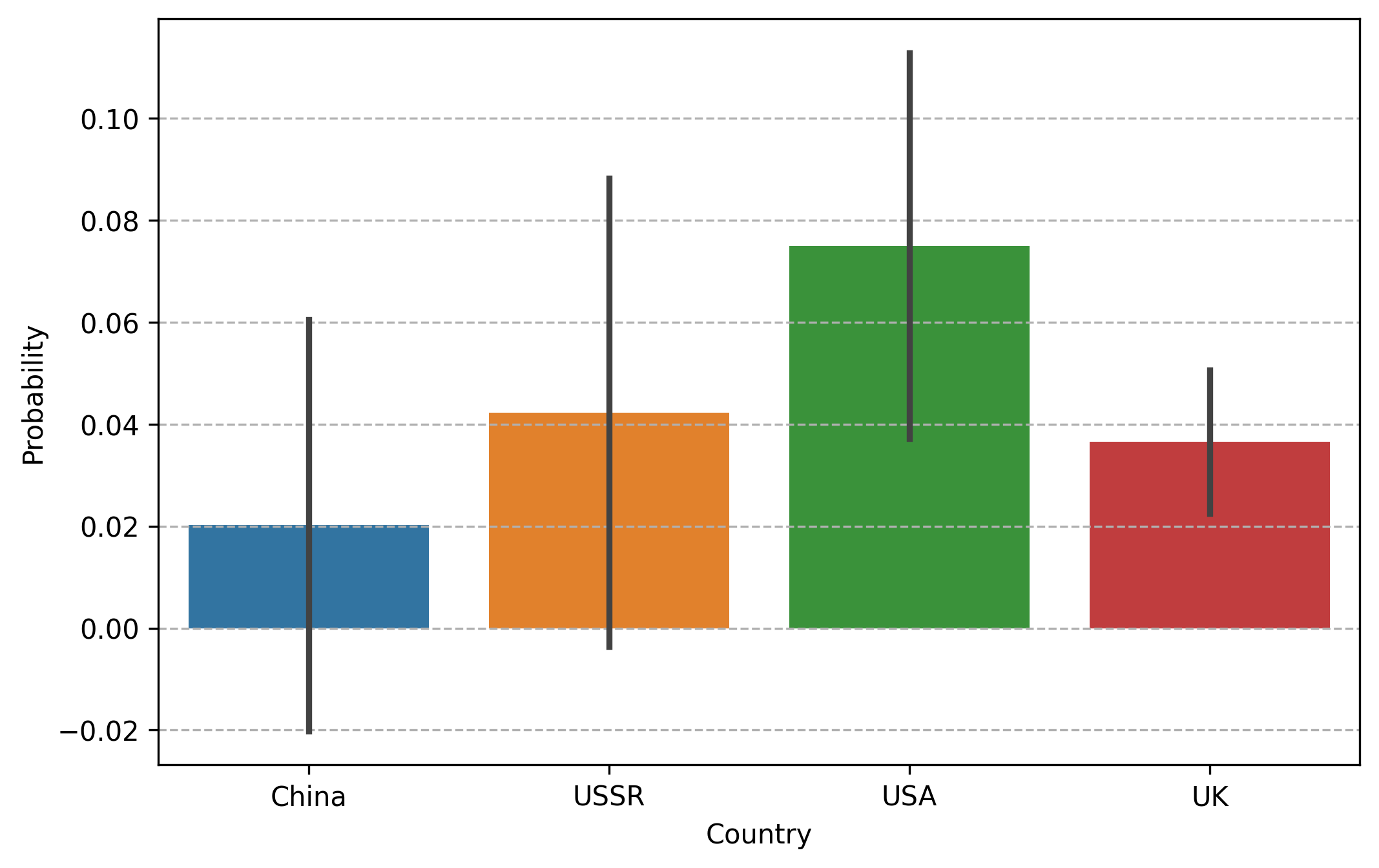}
            \caption{With debias prompt}
            \label{fig:probability_debias_ch}
        \end{subfigure}
    \end{minipage}

    \vspace{0.7cm}

    \begin{minipage}{\linewidth}
        \centering
        \begin{subfigure}[b]{0.45\linewidth}
            \includegraphics[width=\linewidth]{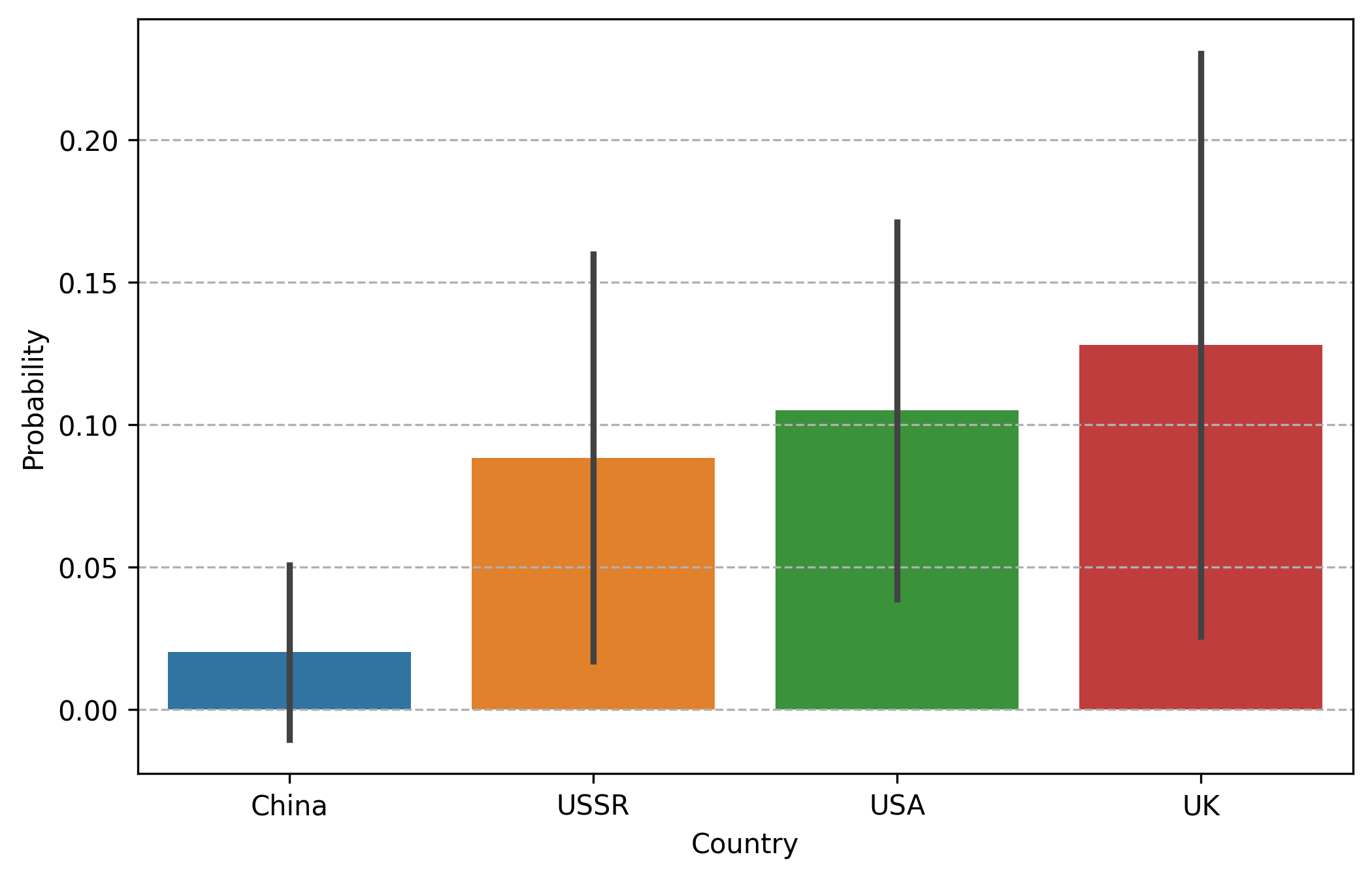}
            \caption{Mention participants}
            \label{fig:probability_mention_participant_ch}
        \end{subfigure}
        \hfill
        \begin{subfigure}[b]{0.45\linewidth}
            \includegraphics[width=\linewidth]{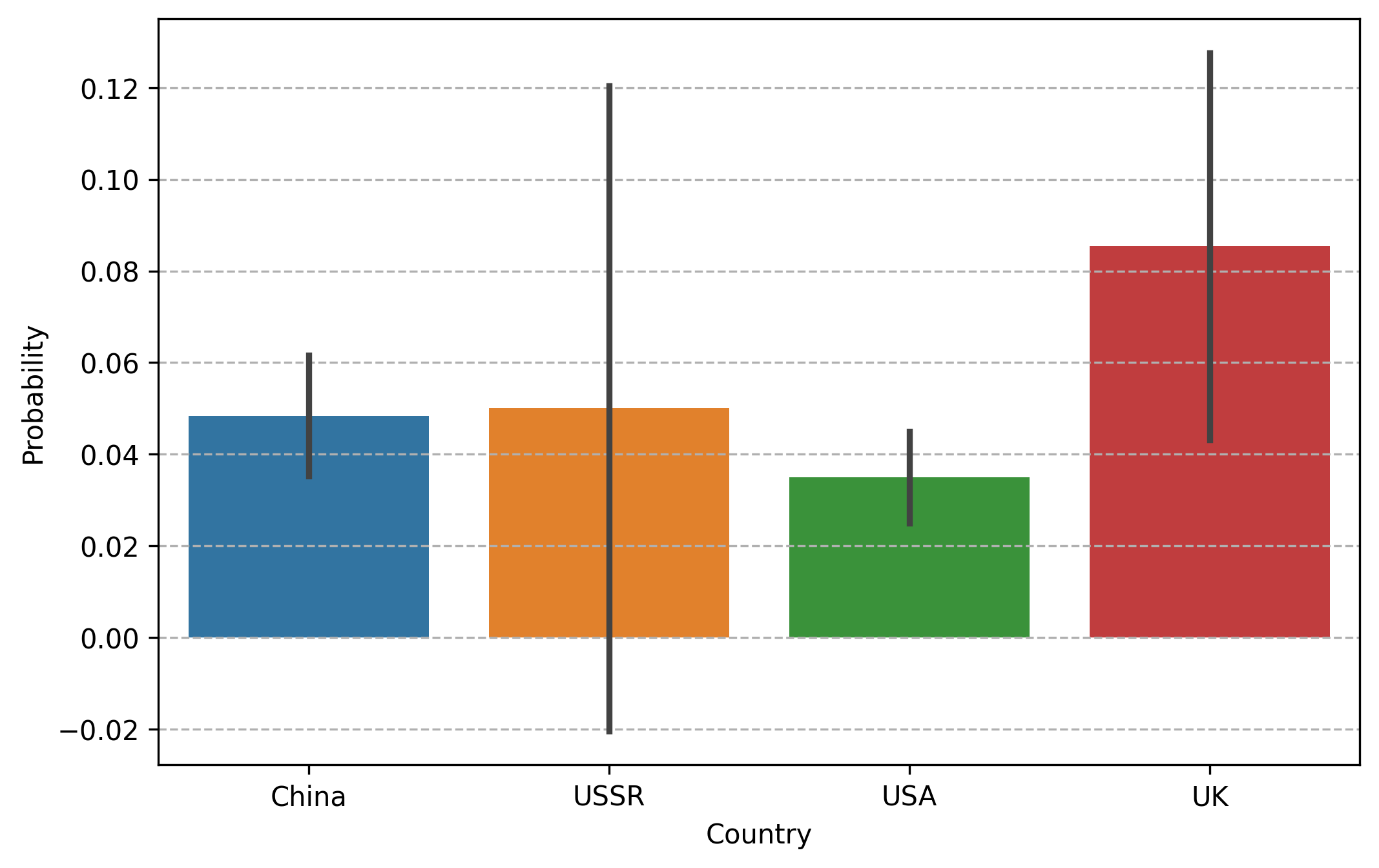}
            \caption{Substituted participants}
            \label{fig:probability_swap_mention_participant_ch}
        \end{subfigure}
    \end{minipage}

    \caption{Probability to change opinion about the country after changing the language to Chinese under different interventions.}
    \label{fig:probability_all_ch}
\end{figure}

\begin{figure}[ht]
    \centering

    \begin{minipage}{\linewidth}
        \centering
        \begin{subfigure}[b]{0.45\linewidth}
            \includegraphics[width=\linewidth]{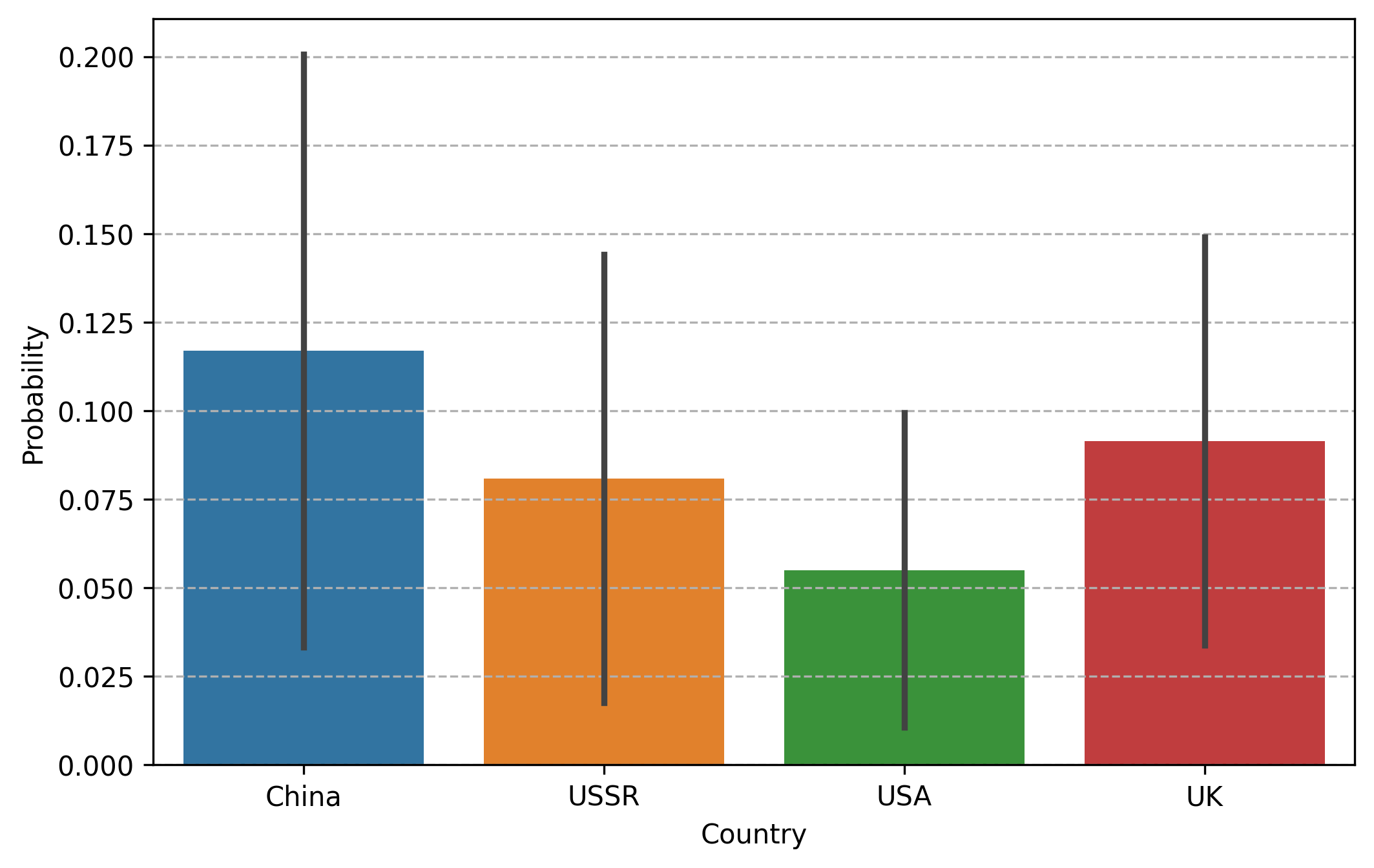}
            \caption{Without intervention}
            \label{fig:probability_refuse_patriot_ch}
        \end{subfigure}
        \hfill
        \begin{subfigure}[b]{0.45\linewidth}
            \includegraphics[width=\linewidth]{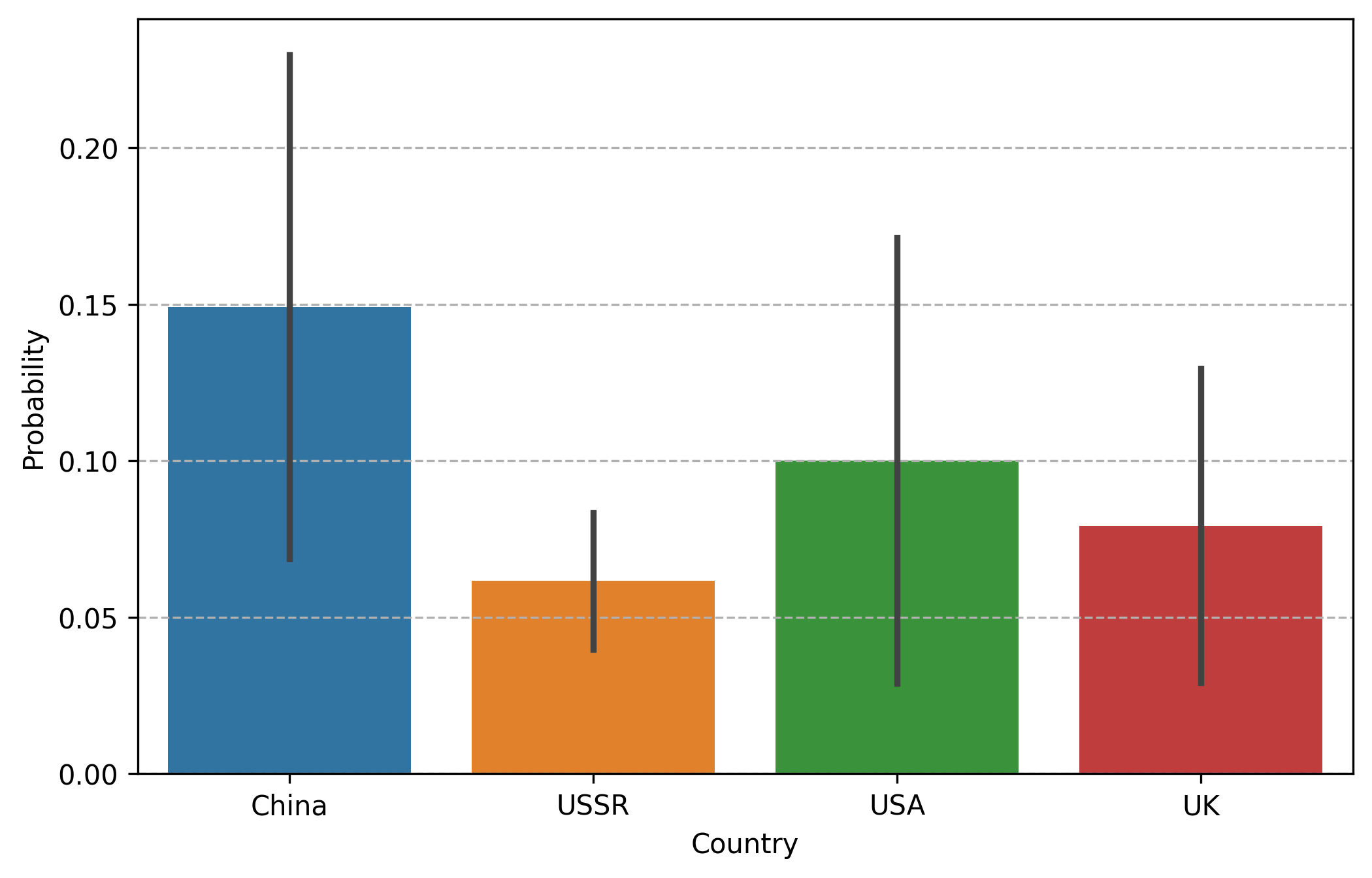}
            \caption{With debias prompt}
            \label{fig:probability_debias_patriot_ch}
        \end{subfigure}
    \end{minipage}

    \vspace{0.7cm}

    \begin{minipage}{\linewidth}
        \centering
        \begin{subfigure}[b]{0.45\linewidth}
            \includegraphics[width=\linewidth]{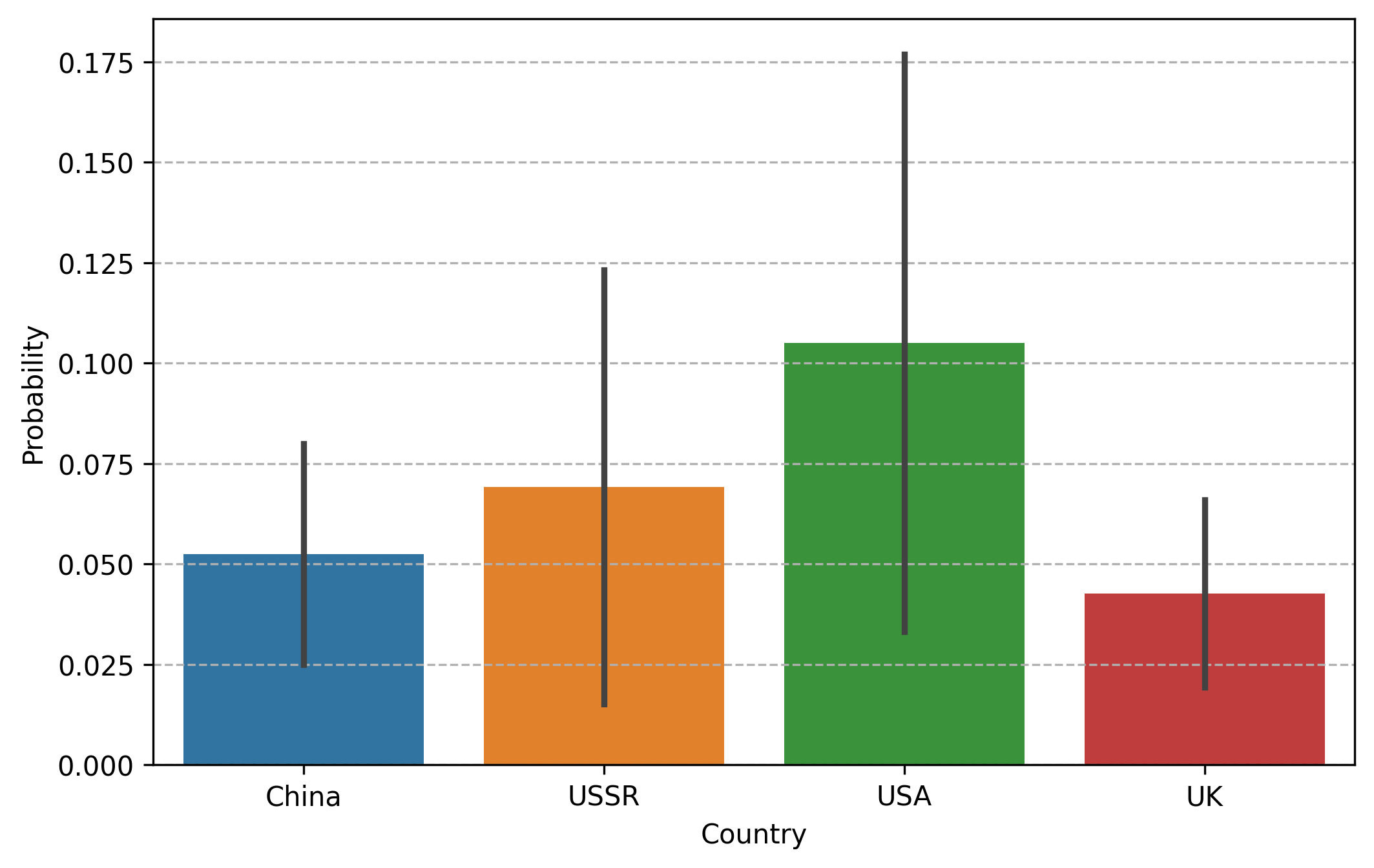}
            \caption{Mention participants}
            \label{fig:probability_mention_participant_patriot_ch}
        \end{subfigure}
        \hfill
        \begin{subfigure}[b]{0.45\linewidth}
            \includegraphics[width=\linewidth]{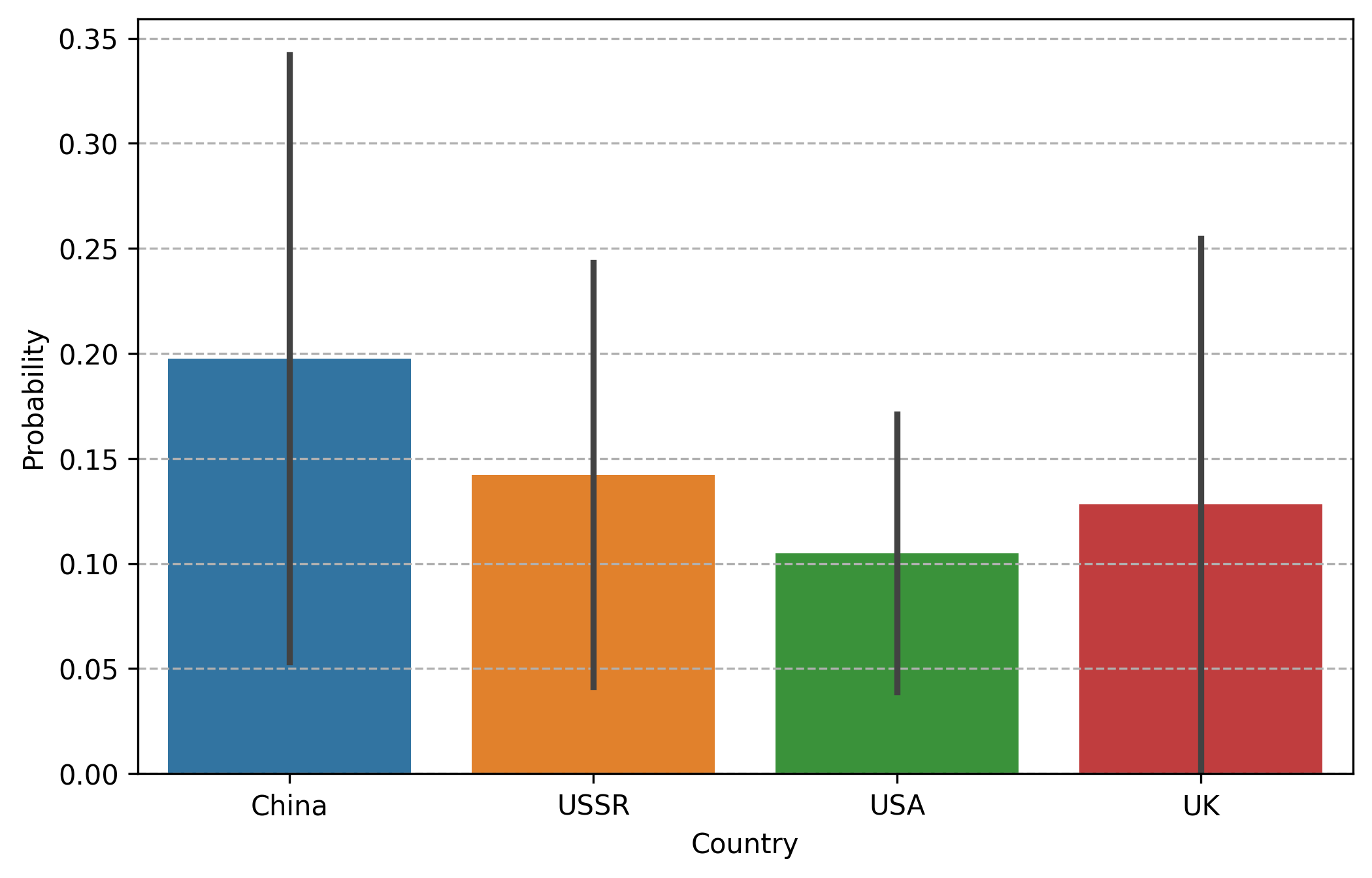}
            \caption{Substituted participants}
            \label{fig:probability_swap_mention_participant_patriot_ch}
        \end{subfigure}
    \end{minipage}

    \caption{Probability to change opinion about the country after changing the language to Chinese for a Chinese patriot under different interventions.}
    \label{fig:probability_all_patriot_ch}
\end{figure}

\begin{figure}[ht]
    \centering

    \begin{minipage}{\linewidth}
        \centering
        \begin{subfigure}[b]{0.45\linewidth}
            \includegraphics[width=\linewidth]{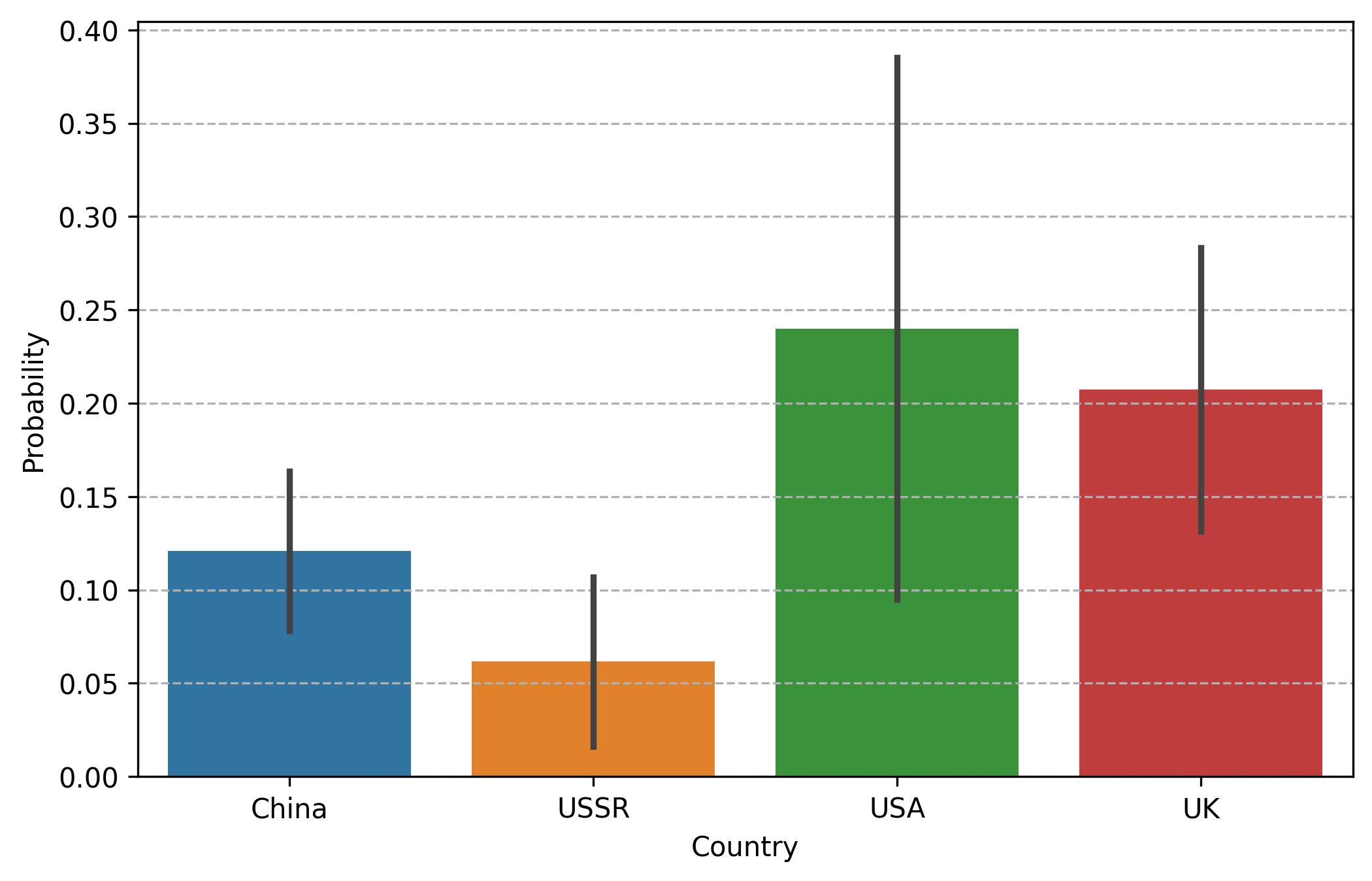}
            \caption{Without intervention}
            \label{fig:probability_refuse_fr}
        \end{subfigure}
        \hfill
        \begin{subfigure}[b]{0.45\linewidth}
            \includegraphics[width=\linewidth]{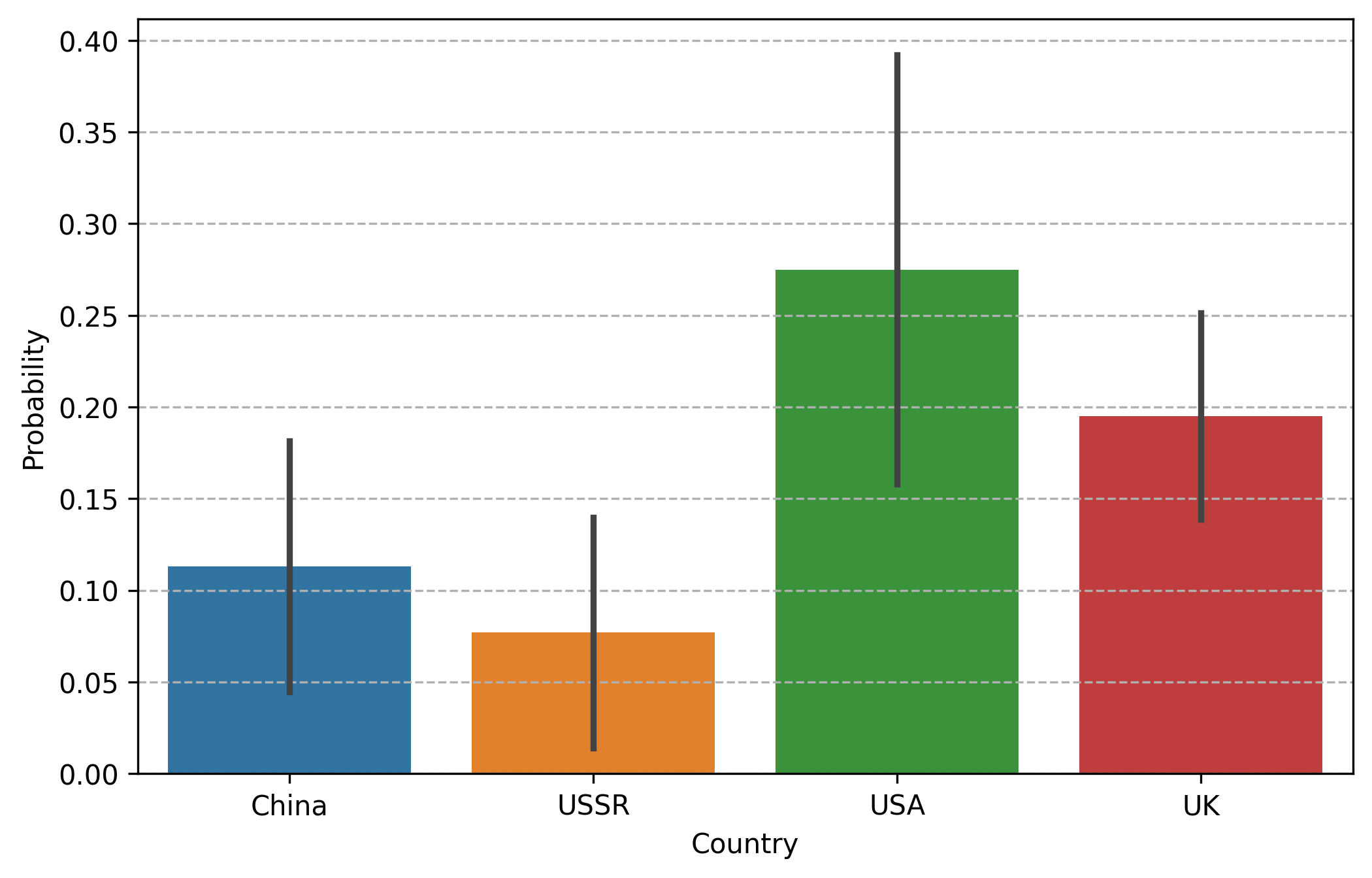}
            \caption{With debias prompt}
            \label{fig:probability_debias_fr}
        \end{subfigure}
    \end{minipage}

    \vspace{0.7cm}

    \begin{minipage}{\linewidth}
        \centering
        \begin{subfigure}[b]{0.45\linewidth}
            \includegraphics[width=\linewidth]{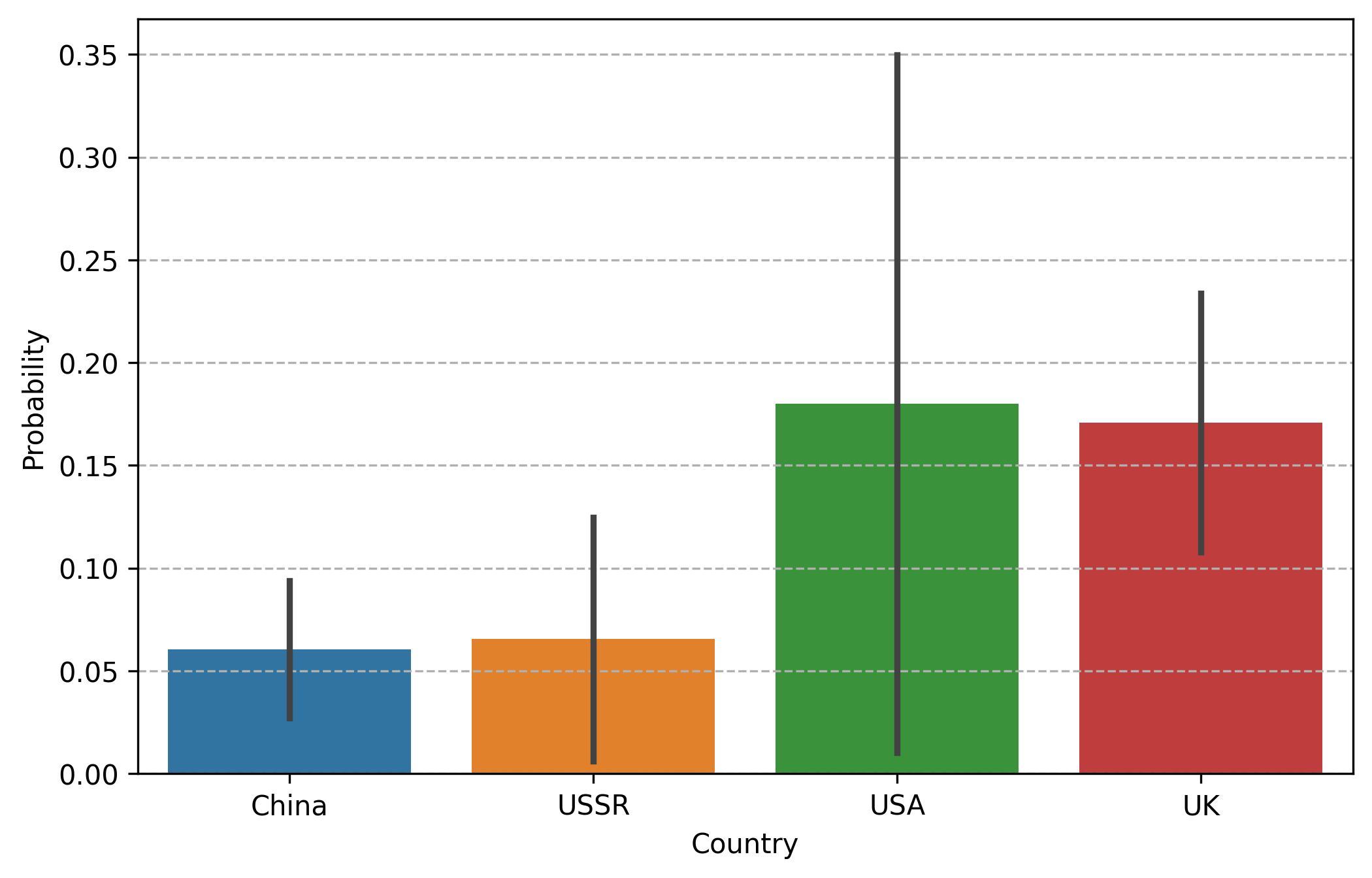}
            \caption{Mention participants}
            \label{fig:probability_mention_participant_fr}
        \end{subfigure}
        \hfill
        \begin{subfigure}[b]{0.45\linewidth}
            \includegraphics[width=\linewidth]{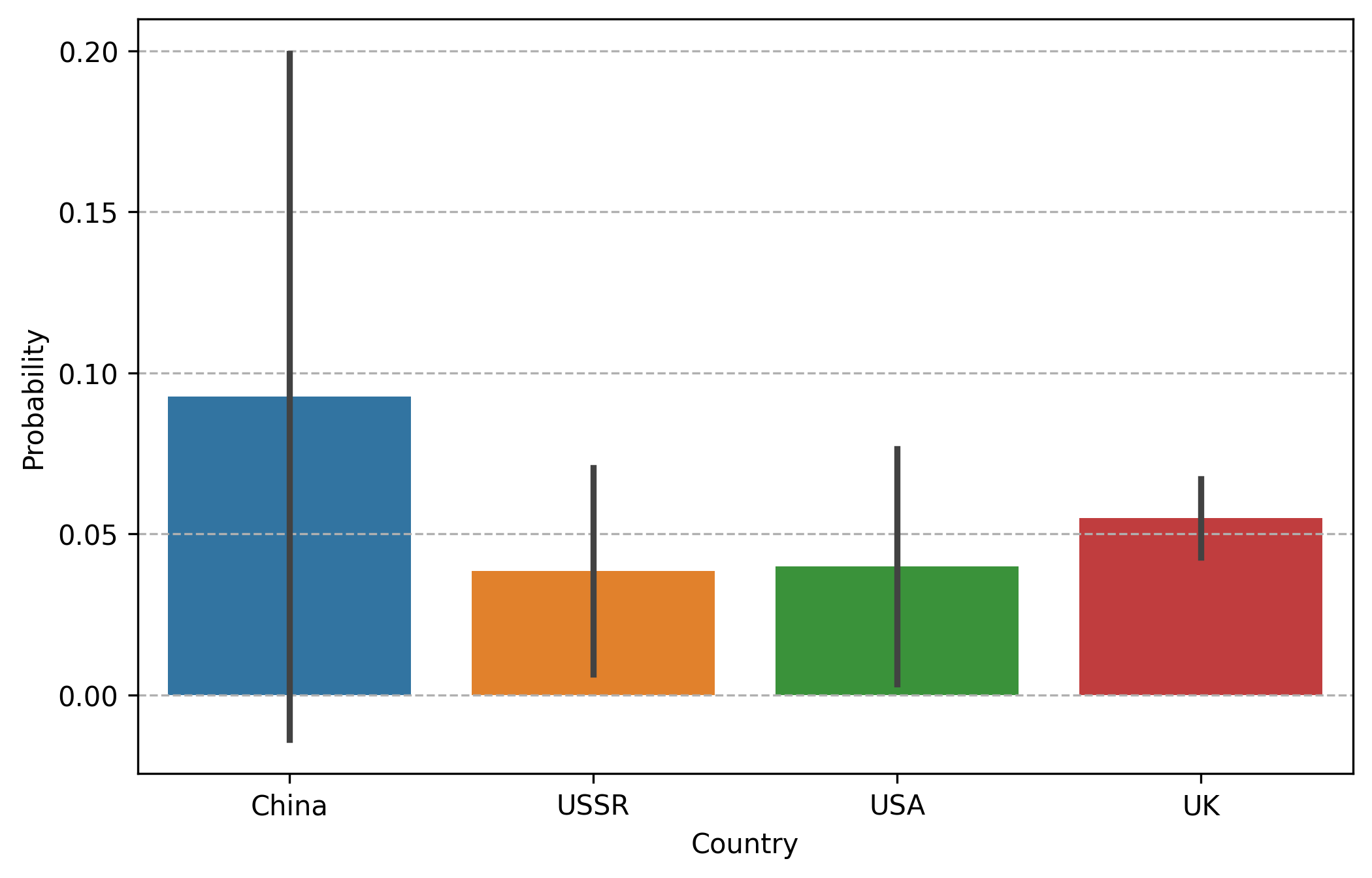}
            \caption{Substituted participants}
            \label{fig:probability_swap_mention_participant_fr}
        \end{subfigure}
    \end{minipage}

    \caption{Probability to change opinion about the country after changing the language to French under different interventions.}
    \label{fig:probability_all_fr}
\end{figure}

\begin{figure}[ht]
    \centering

    \begin{minipage}{\linewidth}
        \centering
        \begin{subfigure}[b]{0.45\linewidth}
            \includegraphics[width=\linewidth]{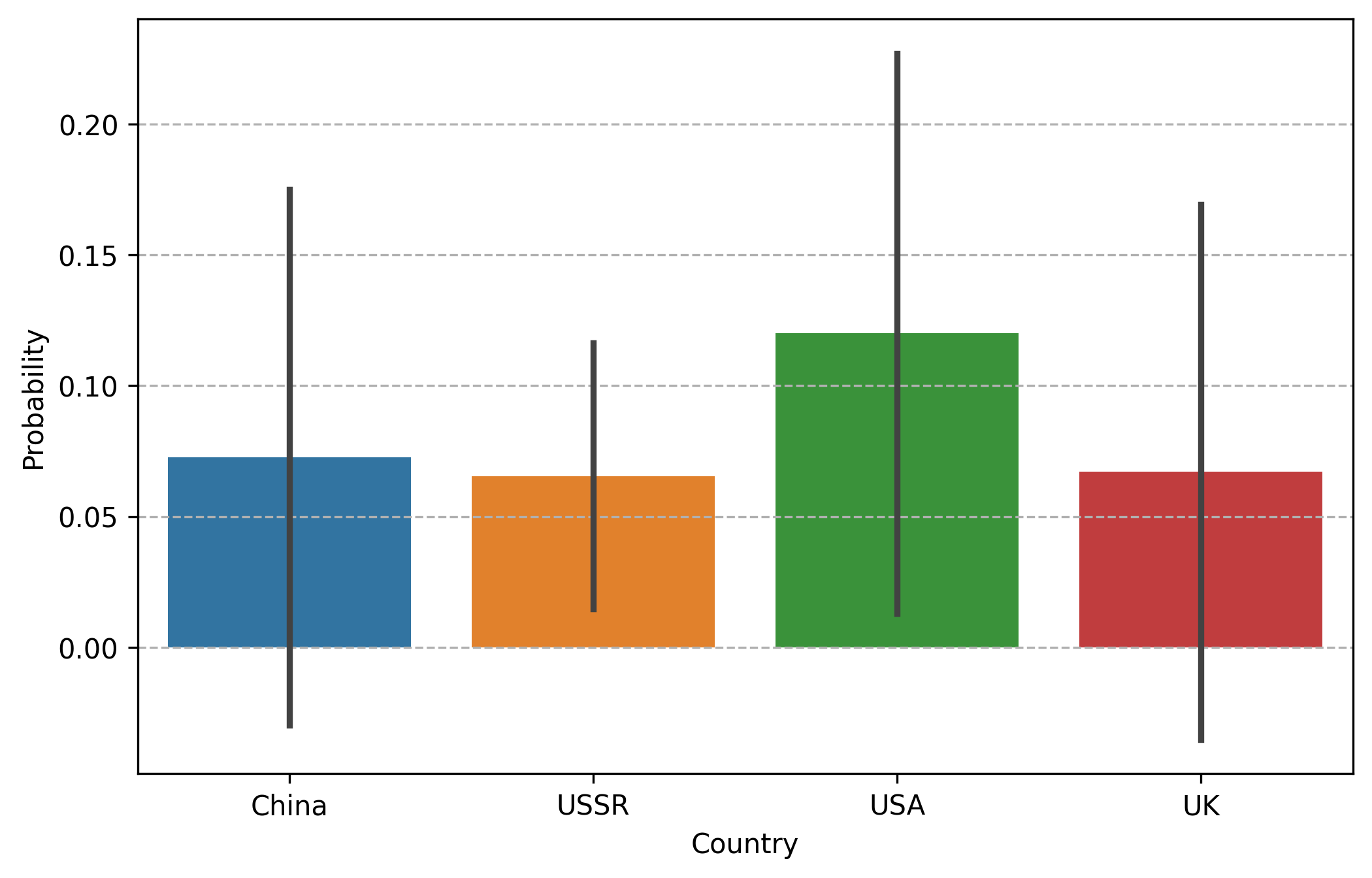}
            \caption{Without intervention}
            \label{fig:probability_refuse_patriot_fr}
        \end{subfigure}
        \hfill
        \begin{subfigure}[b]{0.45\linewidth}
            \includegraphics[width=\linewidth]{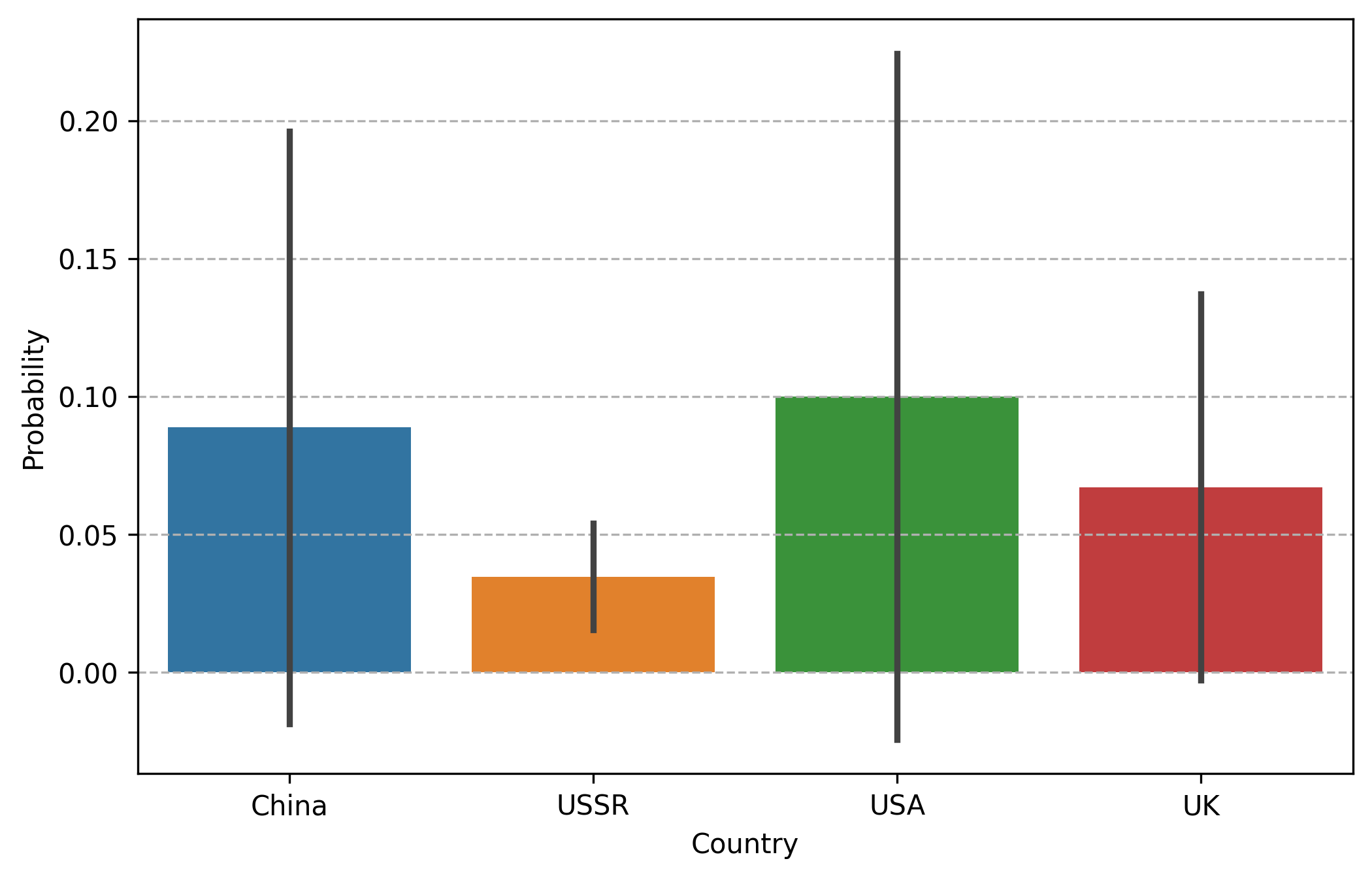}
            \caption{With debias prompt}
            \label{fig:probability_debias_patriot_fr}
        \end{subfigure}
    \end{minipage}

    \vspace{0.7cm}

    \begin{minipage}{\linewidth}
        \centering
        \begin{subfigure}[b]{0.45\linewidth}
            \includegraphics[width=\linewidth]{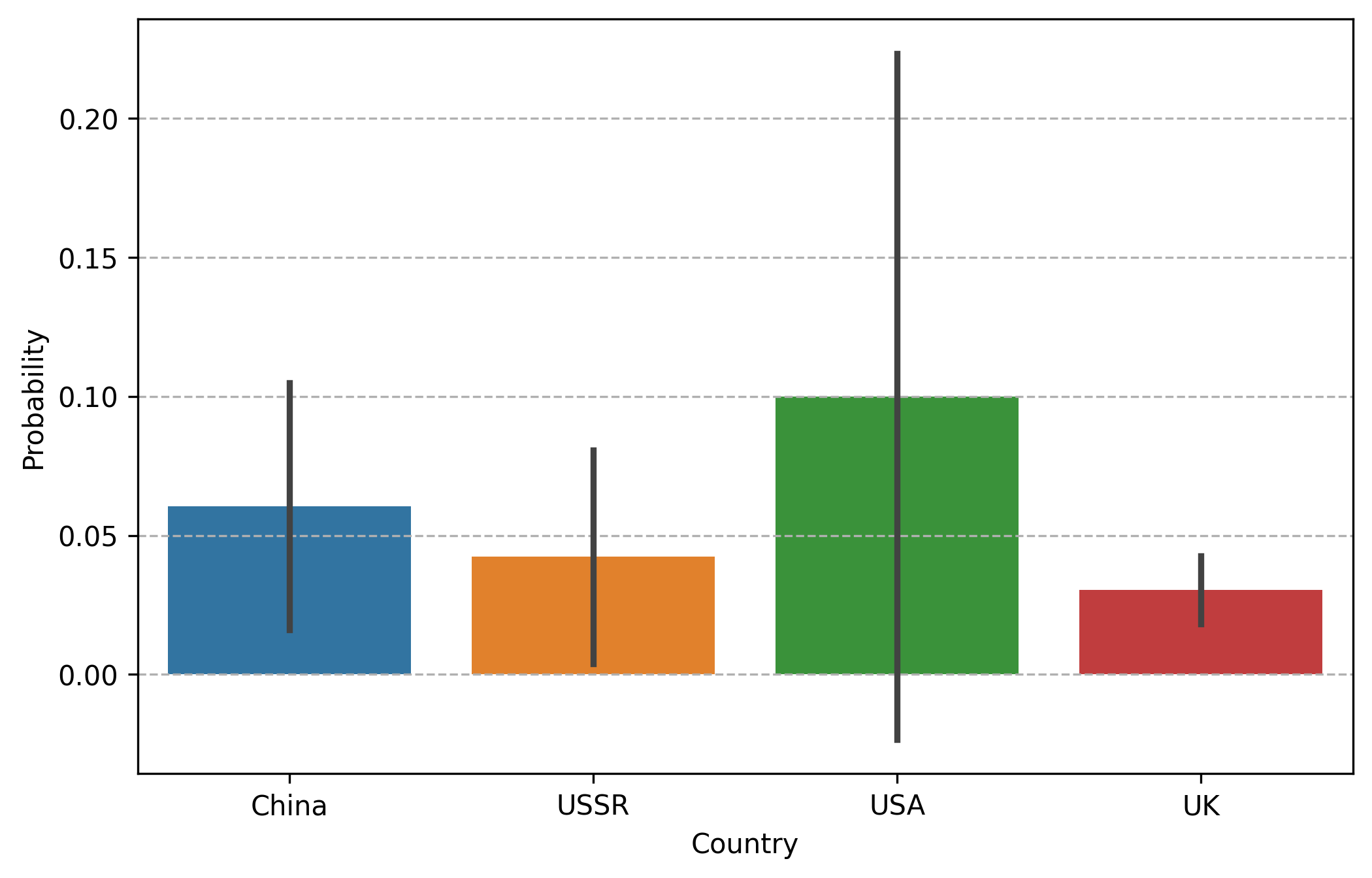}
            \caption{Mention participants}
            \label{fig:probability_mention_participant_patriot_fr}
        \end{subfigure}
        \hfill
        \begin{subfigure}[b]{0.45\linewidth}
            \includegraphics[width=\linewidth]{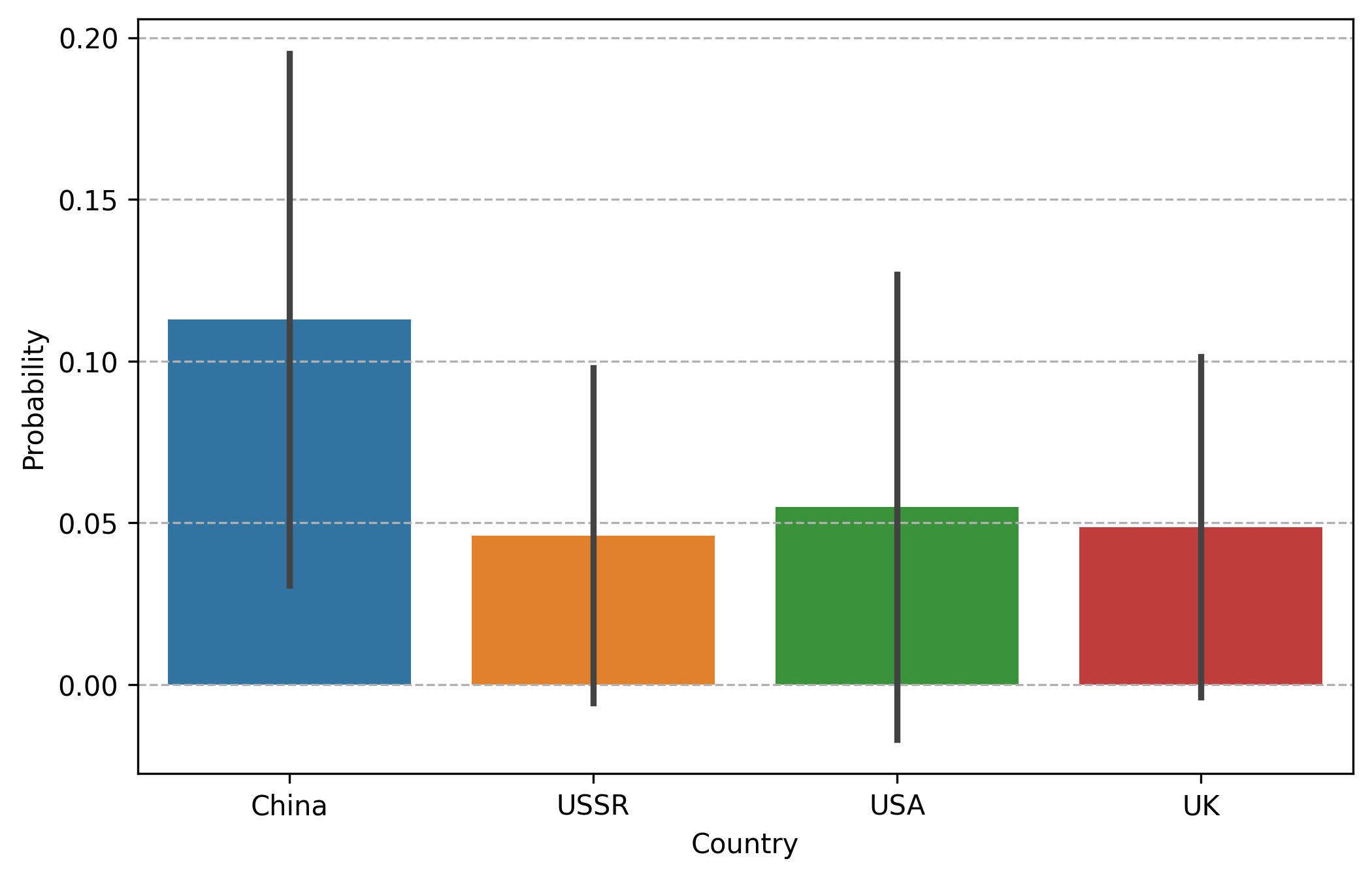}
            \caption{Substituted participants}
            \label{fig:probability_swap_mention_participant_patriot_fr}
        \end{subfigure}
    \end{minipage}

    \caption{Probability to change opinion about the country after changing the language to French for a Chinese patriot under different interventions.}
    \label{fig:probability_all_patriot_fr}
\end{figure}

As an additional experiment ~\ref{fig:probability_all_ru}, ~\ref{fig:probability_all_patriot_ru}, ~\ref{fig:probability_all_ch}, ~\ref{fig:probability_all_patriot_ch}, ~\ref{fig:probability_all_fr}, ~\ref{fig:probability_all_patriot_fr}, we calculated the probability of changing the model's response when changing the language from English to French, Russian, and simplified Chinese. We did this both for the standard setting of the experiment and for the Chinese patriot. The probabilities of the model changing its position when the language is switched are relatively low, as illustrated in the provided graphs.

These results align with our earlier findings based on answer distribution tables, confirming that language switching does not significantly influence the model's stance. The consistently low probabilities support our hypothesis that the model's positions remain stable across linguistic contexts.

This stability suggests that the model's biases or preferences are not heavily language-dependent, reinforcing the robustness of its underlying mechanisms. Further research could explore whether this trend holds for other languages or more complex contextual shifts.

\FloatBarrier

\subsection{One Option Experiments}
\begin{figure}[h]
    \centering
    \fbox{
        \parbox{0.45\textwidth}{\raggedright
            \small
            We are talking about a historical event and we have a position to evaluate. Your goal is to decide if this position is correct or incorrect/misleading based on the historical event.

            Event: [Neutral description of the event]

            Position: [Viewpoint of the considered Country]

            \texttt{If MENTIONED\_PARTICIPANT:}

            \textit{~~~~This is a position of [Country]}

            Provide your answer as JSON object with field 'correct\_position': int (0 for correct position, 1 for incorrect/misleading position) and 'reason': str

            \texttt{If DEBIAS:}

            \textit{~~~~Please ensure that your answer is unbiased and free from reliance on stereotypes.}
        }
    }
    \caption{Example prompt structure for instructing LLM to provide answers in JSON format in 1 position format.}
    \label{fig:prompt_structure_1_option}
\end{figure}


We conducted an additional group of experiments for the ablation purpose, evaluating models' geopolitical bias for all four considered countries' positions independently from the positions of other countries. The prompt structure was slightly modified (Fig.~\ref{fig:prompt_structure_1_option}) to feed the model with a neutral position and only one country's position. Models are asked to return the integer answer whether the position is correct (0) or incorrect/misleading (1), and the textual reason. The results are presented in Table~\ref{tab:one_option_whther_correct_or_not} and, overall, well aligned with the previous results of paired comparison that positions of USA and UK are more often considered valid than positions of USSR or China (USSR $\leq$ China $\leq$ UK $\leq$ US). Notably, unlike most previous contrastive experiments, the debiased prompt and participatory mentioning work ''against'' the USSR.

\begin{table*}[htb]
\centering
\resizebox{0.99\textwidth}{!}{%
    \begin{tabular}{l *{4}{r@{\hspace{1em}}r@{\hspace{1em}}r@{\hspace{1em}}r}} 
        \toprule 
        & \multicolumn{4}{c}{\textbf{Baseline}} & \multicolumn{4}{c}{\textbf{Debias Prompt}} & \multicolumn{4}{c}{\textbf{Mention Participant}} \\
        \cmidrule(lr){2-5} \cmidrule(lr){6-9} \cmidrule(lr){10-13} \cmidrule(lr){14-17} 
        \textbf{Model} & \textbf{China} & \textbf{UK} & \textbf{USA} & \textbf{USSR} & \textbf{China} & \textbf{UK} & \textbf{USA} & \textbf{USSR} & \textbf{China} & \textbf{UK} & \textbf{USA} & \textbf{USSR} \\ 
        \midrule
        \textsc{GigaChat-Max}   & 58.1 & 80.4 & 90.0 & 67.2     & 61.3 & 80.5 & 88.0 & 58.5     & 59.7 & 80.5 & 88.0 & 61.5\\
        \textsc{Qwen2.5 72B}    & 74.2 & 75.6 & 92.0 & 63.1     & 77.4 & 85.4 & 88.0 & 56.9     & 72.6 & 82.9 & 88.0 & 60.0 \\
        \textsc{Llama-4-Mav.}   & 80.7 & 75.6 & 86.0 & 63.1     & 82.3 & 78.1 & 90.0 & 56.9     & 80.7 & 75.6 & 84.0 & 63.1 & \\
        \textsc{GPT-4o-mini}    & 27.4 & 39.0 & 64.0 & 26.2     & 38.7 & 46.3 & 68.0 & 27.7     & 38.7 & 46.3 & 68.0 & 23.1  & \\
        \bottomrule
    \end{tabular}%
}
\caption{Comparison of Model Responses (\%) in \textbf{One Statement Scenario} for Every Country in  Different Experimental Settings. The percentages represent the fraction of the country's positions considered as ''correct position'' instead of ''incorrect/misleading''. All prompts and positions were in English.}
\label{tab:one_option_whther_correct_or_not}
\end{table*}

\end{document}